\documentclass[10pt,twocolumn,letterpaper]{article}

\usepackage{iccv}
\usepackage{times}
\usepackage{epsfig}
\usepackage{xspace}
\usepackage{graphicx}
\usepackage{amsmath}
\usepackage{amssymb}
\usepackage{booktabs}
\usepackage{tabulary}
\usepackage[dvipsnames]{xcolor}
\usepackage[caption=false]{subfig}
\usepackage[british,english,american]{babel}
\usepackage{soul}
\usepackage{xspace}\xspace
\usepackage{adjustbox}
\usepackage{array}

\usepackage{dblfloatfix}
\usepackage{transparent}

\usepackage{algorithm}
\usepackage{listings}
\usepackage{stackengine}
\usepackage{multirow}
\usepackage{algorithmic}

\definecolor{citecolor}{HTML}{0071bc}
\usepackage[pagebackref=false,breaklinks=true,letterpaper=true,colorlinks,citecolor=citecolor,bookmarks=false]{hyperref}

\iccvfinalcopy %

\def\iccvPaperID{7936} %

\definecolor{Highlight}{HTML}{39b54a}  %

\newcolumntype{x}[1]{>{\centering\arraybackslash}p{#1pt}}
\newcolumntype{y}[1]{>{\raggedright\arraybackslash}p{#1pt}}
\newcolumntype{z}[1]{>{\raggedleft\arraybackslash}p{#1pt}}
\newlength\savewidth

\makeatletter\renewcommand\paragraph{\@startsection{paragraph}{4}{\z@}
  {.5em \@plus1ex \@minus.2ex}{-.5em}{\normalfont\normalsize\bfseries}}\makeatother

\ificcvfinal\fi

\newcommand{\imnet}{ImageNet\xspace}
\newcommand{\inet}{ImageNet\xspace}

\definecolor{myblue}{RGB}{60, 87, 153}
\definecolor{myred}{RGB}{248, 16, 29}

\begin{document}

\title{Self-Supervised Pretraining Improves Self-Supervised Pretraining}

\author{Colorado J Reed$^{*1}$ \and  Xiangyu Yue$^{*1}$ \and Ani Nrusimha$^1$ \and Sayna Ebrahimi$^1$ \and Vivek Vijaykumar$^3$ \and Richard Mao$^1$ \and Bo Li$^1$ \and Shanghang Zhang$^1$ \and Devin Guillory$^1$ \and Sean Metzger$^{1,2}$ \and Kurt Keutzer$^1$ \and Trevor Darrell$^1$ }

\maketitle
\ificcvfinal\thispagestyle{empty}\fi

\begin{abstract}
While self-supervised pretraining has proven beneficial for many computer vision tasks, it requires expensive and lengthy computation, large amounts of data, and is sensitive to data augmentation. 
Prior work demonstrates that models pretrained on datasets dissimilar to their target data, such as chest X-ray models trained on ImageNet, underperform models trained from scratch.
Users that lack the resources to pretrain must use existing models with lower performance.
This paper explores Hierarchical PreTraining (HPT), which decreases convergence time and improves accuracy by initializing the pretraining process with an existing pretrained model.
Through experimentation on 16 diverse vision datasets, we show
HPT converges up to 80$\times$ faster, improves accuracy across tasks, and improves the robustness of the self-supervised pretraining process to changes in the image augmentation policy or amount of pretraining data. 
Taken together, HPT provides a simple framework for obtaining better pretrained representations with less computational resources.\footnote{Code and pretrained models are available at \url{https://github.com/cjrd/self-supervised-pretraining}.}

\end{abstract}

\section{Introduction}
Recently, self-supervised pretraining -- an unsupervised pretraining method that self-labels data to learn salient feature representations -- has outperformed supervised pretraining in an increasing number of computer vision applications \cite{chen_simple_2020, chen_improved_2020, caron2020unsupervised}.
These advances come from \emph{instance contrastive learning}, where a model is trained to identify visually augmented images that originated from the same image from a set~\cite{dosovitskiy2015discriminative, wu2018unsupervised}.
Typically, self-supervised pretraining uses unlabeled \emph{source} data to pretrain a network that will be \emph{transferred} to a supervised training process on a \emph{target} dataset. Self-supervised pretraining is particularly useful when labeling is costly, such as in medical and satellite imaging~\cite{medical,cheng2017remote}. 

\begin{figure}[t] 
    \centering
    \includegraphics[width=\linewidth]{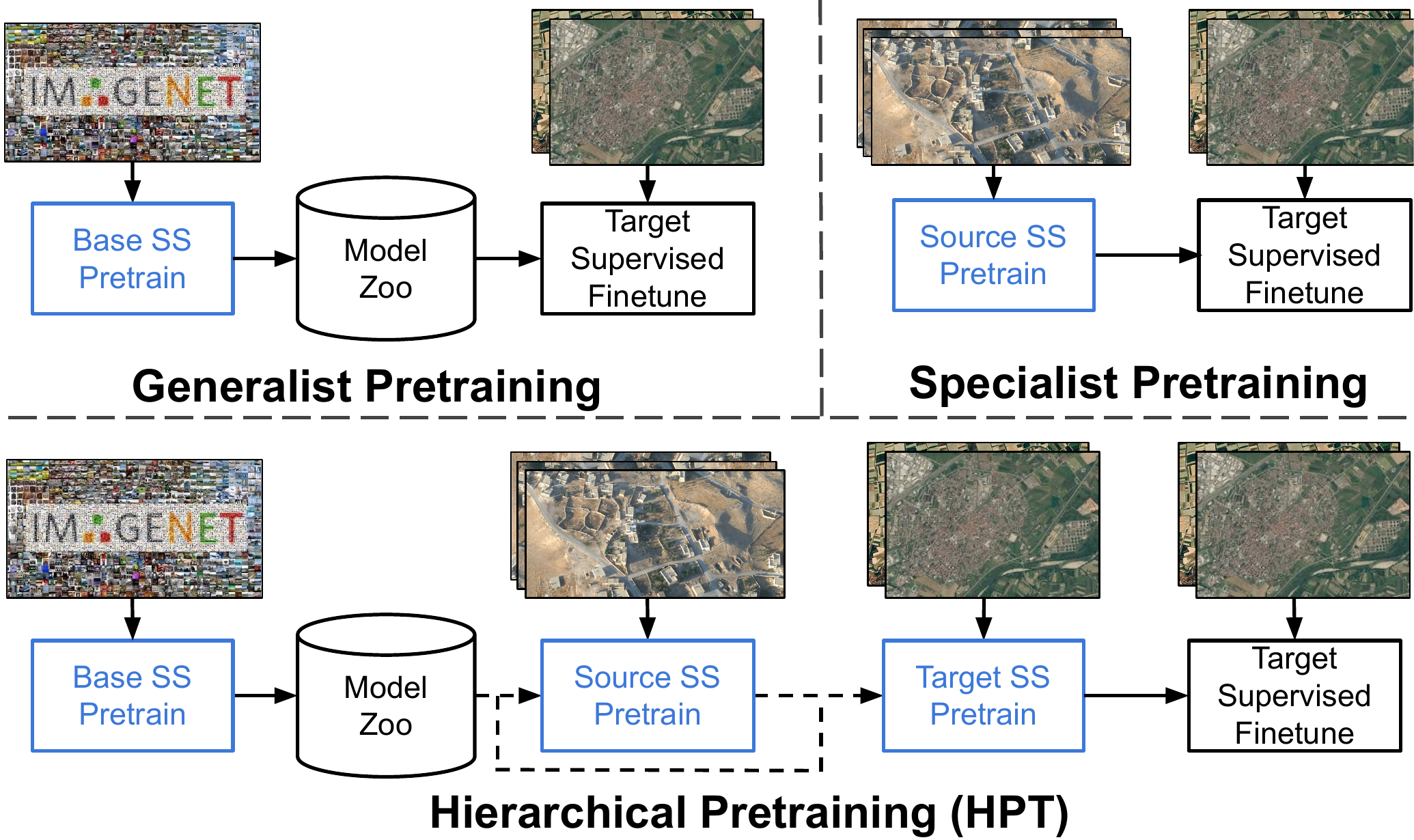}
     \caption{\emph{Methods of using self-supervision.}
     The top row are the two common prior approaches to using self-supervised (SS) pretraining. 
     In \emph{Generalist Pretraining}, a large, general, \emph{base} dataset is used for pretraining, e.g. \inet.
     In \emph{Specialist Pretraining}, a large, specialized \emph{source} dataset is collected and used for pretraining, e.g. aerial images. 
     In this paper, we explore \emph{Hierarchical Pre-Training} (HPT), which sequentially pretrains on datasets that are similar to the target data, thus providing the improved performance of specialist pretraining while leveraging existing generalist models.
     }
     \label{fig:hpt-pipeline}
\end{figure}

However, self-supervised pretraining requires long training time on large datasets, e.g.~SimCLR~\cite{chen_simple_2020} showed improved performance out to $3200$ epochs on \imnet's 1.2 million images~\cite{ILSVRC15}.
In addition, instance contrastive learning is sensitive to the choice of data augmentation policies and many trials are often required to determine good augmentations~\cite{reed2020selfaugment, xiao2020should}.

The computational intensity and sensitivity of self-superivsed pretraining may lead researchers to seek self-supervised models from model zoos and research repositories. However, models pretrained on domain-specific datasets are not commonly available.
In turn, many practitioners do not use a model pretrained on data similar to their target data, but instead, use a pretrained, publicly available model trained on a large, general dataset, such as \imnet.
We refer to this process as \textbf{generalist pretraining}. %
A growing body of research indicates that pretraining on domain-specific datasets, which we refer to as \textbf{specialist pretraining}, leads to improved transfer performance \cite{raghu2019transfusion, neumann2019domain,niu2020distant}.

Figure~\ref{fig:hpt-pipeline} formalizes this categorization of self-supervised pretraining methods.
Generalist and specialist pretraining are as described above, with one round of self-supervised pretraining on a domain-general and domain-specific dataset, respectively.
\textbf{Hierarchical Pretraining} refers to models pretrained on datasets that are progressively more similar to the target data. 
HPT first pretrains on a domain-general dataset (referred to as the \emph{base pretrain}), then optionally pretrains on domain-specific datasets (referred to as the \emph{source pretrain}), before finally pretraining on the target dataset (referred to as the \emph{target pretrain}).
In all cases, pretraining is followed by supervised finetuning on the target task.

Specialist pretraining presents the same core challenge that transfer learning helps alleviate: a sensitive training process that requires large datasets and significant computational resources \cite{kornblith2019better}.
While transfer learning has been carefully investigated in supervised and semi-supervised settings for computer vision~\cite{tan2018survey}, it has not been formally studied for self-supervised pretraining, itself. 
Furthermore, several recent papers that apply self-supervised learning to domain-specific problems did not apply transfer learning to the pretraining process itself, which motivated our work~\cite{tao2020remote,ayush2020geography, lamm2020vehicle}. 

In this paper, we investigate the HPT framework with a diverse set of pretraining procedures and downstream tasks. We test 16 datasets spanning visual domains, such as medical, aerial, driving, and simulated images.
In our empirical study, we observe that HPT shows the following benefits compared to self-supervised pretraining from scratch:
\begin{itemize}
    \item HPT reduces self-supervised pretraining convergence time up to 80$\times$.
    \item HPT consistently converges to better performing representations than generalist or specialist pretraining for 15 of the 16 studied datasets on image classification, object detection, and semantic segmentation tasks.
    \item HPT is significantly more resilient to the set of image augmentations and amount of data used during self-supervised pretraining.
\end{itemize}

In the following sections, we discuss the relevant background for our investigation, formalize our experimental settings, present the results and ablations, and include a discussion of the results and their implications and impact on future work. 
Based on the presented analyses, we provide a set of guidelines for practitioners to successfully apply self-supervised pretraining to new datasets and downstream applications. 
Finally, in the appendix, we provide many additional experiments that generalize our results to include supervised pretraining models. In summary, across datasets, metrics, and methods, \emph{self-supervised pretraining improves self-supervised pretraining}.

\section{Background and Related Work}
\textbf{Transfer learning} studies how a larger, more general, or more specialized \emph{source} dataset can be leveraged to
improve performance on \emph{target} downstream datasets/tasks~\cite{raina2007self, quattoni2008transfer, bengio2012deep, devlin2018bert, henaff2019data, he2019momentum,  donahue2014decaf, erhan2010does, zeiler2014visualizing, goodfellow2016deep, lecun2015deep, radford2018improving}. 
This paper focuses on a common type of transfer learning in which model weights trained on source data are used to initialize training on the target task~\cite{yosinski2014transferable}. 
Model performance generally scales with source dataset size and the similarity between the source and target data \cite{raghu2019transfusion, neumann2019domain,niu2020distant}.

A fundamental challenge for transfer learning is to improve the performance on target data when it is not similar to source data. 
Many papers have tried to increase performance when the target and source datasets are not similar.
Recently, \cite{puigcerver2020scalable} proposed first training on the base dataset and then training with subsets of the base dataset to create specialist models, and finally using the target data to select the best specialist model. 
Similarly, \cite{ngiam2018domain} used target data to reweight the importance of base data. 
Unlike these works, we do not revisit the base data, modify the pretrained architecture, or require expert model selection or a reweighting strategy.

\textbf{Self-supervised pretraining} is a form of unsupervised training that captures the intrinsic patterns and properties of the data without using human-provided labels to learn discriminative representations for the downstream tasks~\cite{doersch2015unsupervised,doersch2017multi,zhang2016colorful,gidaris2018unsupervised, wang2020pre}. 
In this work we focus on a type of self-supervised pretraining called \emph{instance contrastive learning}~\cite{dosovitskiy2015discriminative,wu2018unsupervised,he2019momentum}, which trains a network by determining which visually augmented images originated from the same image, when contrasted with augmented images originating from different images. 
Instance contrastive learning has recently outperformed supervised pretraining on a variety of transfer tasks \cite{he2019momentum, chen2020big}, which has lead to increased adoption in many applications.
 Specifically, we use the MoCo algorithm~\cite{chen_improved_2020} due to its popularity, available code base, reproducible results without multi-TPU core systems, and similarity to other self-supervised algorithms \cite{le2020contrastive}.  We also explore additional self-supervised methods in the appendix. 
 \par
Our focus is on self-supervised learning for vision tasks.
Progressive self-supervised pretraining on multiple datasets has also been explored for NLP tasks, e.g.~ see \cite{gururangan2020don, pfeiffer2020adapterhub} and the citations within. 
In \cite{gururangan2020don}, the authors compare NLP generalist models and NLP models trained on additional source and task-specific data. While our work is similar in spirit to the language work of \cite{gururangan2020don}, our work focuses on computer vision, includes a greater variation of pretraining pipelines and methods, and allows for adaptation with fewer parameter updates.

\textbf{Label-efficient learning} includes weak supervision methods \cite{mahajan2018exploring}
that assume access to imperfect but related labels, and semi-supervised methods that assume labels are only available for a subset of available examples~\cite{chen2020big, kolesnikov2020big, yalniz2019billion}. 
While some of the evaluations of the learned representations are done in a semi-supervised manner, HPT is complementary to these approaches and the representations learned from HPT can be used in conjunction with them.

\section{Hierarchical pretraining}
HPT
sequentially performs a small amount of self-supervised pretraining on data that is increasingly similar to the target dataset. In this section, we formalize each of the HPT components as depicted in Figure~\ref{fig:hpt-pipeline}.

\textbf{Base pretraining:} 
We use the term \emph{base pretraining} to describe the initial pretraining step where a large, general vision dataset (\emph{base dataset}) is used to pretrain a model from scratch. Practically, few users will need to perform base pretraining, and instead, can use publicly available pretrained models, such as ImageNet models. Because base pretraining, like many prior transfer learning approaches, is domain agnostic, most practitioners will select the highest performing model on a task with a large domain~\cite{kolesnikov2019big}. 

\textbf{Source pretraining:}
Given a base trained model, we select a source dataset that is both larger than the target dataset and more similar to the target dataset than the base dataset. 
Many existing works have explored techniques to select a model or dataset that is ideal for transfer learning with a target task~\cite{renggli2020model}.
Here, we adopt an approach studied by \cite{kornblith2019better, renggli2020model} in a supervised context called a \emph{task-aware search strategy}: each potential source dataset is used to perform self-supervised pretraining on top of the base model for a very short amount of pretraining, e.g. $\sim$5k pretraining steps as discussed in Section~\ref{sec:exp}. 
The supervised target data is then used to train a linear evaluator on the frozen pretrained source model. 
The source model is then taken to be the model that produces the highest linear evaluation score on the target data, and is then used for additional target pretraining.

Experimentally, we have found that using a single, similar, and relatively large (e.g.~$> 30\mathrm{K}$ images) source dataset consistently improves representations for the target task. 
Furthermore, we view source pretraining as an optional step, and as shown in Section~\ref{sec:exp}, HPT still leads to improved results when directly performing self-supervised pretraining on the target dataset following the base pretraining.
We further discuss source model selection in the appendix.

\textbf{Target pretraining:} 
Finally, we perform self-supervised pretraining with the target dataset, initialized with the final source model, or the base model in the case when no source model was used. This is also the stage where layers of the model can be frozen to prevent overfitting to the target data and enable faster convergence speed.
Experimentally, we have found that freezing all parameters except the modulation parameters of the batch norm layers leads to consistently strong performance for downstream tasks, particularly when the target dataset is relatively small ($< 10\mathrm{K}$ images).

\textbf{Supervised finetune:}
Given the self-supervised pretrained model on the target dataset, we transfer the final model to the downstream target task, e.g. image classification or object detection.

\begin{figure*}[tbh] 
    \centering
    \includegraphics[width=\linewidth]{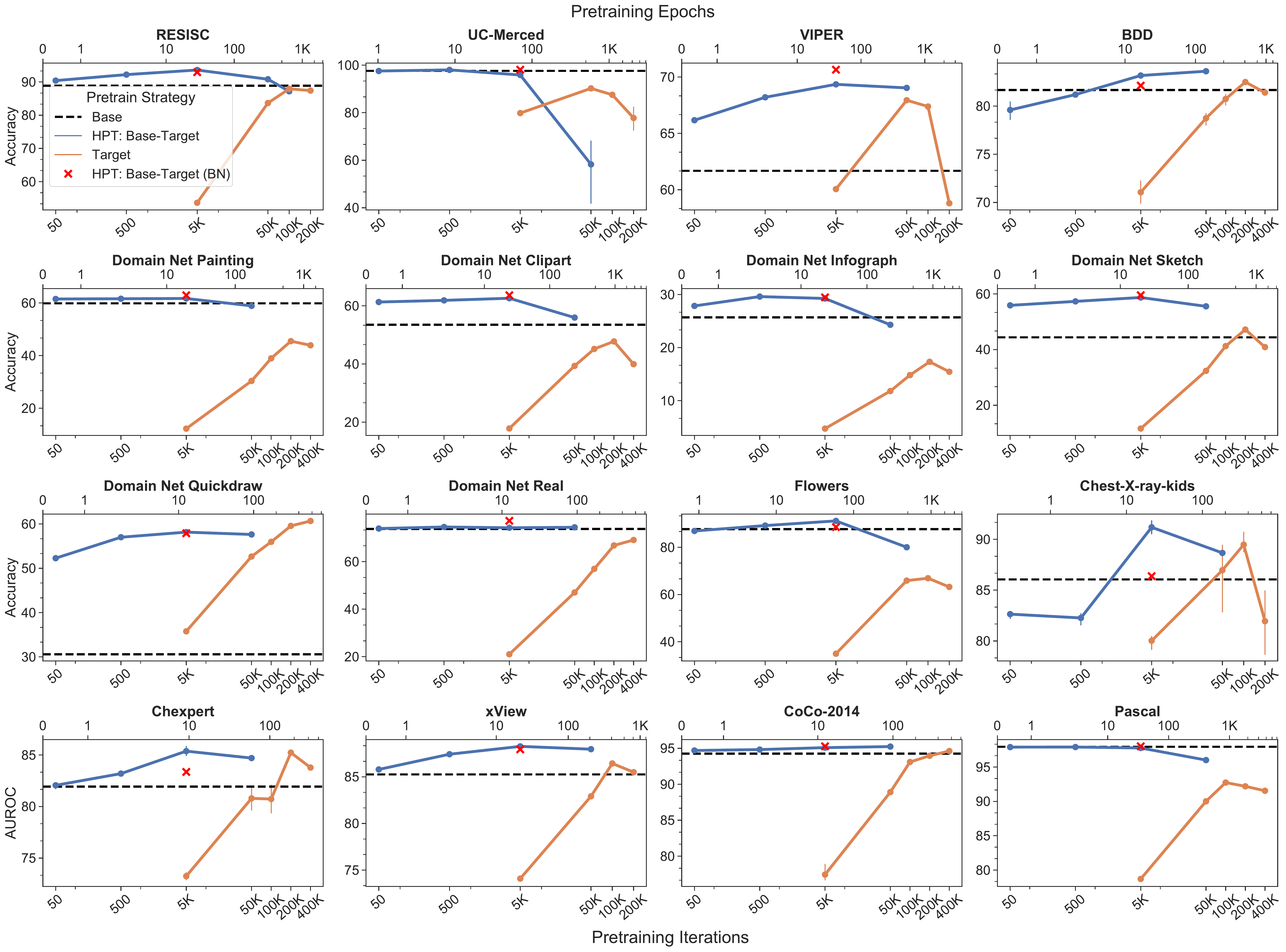}
     \caption{\emph{Linear separability evaluation}. For each of the 16 datasets, we train a generalist model for 800 epochs on ImageNet (Base). We either train the whole model from 50-50k iters (HPT Base-Target) or just the batch norm parameters for 5k iters (HPT Base-Target (BN)). We compare HPT to a Specialist model trained from a random initialization (Target).
     For each, we train a linear layer on top of the final representation. HPT obtains the best results on 15 out of 16 datasets without hyperparameter tuning.}
     \label{fig:hpt-exp1-linear-analysis}
\end{figure*}
\section{Experiments}
\label{sec:exp}

Through the following experiments, we investigate the quality, convergence, and robustness of self-supervised pretraining using the HPT framework. 

\subsection{Datasets}
We explored self-supervised pretraining on the following datasets that span several visual domains (see the appendix for all details). Dataset splits are listed with a train/val/test format in square brackets after the dataset description.

\textit{\textbf{Aerial}}: \textbf{xView}~\cite{lam2018xview} is a 36-class object-centric, multi-label aerial imagery dataset [$39133/2886/2886$]. \textbf{RESISC}~\cite{cheng2017remote} is a 45-class scene classification dataset for remote sensing [$18900/6300/6300$]. \textbf{UC-Merced}~\cite{yang2010bag} is a 21-class aerial imagery dataset [$1260/420/420$].

\textit{\textbf{Autonomous Driving}}:
\textbf{BDD}~\cite{yu2020bdd100k} is a high resolution driving dataset with $10$ object detection labels and $6$ weather classification labels. We evaluate HPT performance over the object detection task, as well as the weather classification task [$60k/10k/10k$].
\textbf{VIPER}~\cite{richter2017playing} is a 23-class simulated driving dataset for which we perform multi-label each object in the image [$13367/2868/4959$].

\textit{\textbf{Medical}}: \textbf{Chexpert}~\cite{irvin2019chexpert} is a large, multi-label X-ray dataset, where we determine whether each image has any of 5 conditions [$178731/44683/234$]. \textbf{Chest-X-ray-kids}~\cite{kermany2018large} provides pediatric X-rays used for 4-way pneumonia classification [$4186/1046/624$]. 

\textit{\textbf{Natural, Multi-object}}: \textbf{COCO-2014}~\cite{lin2014microsoft} is an 81-class object detection benchmark. 
We perform multi-label classification for each object, and we further use the $2017$ split to perform object detection and segmentation [$82783/20252/20252$]. \textbf{Pascal} VOC 2007+2012~\cite{everingham2015pascal} is a standard 21-class object detection benchmark we use for multi-label classification to predict whether each object is in each image. We also use the object detection labels for an object detection transfer task [$13.2k/3.3k/4.9k$].

\textit{\textbf{Assorted}}: \textbf{DomainNet}~\cite{peng2019moment} contains six distinct datasets, where each contains the same 345 categories. The domains consist of \texttt{real} images similar to ImageNet, \texttt{sketch} images of greyscale sketches, \texttt{painting} images, \texttt{clipart} images, \texttt{quickdraw} images of binary black-and-white drawings from internet users, and \texttt{infograph} illustrations. 
We use the original train/test splits with 20\% of the training data used for validation. \textbf{Oxford Flowers}~\cite{nilsback2008automated}: we use the standard split to classify 102 fine-grain flower categories [$1020/1020/6149$].

\begin{figure*}[t]
\tiny
\centering

\stackunder[5pt]{\includegraphics[width=3.8cm]{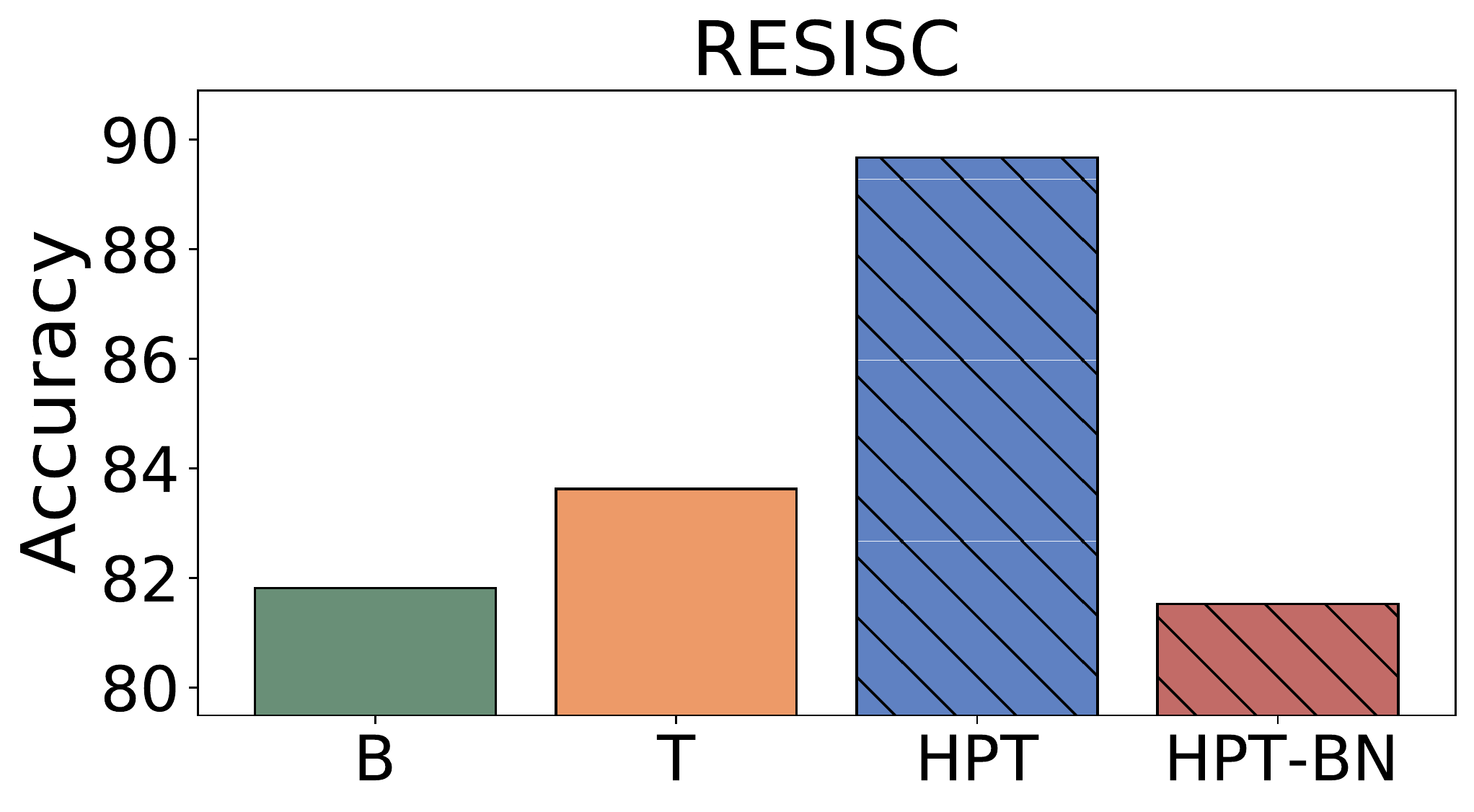}}{} 
\hspace{6pt}%
\stackunder[5pt]{\includegraphics[width=3.8cm]{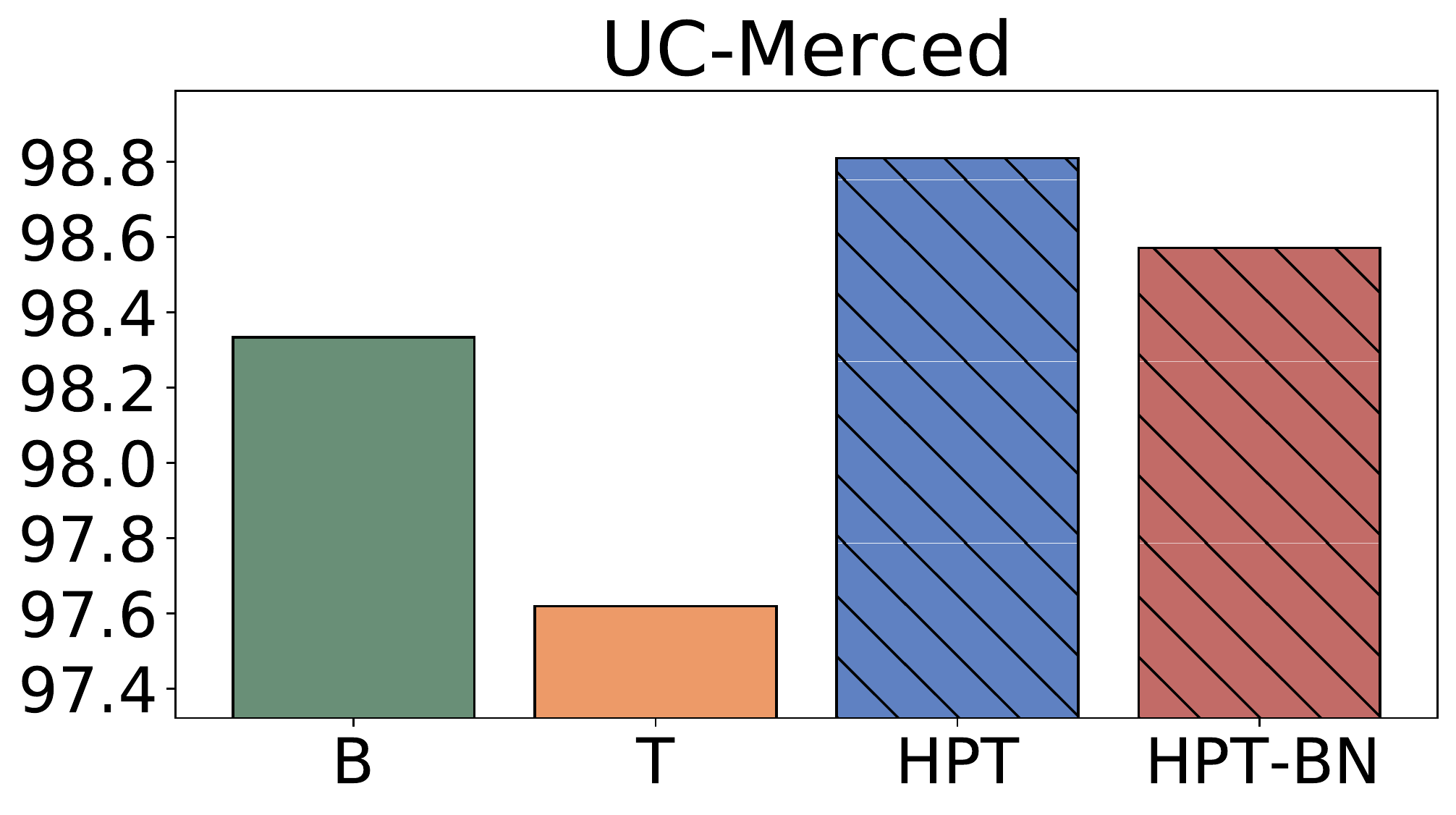}}{}
\hspace{6pt}%
\stackunder[5pt]{\includegraphics[width=3.8cm]{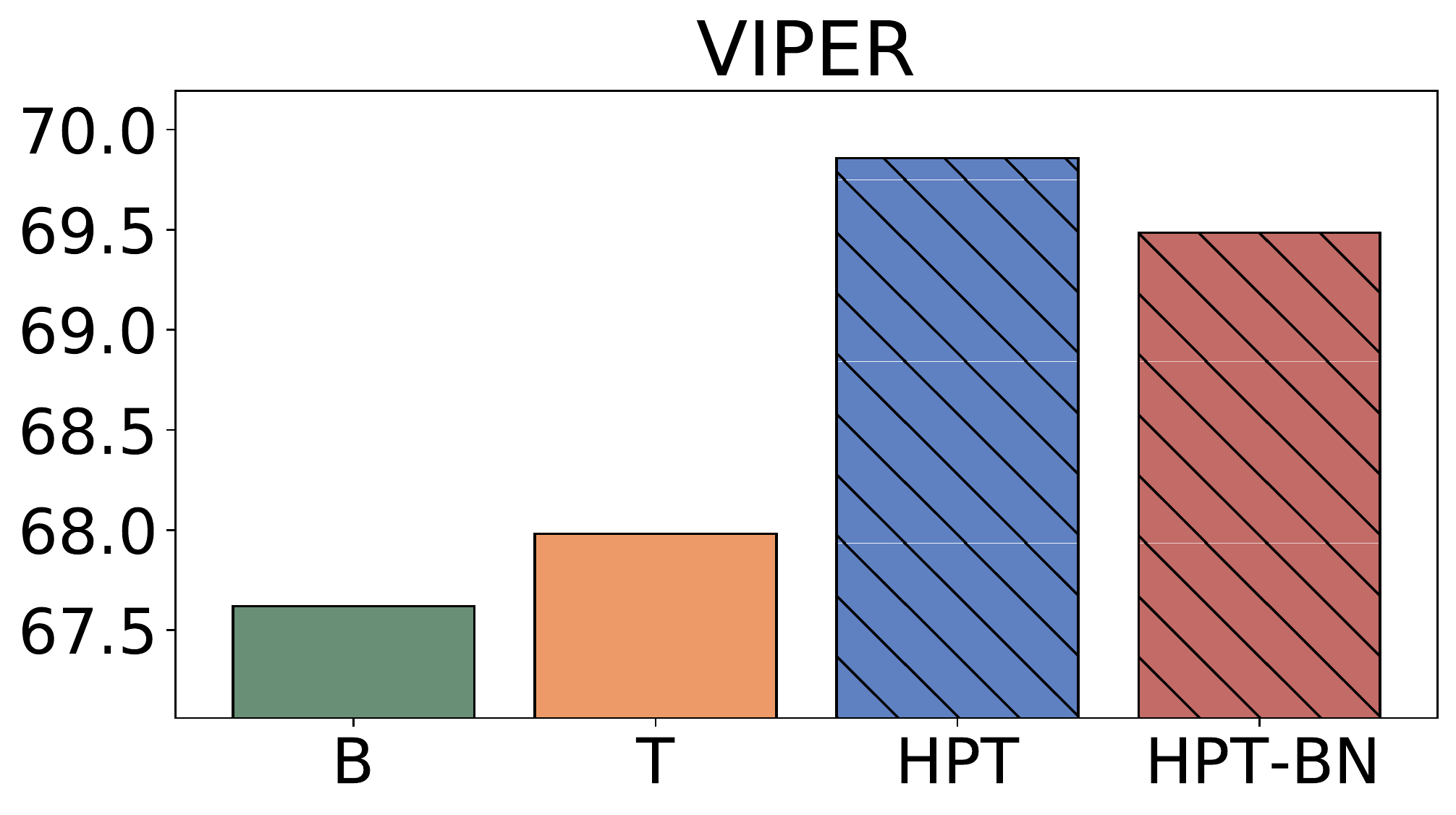}}{} 
\hspace{6pt}%
\stackunder[5pt]{\includegraphics[width=3.8cm]{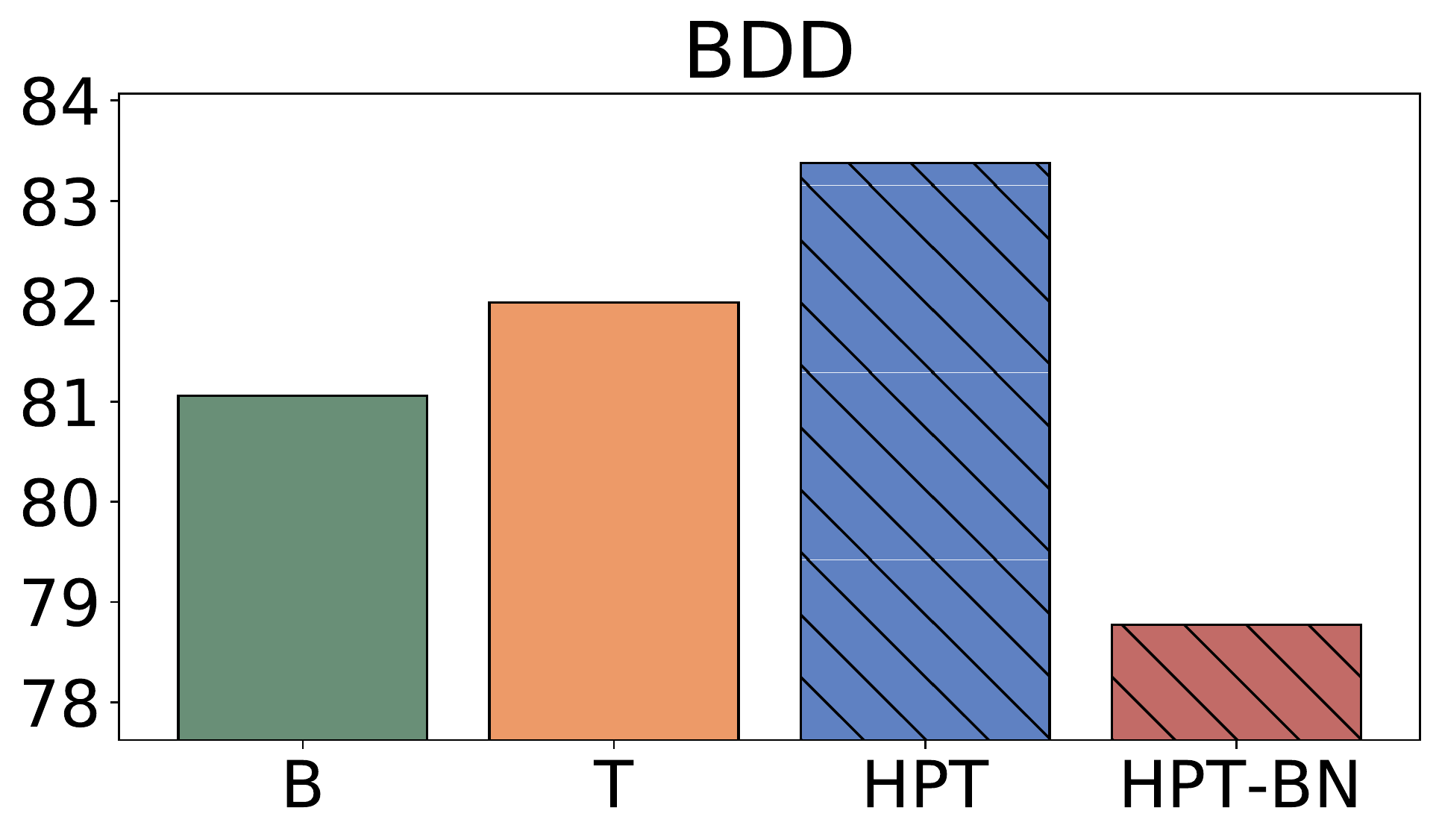}}{}%
\\
\stackunder[5pt]{\includegraphics[width=3.8cm]{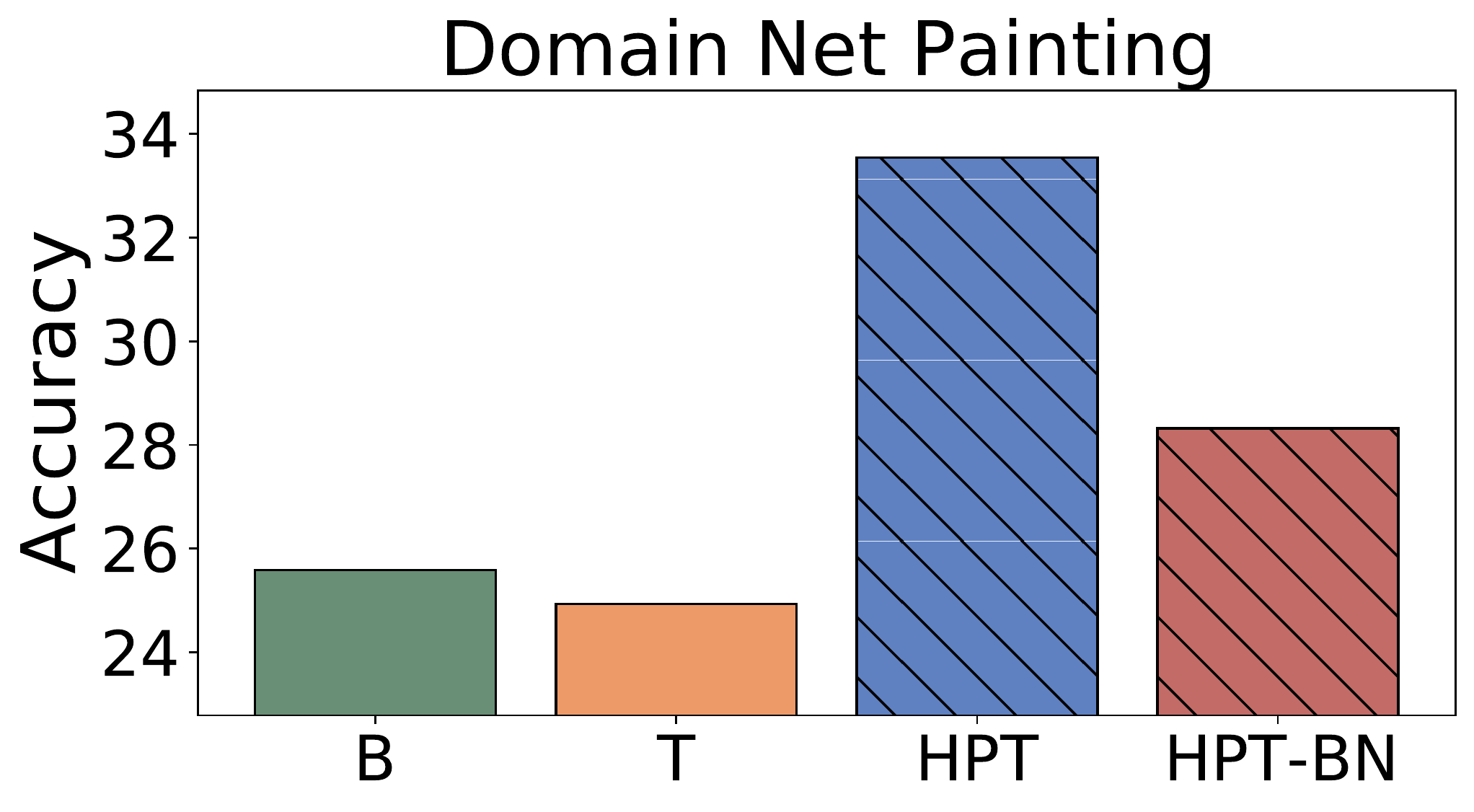}}{}
\hspace{6pt}%
\stackunder[5pt]{\includegraphics[width=3.8cm]{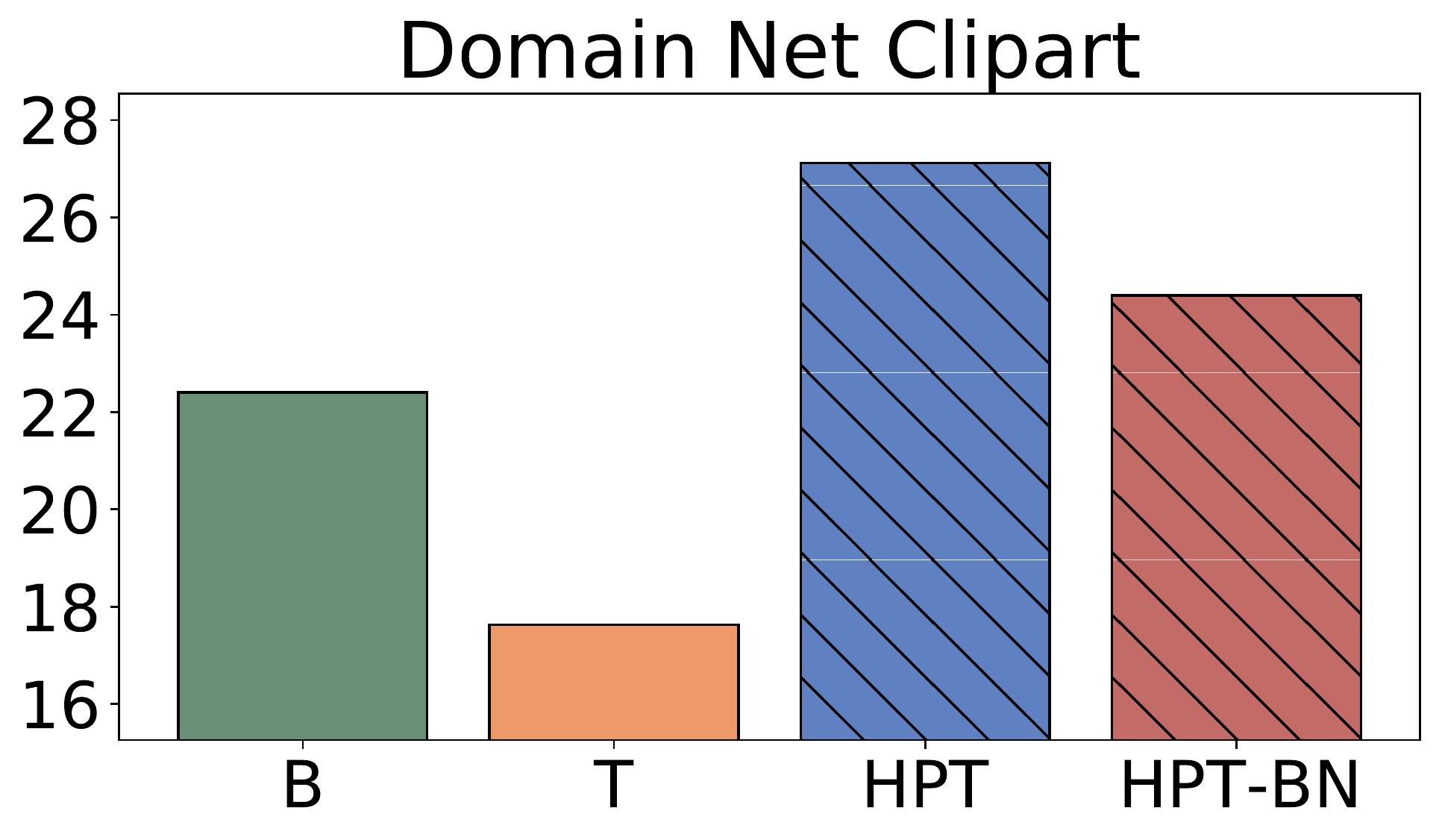}}{}
\hspace{6pt}%
\stackunder[5pt]{\includegraphics[width=3.8cm]{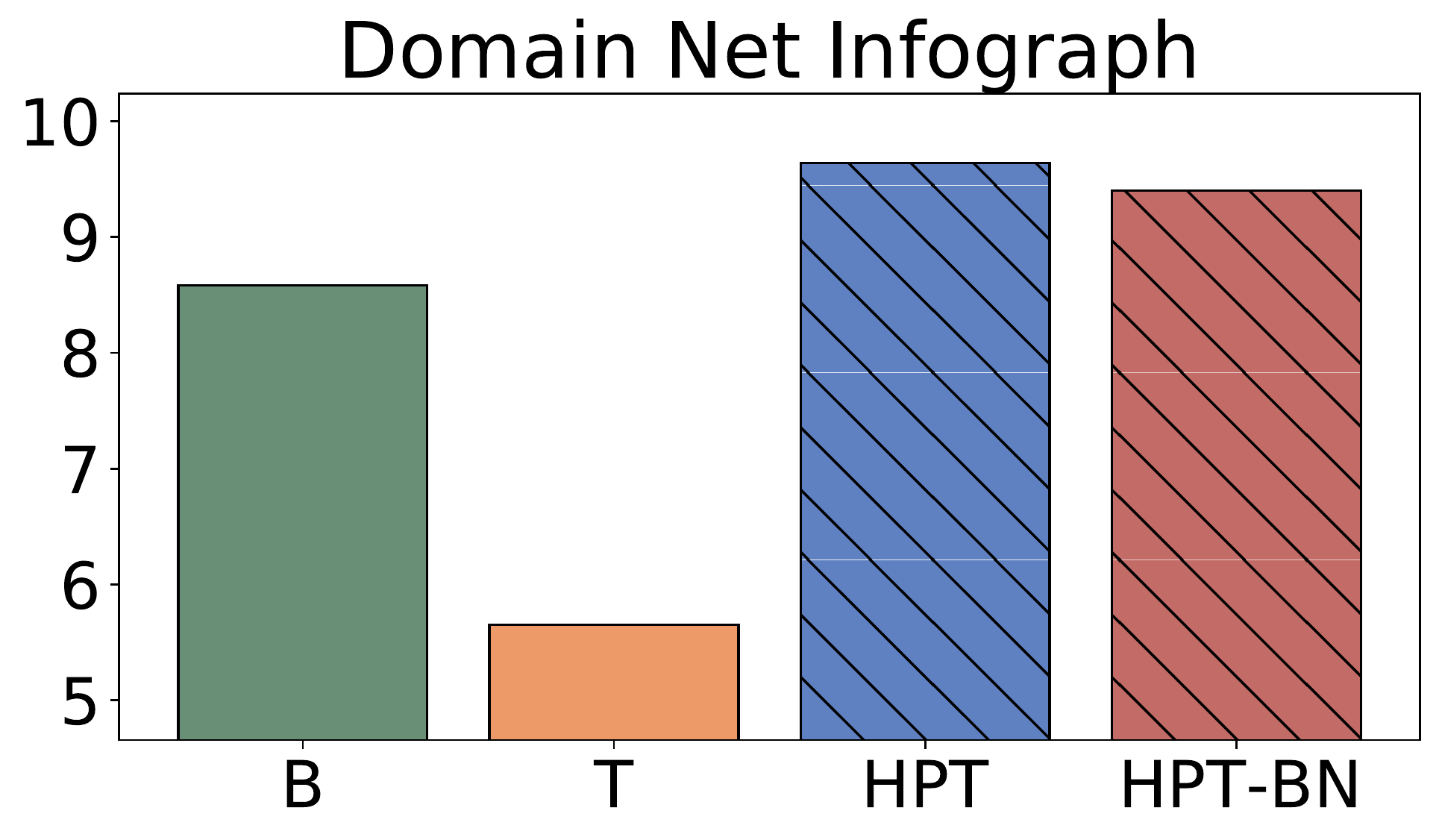}}{}
\hspace{6pt}%
\stackunder[5pt]{\includegraphics[width=3.8cm]{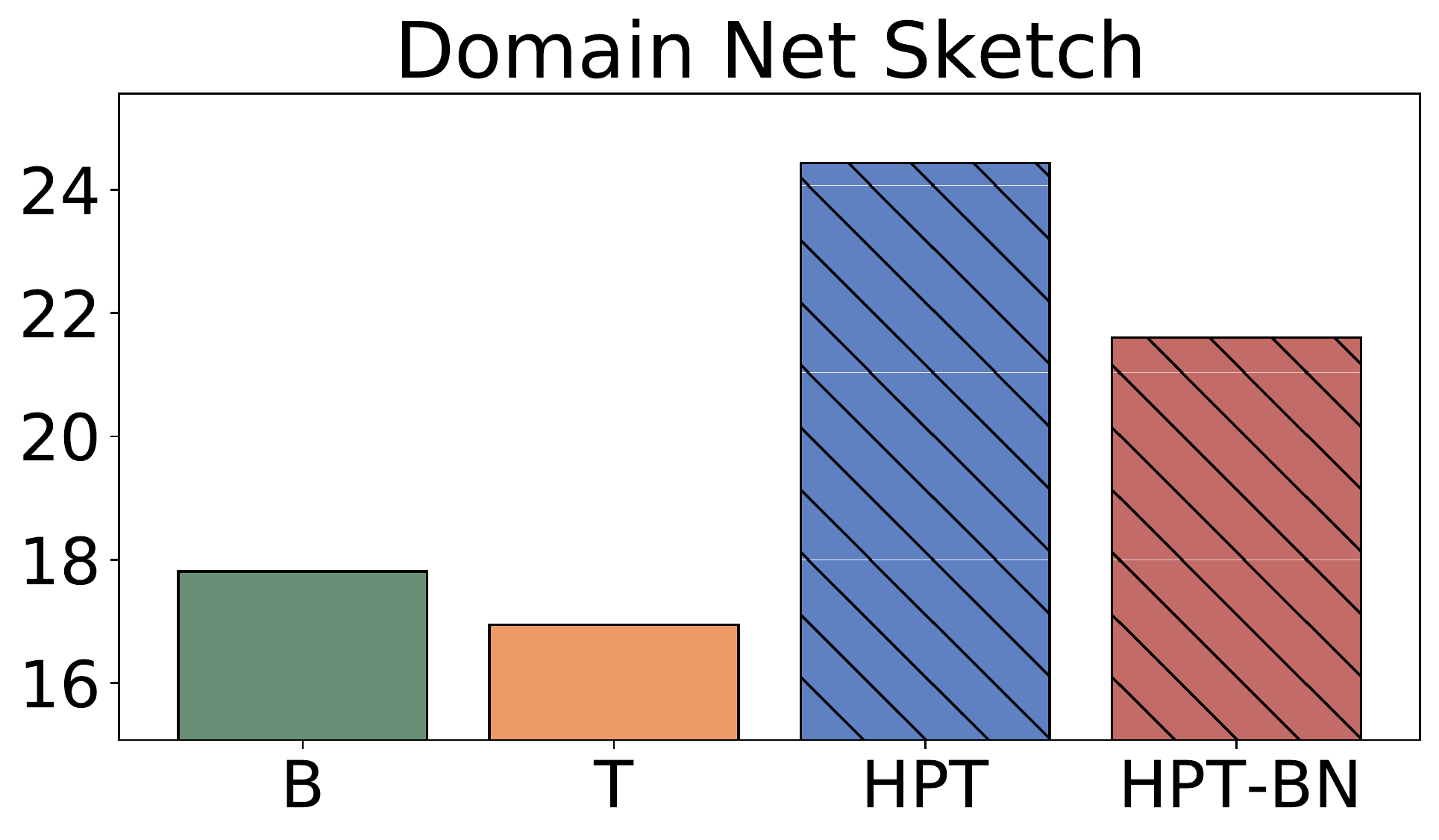}}{}
\\
\stackunder[5pt]{\includegraphics[width=3.8cm]{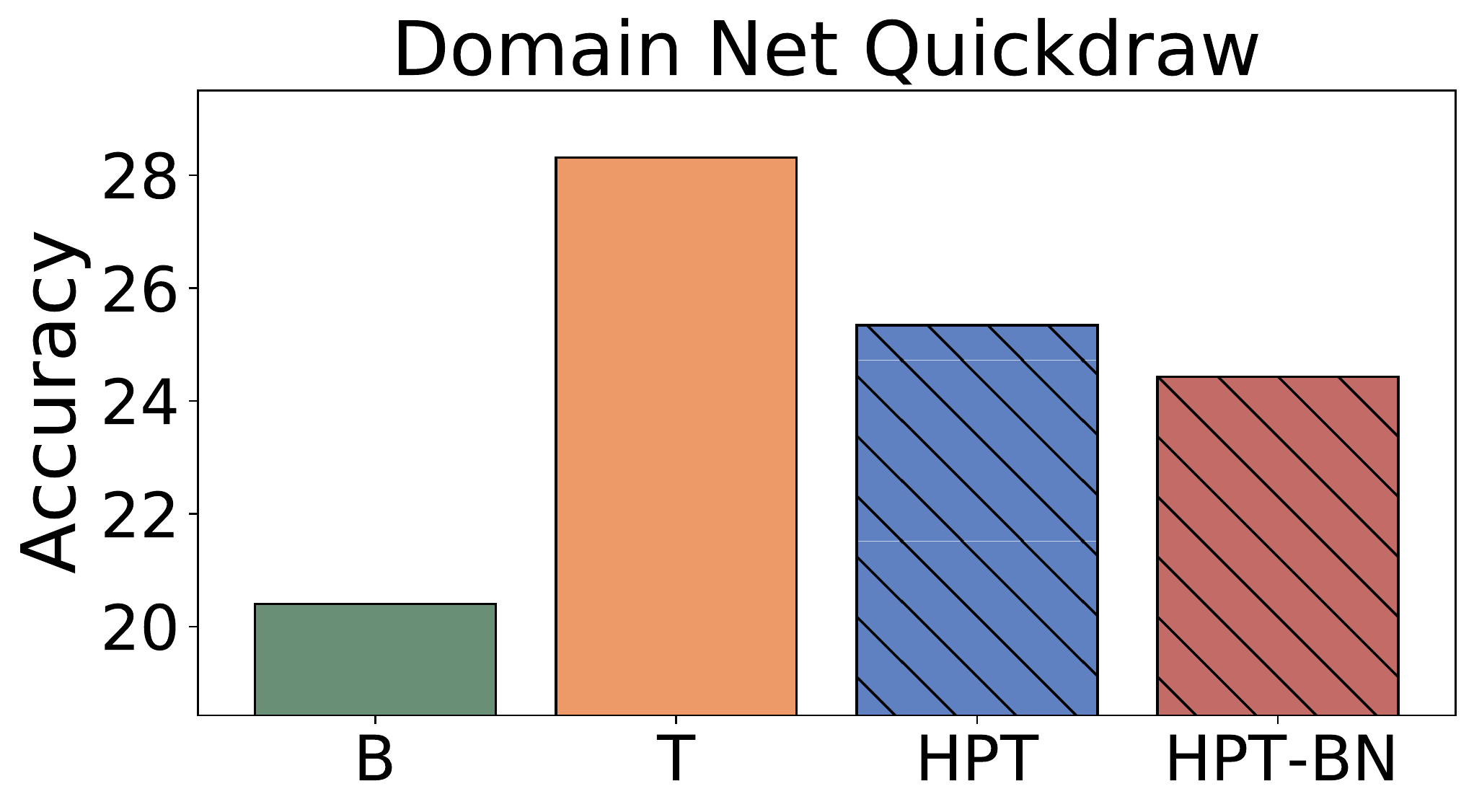}}{} 
\hspace{6pt}%
\stackunder[5pt]{\includegraphics[width=3.8cm]{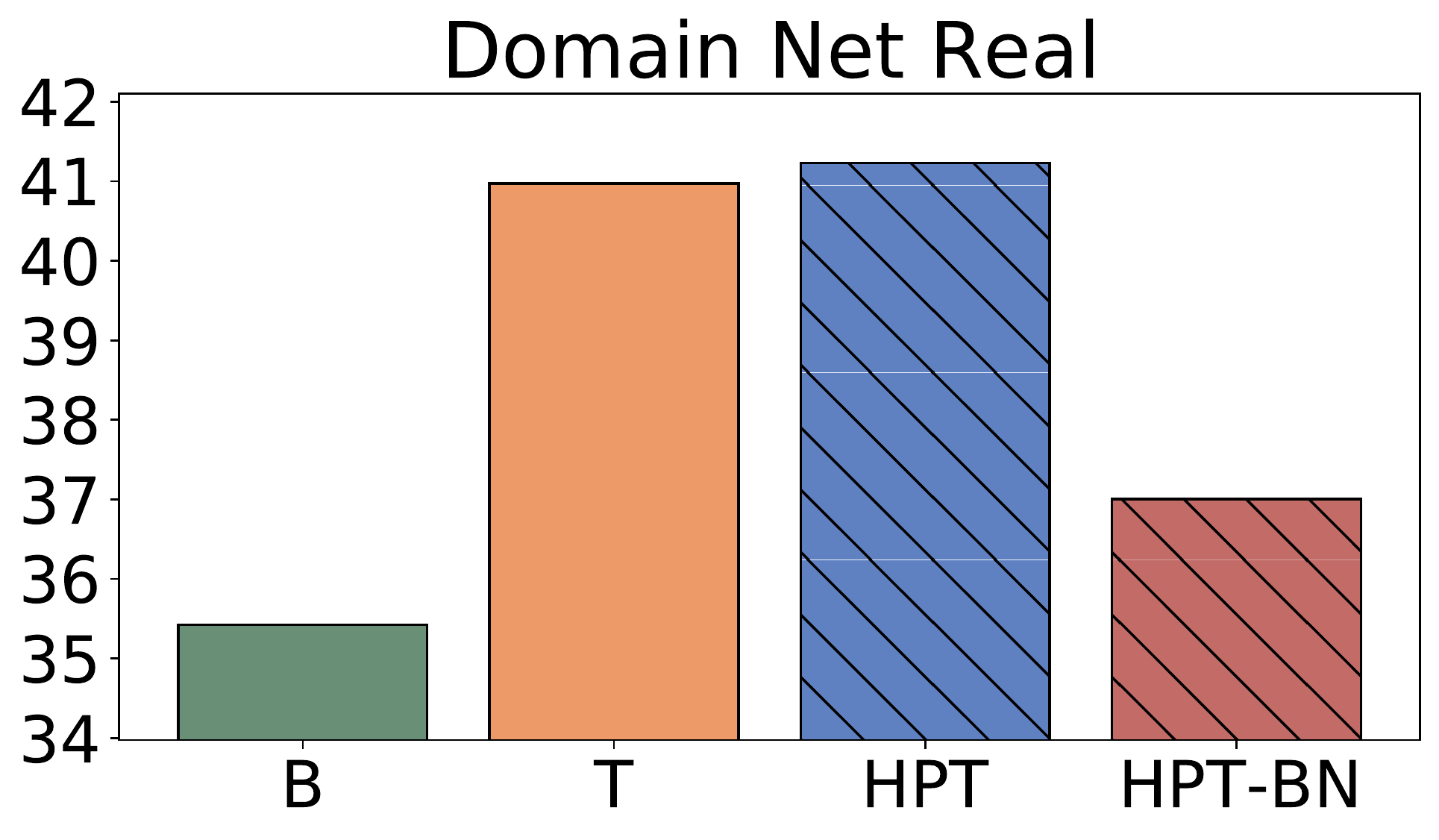}}{}
\hspace{6pt}%
\stackunder[5pt]{\includegraphics[width=3.8cm]{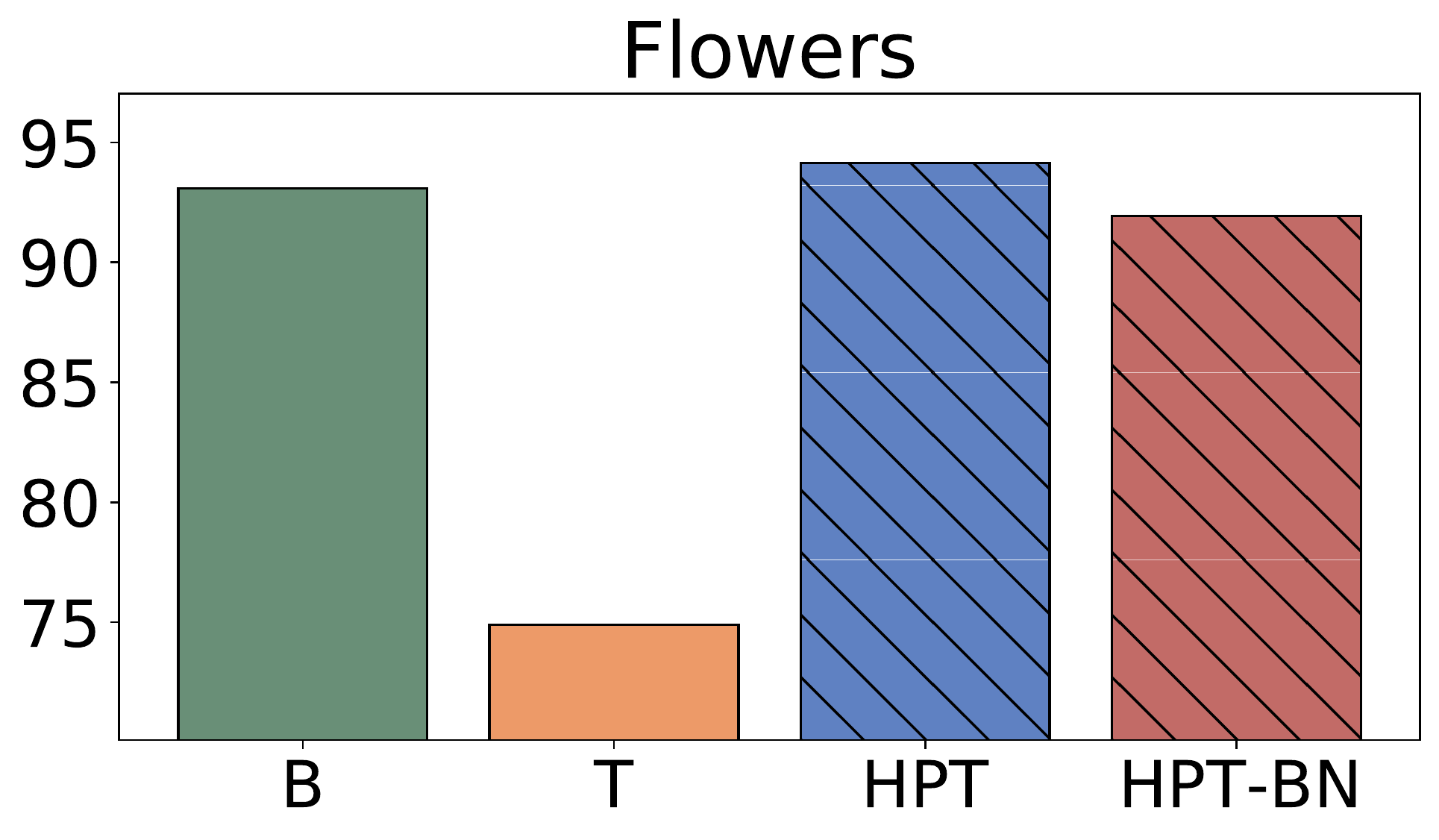}}{} 
\hspace{6pt}%
\stackunder[5pt]{\includegraphics[width=3.8cm]{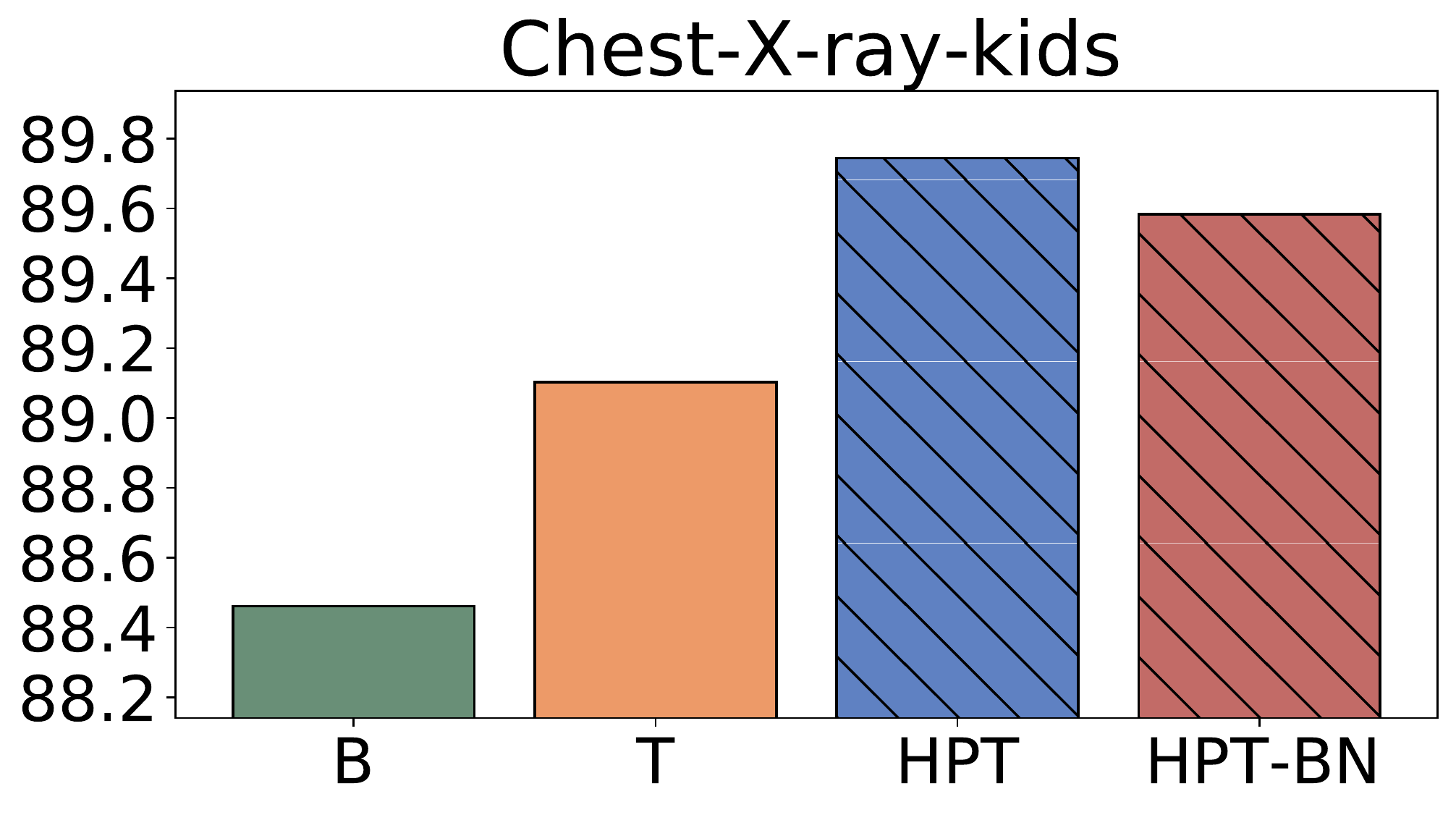}}{}
\\
\stackunder[5pt]{\includegraphics[width=3.8cm]{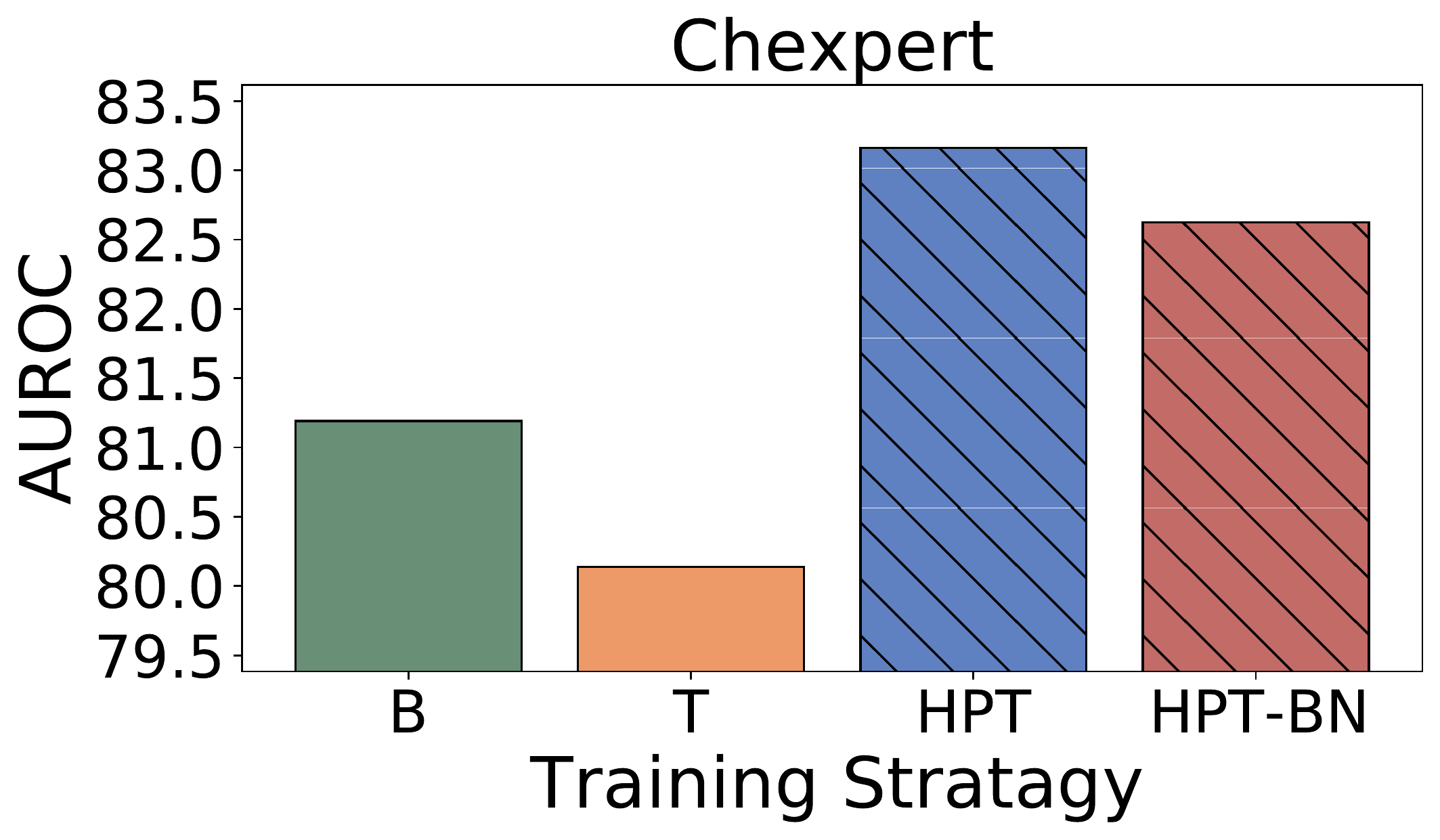}}{} 
\hspace{6pt}%
\stackunder[5pt]{\includegraphics[width=3.8cm]{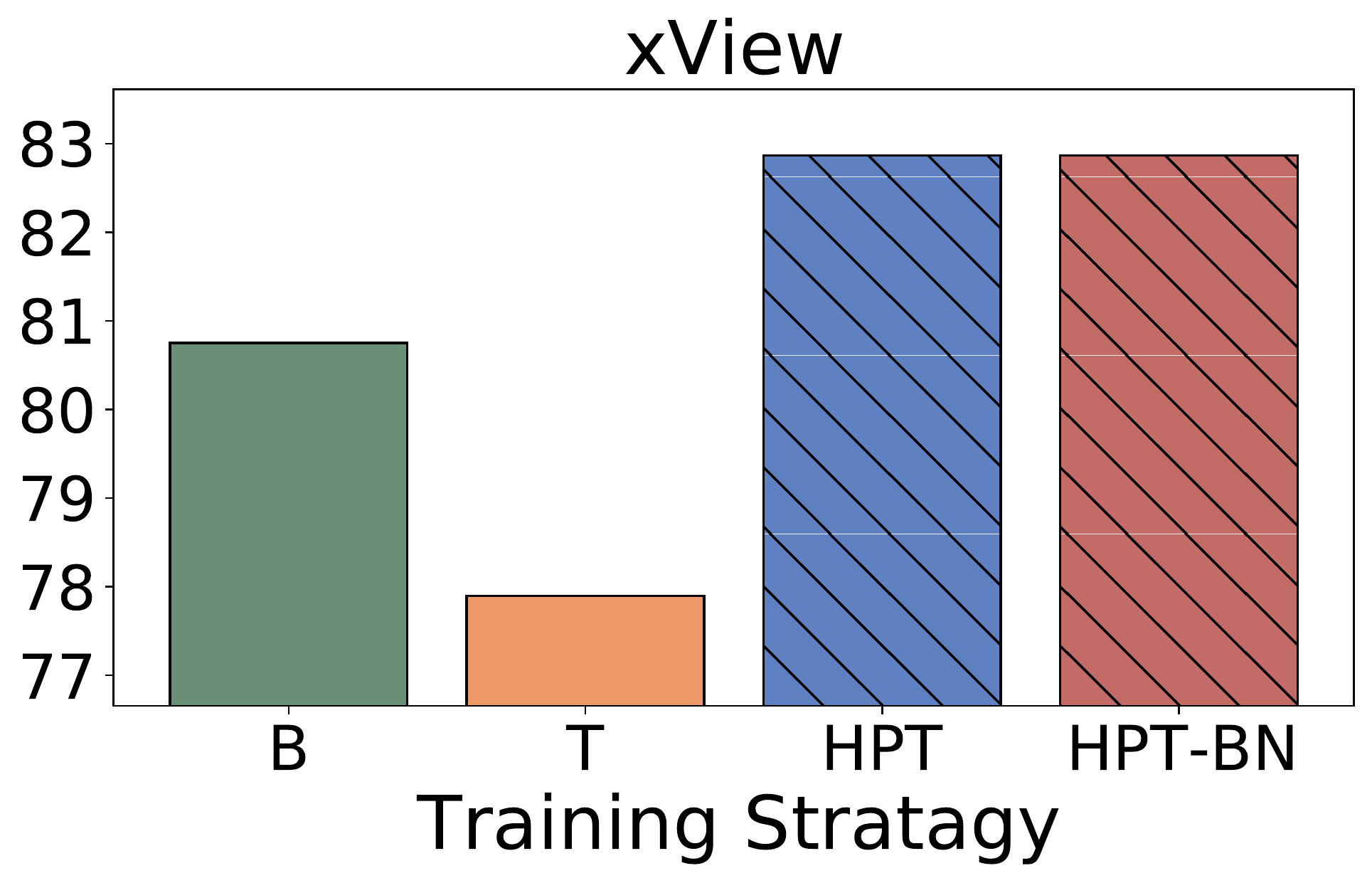}}{}
\hspace{6pt}%
\stackunder[5pt]{\includegraphics[width=3.8cm]{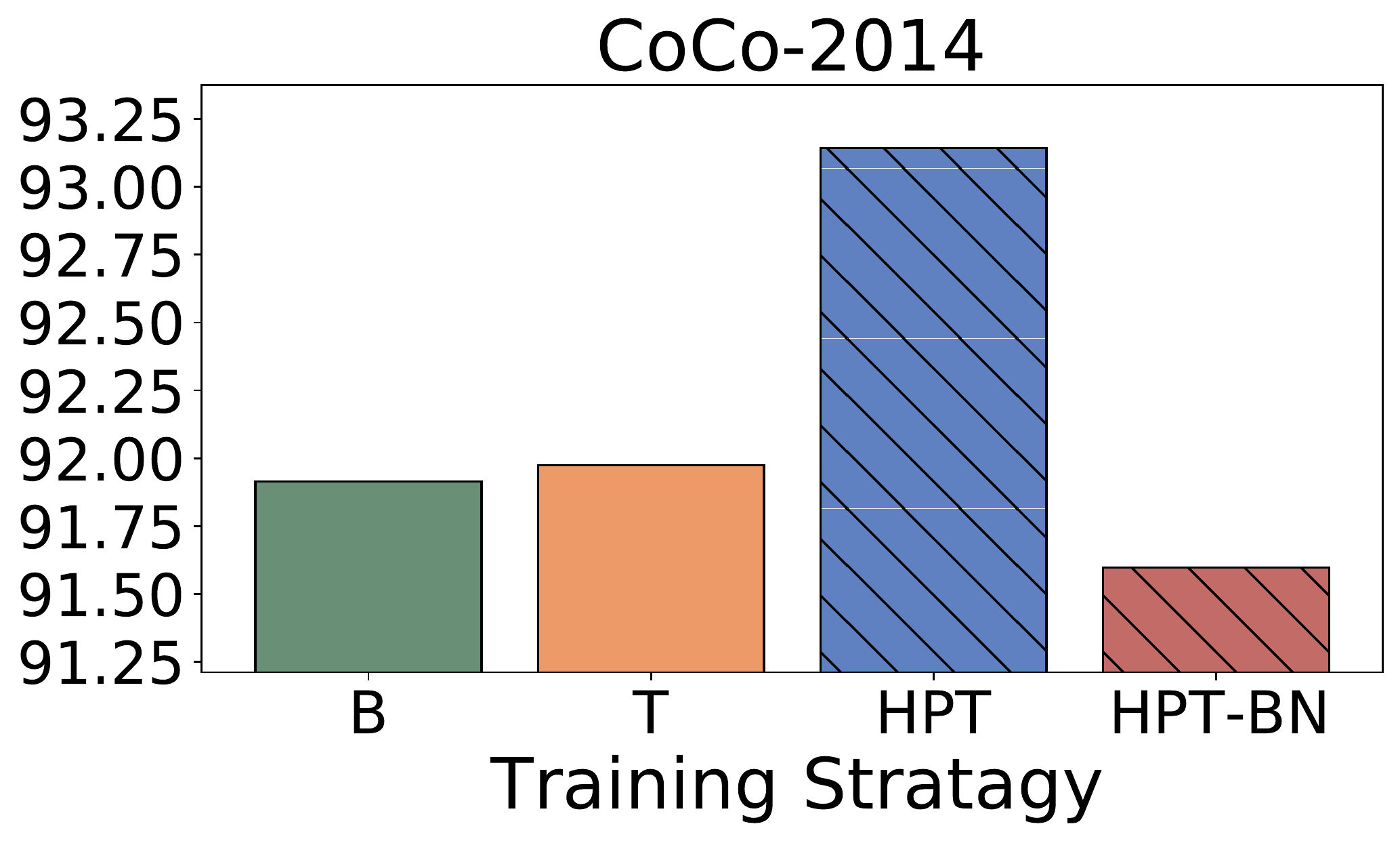}}{} 
\hspace{6pt}%
\stackunder[5pt]{\includegraphics[width=3.8cm]{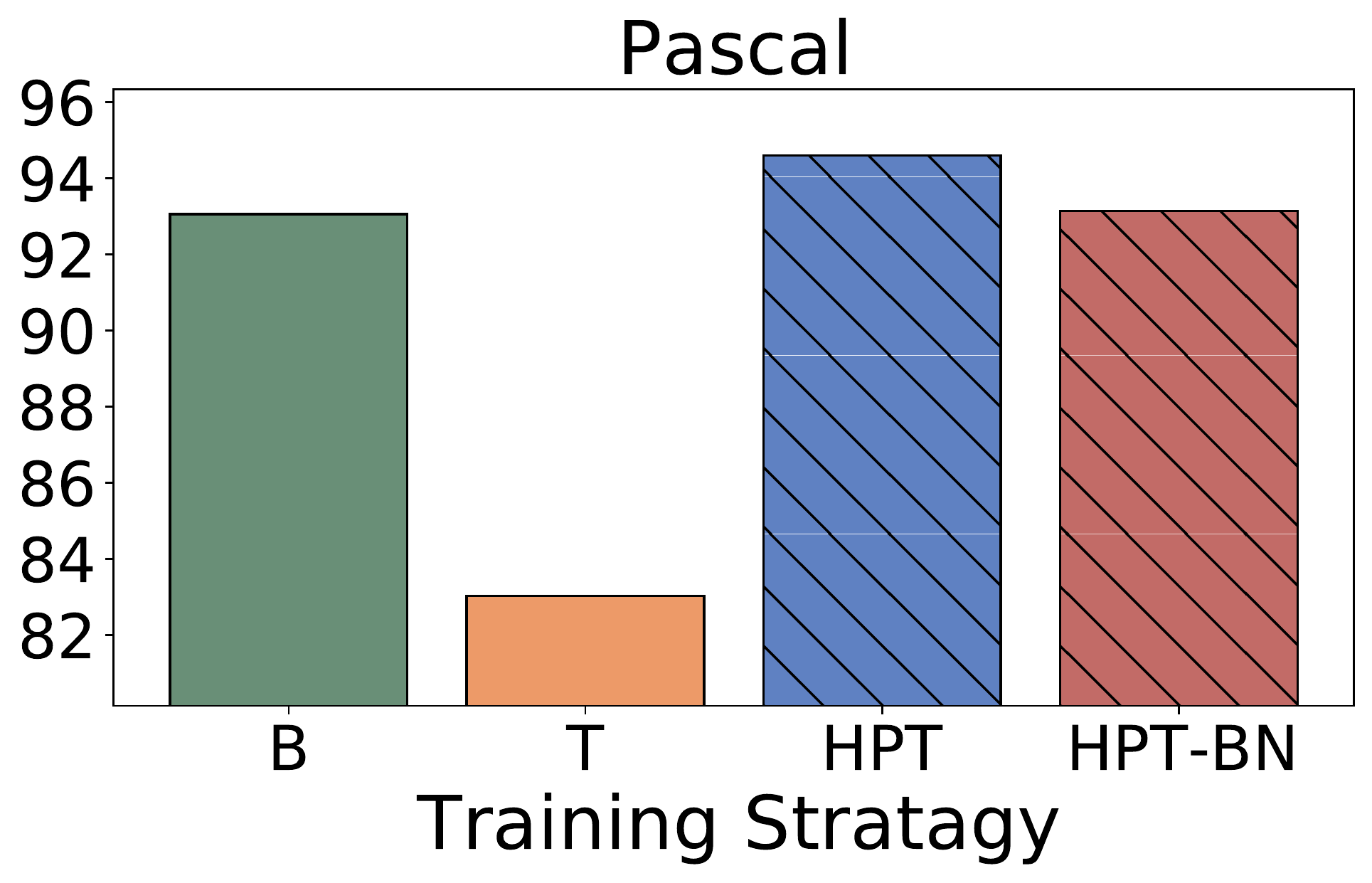}}{} 
\caption{\emph{Semi-supervised evaluation}. We compared the best semi-supervised finetuning performance from the (B)ase model, (T)arget pretrained model, HPT pretrained model, and HPT-BN pretrained model using a 1k labeled subset of each dataset. Despite performing 10x-80x less pretraining, HPT consistently outperformed the Base and Target. HPT-BN generally showed improvement over Base model transfer, but did not surpass HPT's performance.}
\label{fig:ex2}%
\end{figure*}

\subsection{Evaluations}
\label{ss:eval}

The features of self-supervised pretrained models are typically evaluated using one of the following criteria: 
\begin{itemize}
    \item \textbf{Separability}: 
    Tests whether a linear model can distinguish different classes in a dataset using learned features. Good representations should be linearly separable
     \cite{oord2018representation, coates2012learning}. 
    \item \textbf{Transferability}: 
    Tests the performance of the model when finetuned on new datasets and tasks. Better representations will generalize to more downstream datasets tasks \cite{he2019momentum}.
    \item \textbf{Semi-supervised}: 
    Test performance with limited labels. Better representations will suffer less performance degradation \cite{henaff2019data, chen_simple_2020}.
\end{itemize}

We explored these evaluation methods with each of the above datasets. For all evaluations, unless otherwise noted, we used a single, centered crop of the test data with no test-time augmentations. For classification tasks, we used top-1 accuracy and for multi-label classification tasks we used the Area Under the ROC (AUROC)~\cite{bradley1997use}.

In our experiments, we used MoCo-V2~\cite{chen_improved_2020} as the self-supervised training algorithm. We selected MoCo-V2 as it has state-of-the-art or comparable performance for many transfer tasks, and because it uses the InfoNCE loss function~\cite{oord2018representation}, which is at the core of many recent contrastive pretraining algorithms \cite{liu2020self}.
Unless otherwise noted, all training is performed with a standard ResNet-50 backbone~\cite{szegedy2017inception} on 4 GPUs, using default training parameters from \cite{he2019momentum}.
We also explored additional self-supervised pretraining algorithms and hyperparameters in the appendix. 

In the following experiments, we compare implementations of the following self-supervised pretraining strategies:
\noindent{
\begin{itemize}
    \item \textbf{Base}: transfers the 800-epoch MoCo-V2 ImageNet model from \cite{chen_improved_2020} and also updates the batch norm's non-trainable mean and variance parameters using the target dataset (this uniformly led to slightly improved performance for Base transfer).
    \item \textbf{Target}: performs MoCo-V2 on the target dataset from scratch.
    \item \textbf{HPT}: initializes MoCo-V2 pretraining with the 800-epoch MoCo-V2 ImageNet model from \cite{chen_improved_2020}, then optionally performs pretraining on a source dataset before pretraining on the target dataset. The batch norm variant (\textbf{HPT-BN}) only trains the batch norm parameters ($\gamma,\beta$), e.g.~a ResNet-50 has 25.6M parameters, where only ${\sim}0.2\%$ are BN parameters.
\end{itemize}
}
Existing work largely relies on supervised evaluations to tune the pretraining hyperparameters~\cite{chen_simple_2020}, but in practice, it is not possible to use supervised evaluations of unlabeled data to tune the hyperparameters. Therefore, to emphasize the practicality of HPT, we used the default pretraining hyperparameters from \cite{chen_improved_2020} with a batch size of 256 (see the appendix for full details). %

\subsection{Pretraining Quality Analysis}
\begin{figure*}[t]
\captionsetup[subfigure]{labelformat=empty}
\centering
\stackunder{\includegraphics[width=5.1cm]{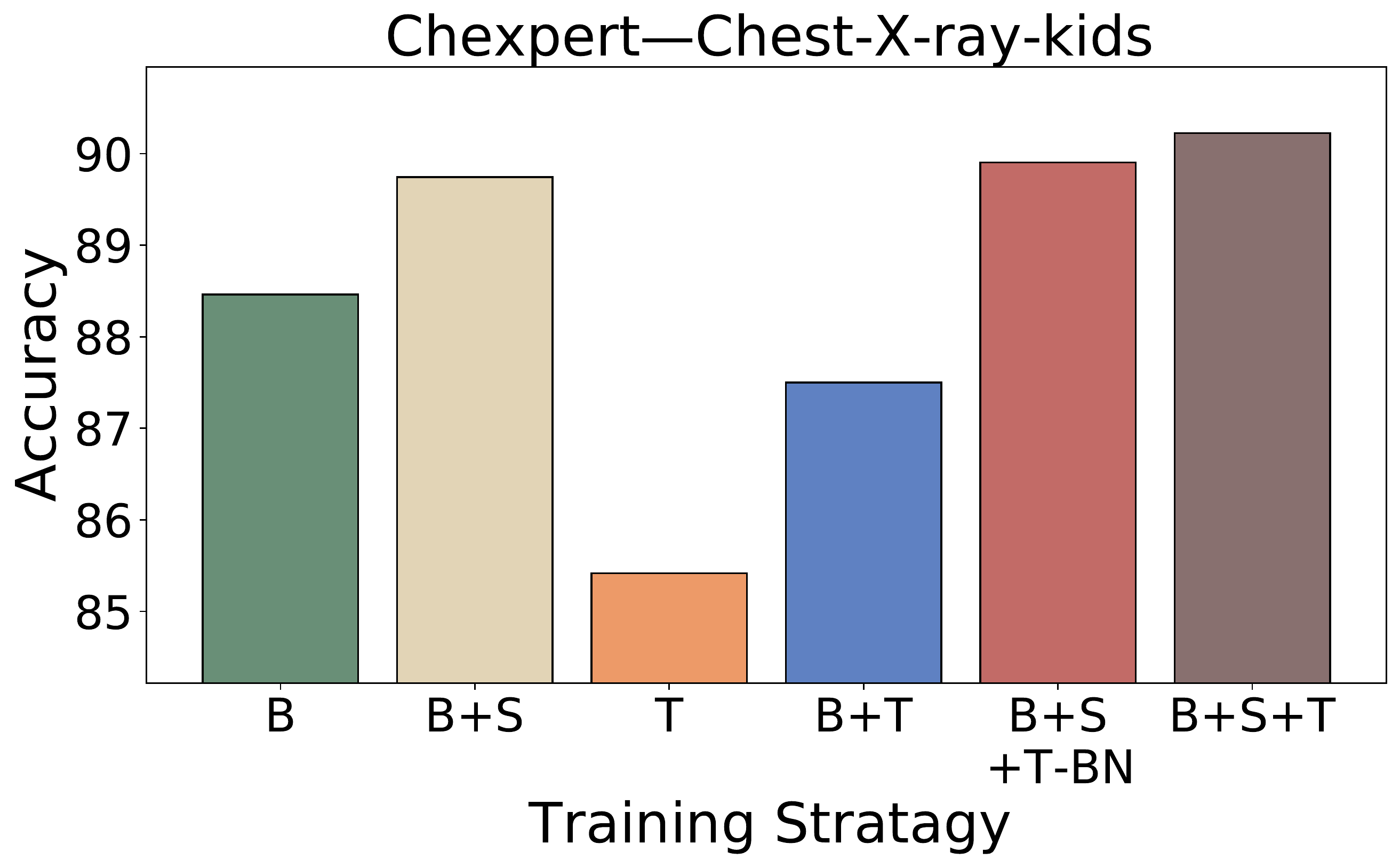}}{} 
\hspace{8pt}%
\stackunder{\includegraphics[width=5.1cm]{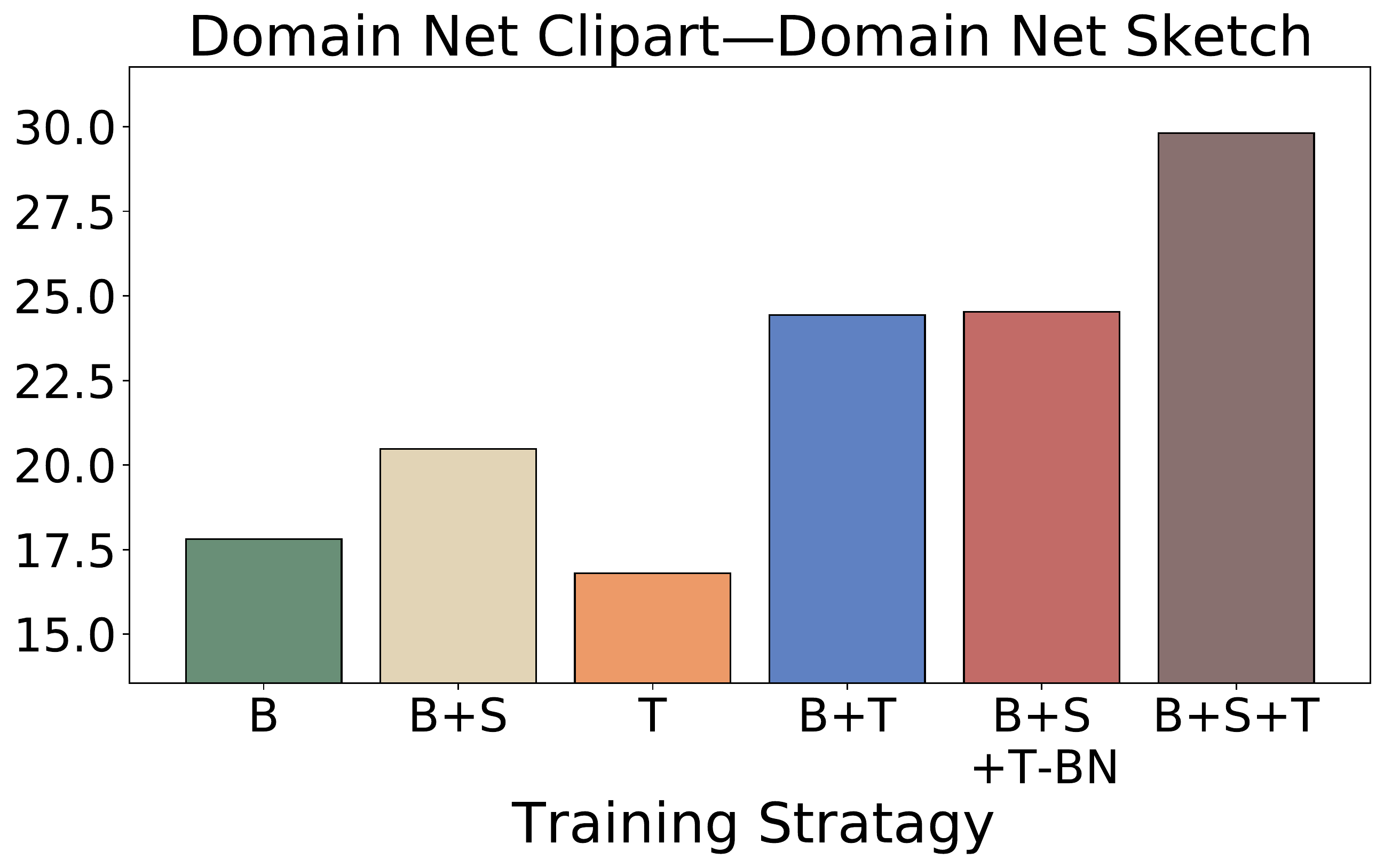}}{} 
\hspace{8pt}%
\stackunder{\includegraphics[width=5.1cm]{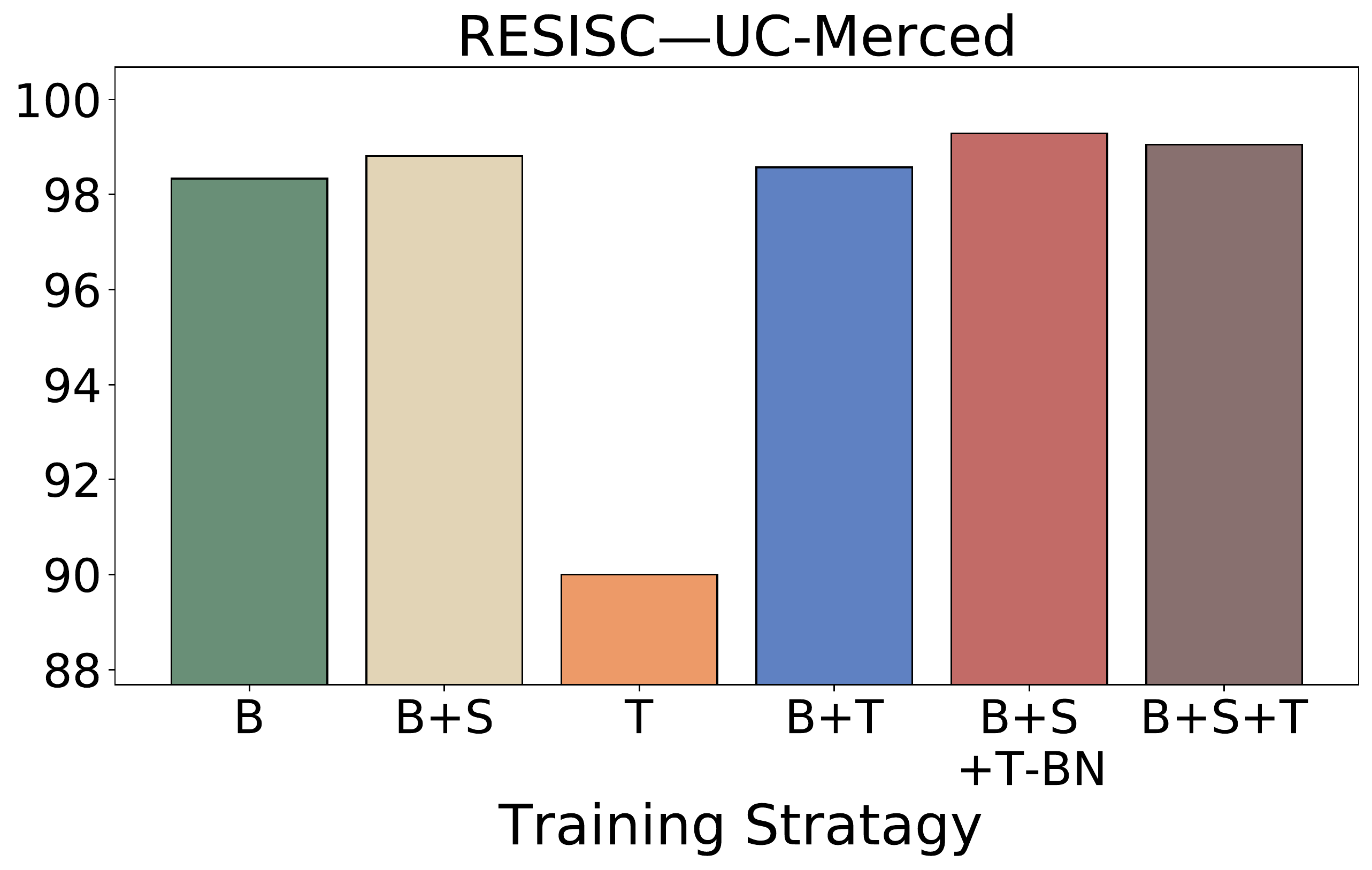}}{}
\caption{\emph{Full finetuning evaluations.} Finetuning performance on target datasets. For these datasets, we evaluated the performance increase on the target dataset by pretraining on sequences of (B)ase (ImageNet), (S)ource (left) dataset, and (T)arget (right) dataset. 
All HPT variants beat all baselines in all cases, with HPT-BN getting slightly better performance on UC Merced and B+S+T having the best performance elsewhere.}
\label{fig:ex3}%
\end{figure*}

\textbf{Separability analysis:}  We first analyzed the quality of the learned representations through a linear separability evaluation~\cite{chen_simple_2020}. We trained the linear model with a batch size of 512 and the highest performing learning rate of $\{0.3,3,30\}$. 
Similar to \cite{kornblith2019better}, we used steps rather than epochs to allow for direct computational comparison across datasets. 
For Target pretraining, we pretrained for \{5k, 50k, 100k, 200k, 400k\} steps, where we only performed 400k steps if there was an improvement between 100k and 200k steps. For reference, one NVIDIA P100 GPU-Day is 25k steps. 
We pretrained HPT for much shorter schedules of \{50, 500, 5k, 50k\} steps, and HPT-BN for 5k steps -- we observed little change in performance for HPT-BN after 5k steps.
 
\textbf{Key observations:} From Figure~\ref{fig:hpt-exp1-linear-analysis}, we observe that HPT typically converges by 5k steps of pretraining \emph{regardless of the target dataset size}, and that for 15 out of 16 datasets, HPT and HPT-BN converged to models that performed as well or better than the Base transfer or Target pretraining at 400k steps (80x longer). 
The only dataset in which the Target pretraining outperformed HPT was \texttt{quickdraw} -- a large, binary image dataset of crowd-sourced drawings.
We note that \texttt{quickdraw} is the only dataset in Target pretraining at 5k steps outperformed directly transferring the Base model, indicating that the direct transfer performance from ImageNet is quite poor due to a large domain gap -- an observation further supported by its relatively poor domain adaptation in \cite{peng2019moment}. 

HPT improved performance on RESISC, VIPER, BDD, Flowers, xView, and \texttt{clipart}, \texttt{infograph}, and \texttt{sketch}: a diverse range of image domains and types. HPT had similar performance as Base transfer for the datasets that were most similar to ImageNet: \texttt{real}, COCO-2014, and Pascal, as well as for UC-Merced, which had 98.2\% accuracy for Base transfer and 99.0\% accuracy for HPT and HPT-BN. The two medical datasets, Chexpert and Chest-X-ray-kids had comparable performance with HPT and Target pretraining, yet HPT reached equivalent performance in 5k steps compared to 200k and 100k, respectively. Finally, HPT exhibited overfitting characteristics after 5k steps, where the overfitting was more pronounced on the smaller datasets (UC-Merced, Flowers, Chest-X-ray-kids, Pascal), leading us to recommend a very short HPT pretraining schedule, e.g.~5k iterations, regardless of dataset size. We further investigate these overfitting characteristics in the appendix.

\textbf{Semi-supervised transferability:}
Next, we conducted a semi-supervised transferability evaluation of the pretrained models. This experiment tested whether the benefit from the additional pretraining is nullified when finetuning all model parameters. Specifically, we selected the top performing models from the linear analysis for each pretraining strategy and fully finetuned the pretrained models using 1000 randomly selected labels without class balance but such that each class occured at least once. We finetune using a combination of two learning rates (0.01, 0.001) and two finetuning schedules (2500 steps, 90 epochs) with a batch size of 512 and report the top result for each dataset and model 
-- see the appendix for all details. %

\textbf{Key observations:} Figure~\ref{fig:ex2} shows the top finetuning performance for each pretraining strategy. The striped bars show the HPT pretraining variants, and we observe that similar to the linear analysis, HPT has the best performing pretrained models on 15 out of 16 datasets, with \texttt{quickdraw} being the exception. One key observation from this experiment is that HPT is beneficial in the semi-supervised settings and that the representational differences from HPT and the Base model are different enough that full model finetuning cannot account for the change.
We further note that while HPT-BN outperformed HPT in several linear analyses, HPT-BN never outperformed HPT when finetuning all parameters. This result indicates that some of the benefit from pretraining only the batch norm parameters is redundant with supervised finetuning. We also note that whether Base or Target pretraining performed better is highly dependent on the dataset, while HPT had uniformly strong performance.

\begin{figure*}[t]
\captionsetup[subfigure]{labelformat=empty}
\centering
\stackunder[5pt]{\includegraphics[width=5.1cm]{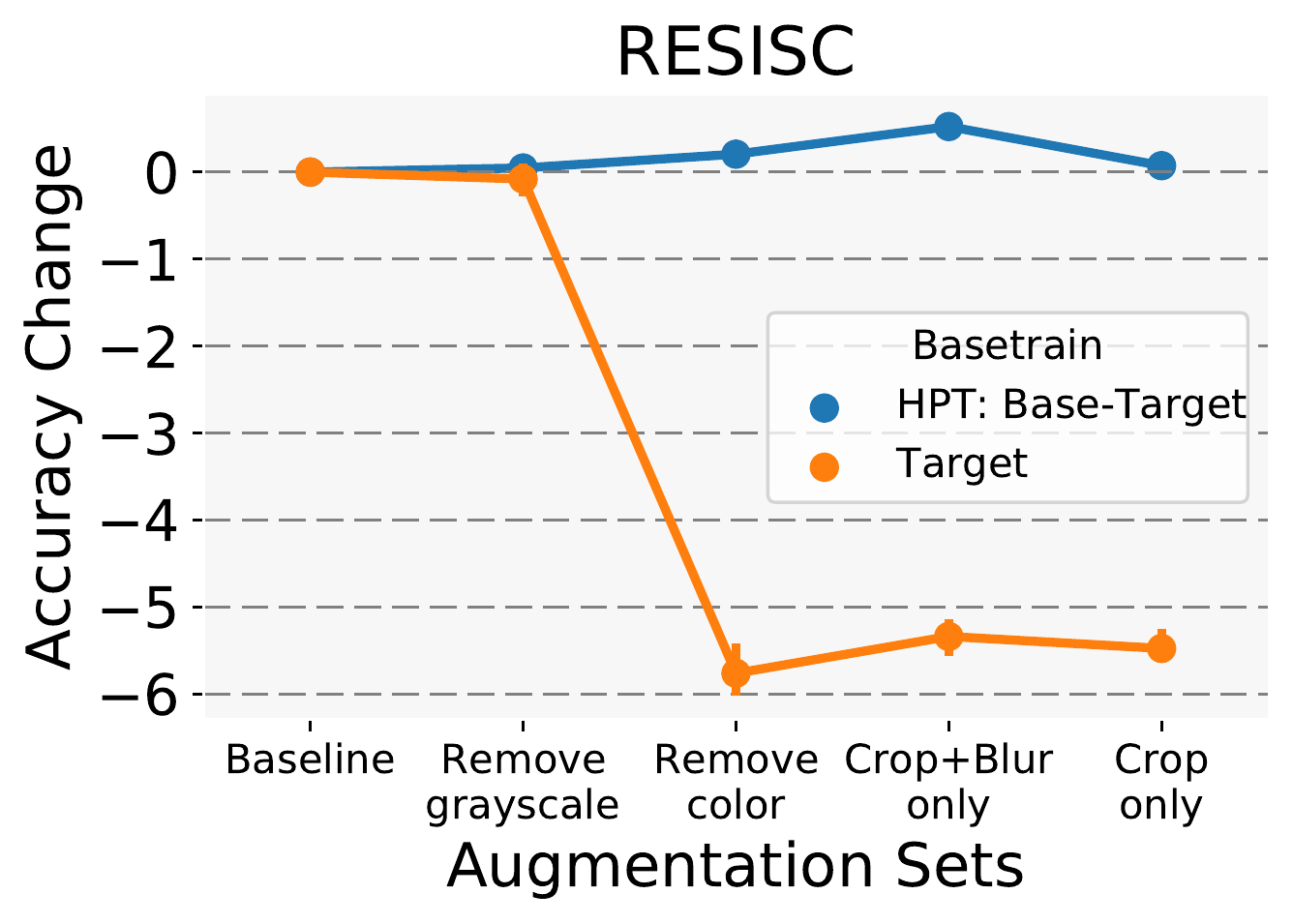}}{}
\hspace{4pt}%
\stackunder[5pt]{\includegraphics[width=5.1cm]{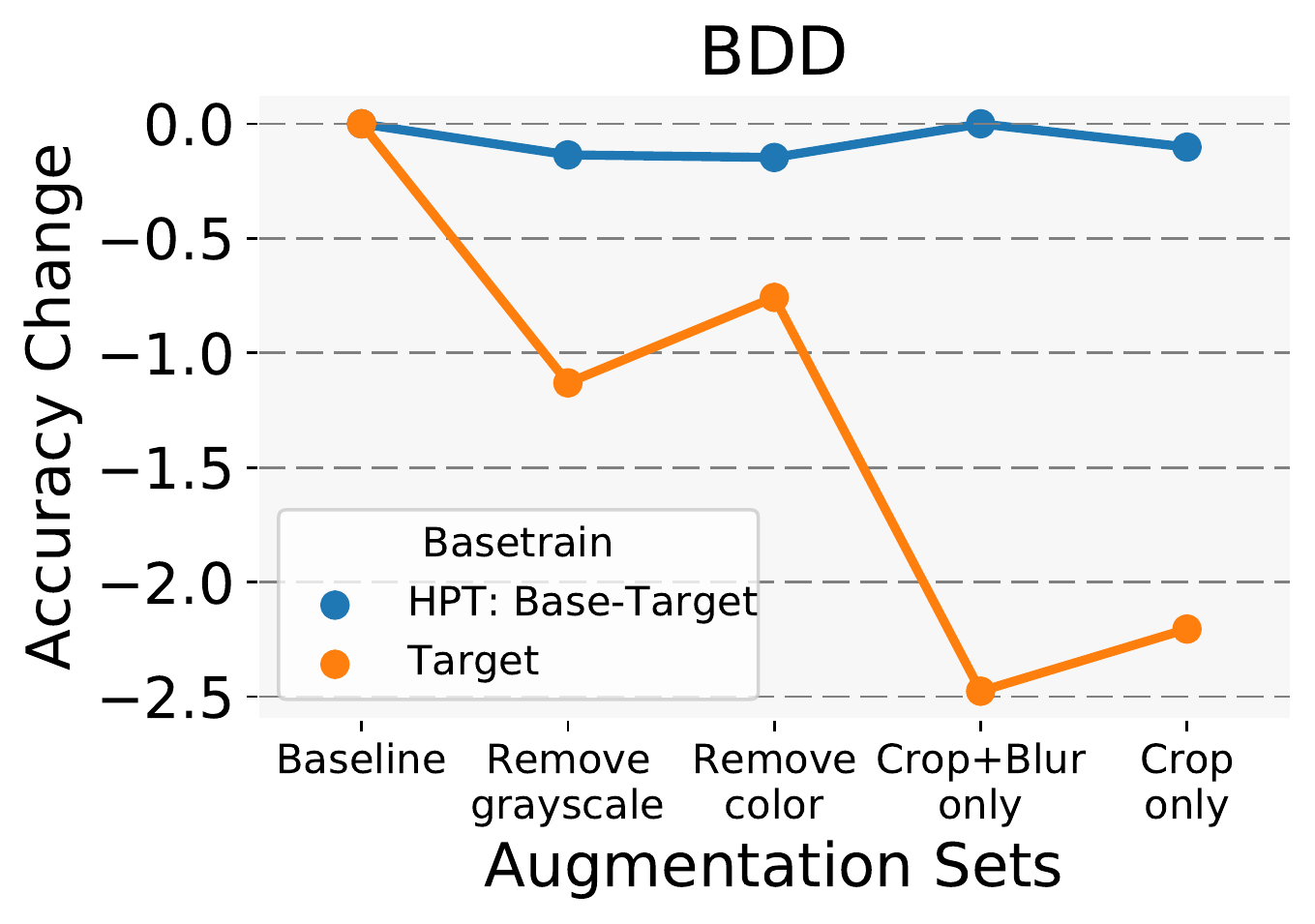}}{}
\hspace{4pt}%
\stackunder[5pt]{\includegraphics[width=5.1cm]{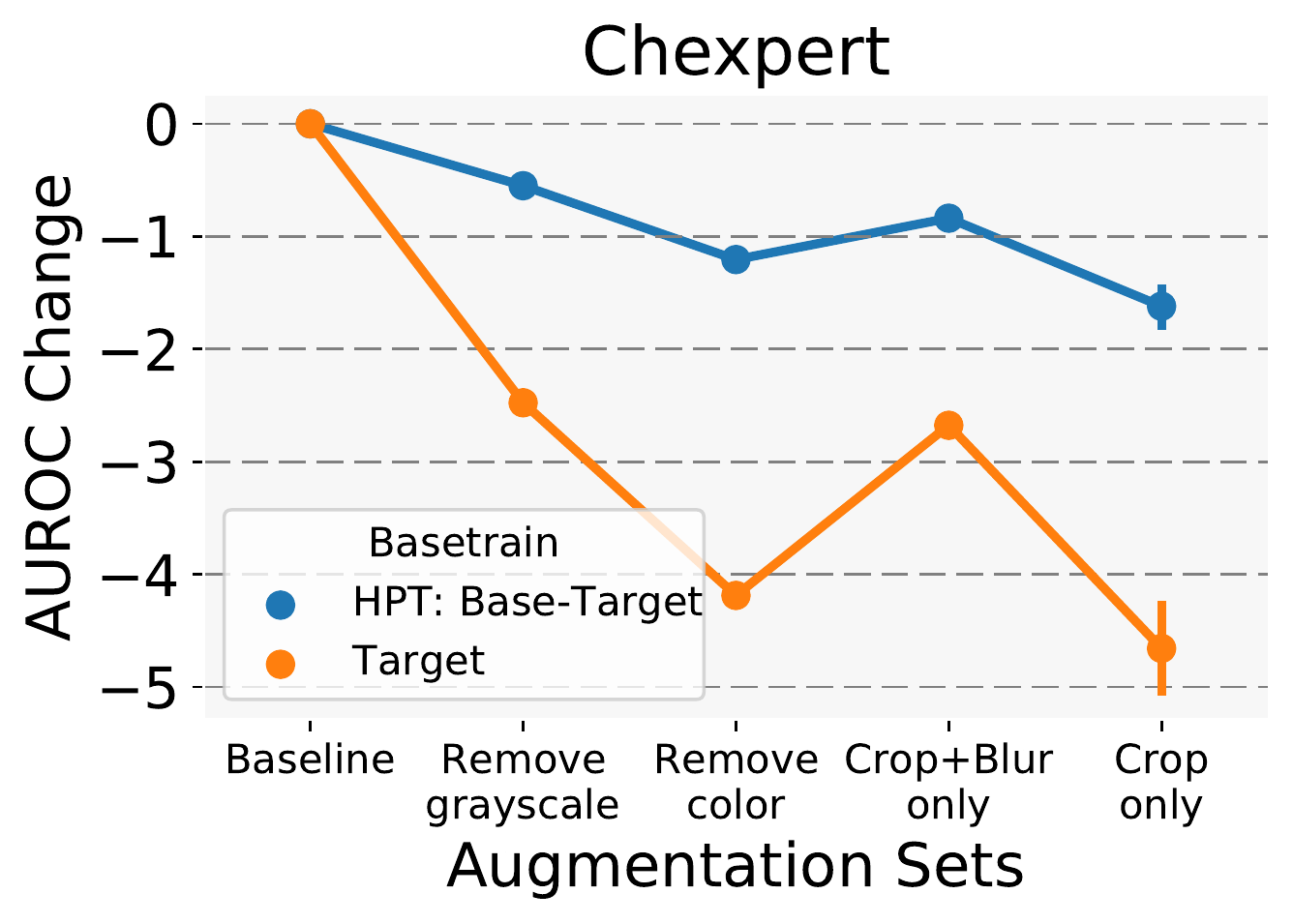}}{}
\caption{\emph{Augmentation robustness}. We compare the accuracy change of sequentially removing data augmentation policies (\texttt{Grayscale},
\texttt{ColorJitter}, 
\texttt{RandomHorizontalFlip}.
\texttt{GaussianBlur})
on linear evaluation performance. HPT performs better with only cropping than any other policy does with any incomplete combination.}
\label{fig:augrobust}%
\end{figure*}

\textbf{Sequential pretraining transferability: }
Here, we explore HPT's performance when pretraining on a source dataset before pretraining on the target dataset and finally transferring to the target task. 
We examined three diverse target datasets: Chest-X-ray-kids, \texttt{sketch}, and UC-Merced.
We select the source dataset for each of the target dataset by choosing the source dataset that yielded the highest linear evaluation accuracy on the target dataset after 5k pretraining steps on top of the base model. This selection yielded the following HPT instantiations: ImageNet then Chexpert then Chest-X-ray-kids, ImageNet then \texttt{clipart} then \texttt{sketch}, and ImageNet then RESISC then UC-Merced.

\textbf{Key observations:} Figure~\ref{fig:ex3} compares finetuning the 1000-label subset of the target data after the following pretraining strategies: directly using the Base model (B), Target pretraining (T), Base then Source pretraining (B+S), Base then Target pretraining (B+T), Base then Source pretraining then Target pretraining (B+S+T), and Base then Source pretraining then Target pretraining on the batch norm parameters (B+S+T-BN). The full HPT pipeline (B+S+T) leads to the top results on all three target datasets. In the appendix, we further show that the impact of an intermediate source model decreases with the size of the target dataset.

\textbf{Object detection and segmentation transferability}: For Pascal and BDD, we transferred HPT pretrained models to a Faster R-CNN R50-C4 model and finetuned the full model; for COCO, we used a Mask-RCNN-C4. 
Over three runs, we report the median results using the COCO AP metric as well as AP$_{50}$/AP$_{75}$. For Pascal, we performed finetuning on the \texttt{train2007+2012} set and performed evaluation on the \texttt{test2007} set. For BDD we used the provided train/test split, with 10k random images in the train split used for validation. For COCO, we used the 2017 splits and trained with the 1x schedule (see appendix for all details).

\textbf{Key observations:} Tables~\ref{tab:bddvoc}-\ref{tab:coco} show the object detection and segmentation results. For Pascal, we tested HPT instantiations of Base-Target, Base-Target (BN), and Base-Source-Target, where COCO-2014 was selected as the source model using the top-linear-analysis selection criteria. For the larger BDD and COCO datasets, we tested Base-Target and Base-Target (BN). Overall, the results are consistent across all datasets for image classification, object detection, and segmentation: HPT: both Base-Target and Base-Target (BN) lead to improvements over directly transferring the Base model to the target task. 
\begin{table}[t]
\small
\caption{Transfer Result: This table reports the median $\mbox{AP}$, $\mbox{AP}_{50}$, $\mbox{AP}_{75}$ over three runs of finetuning a Faster-RCNN C4 detector. For Pascal, the Source dataset is COCO-2014. A bold result indicates a $+0.2$ improvement over all other pretraining strategies.}
\label{tab:bddvoc}
\begin{tabular}{l|rrr}
\hline
\textbf{Pretrain}   & $\mbox{AP}^{\mbox{bb}}$           & $\mbox{AP}^{\mbox{bb}}_{50}$   & $\mbox{AP}^{\mbox{bb}}_{75}$ \\ \hline\hline
 \multicolumn{4}{c}{\texttt{Pascal VOC07}} \\ \hline
Target                         & 48.4                 & 75.9                 & 51.9                 \\
Base                           & 57.0                 & 82.5                 & 63.6                 \\
HPT: Base-Target               & 57.1                 & \textbf{82.7}                 & 63.7                 \\
HPT: Base-Target (BN)          & \textbf{57.5}        & \textbf{82.8}                 & 64.0                 \\
HPT: Base-Source-Target      & \textbf{57.5}          & \textbf{82.7}                 & \textbf{64.4}       \\
HPT: Base-Source-Target (BN) & \textbf{57.6}          & \textbf{82.9}                 & 64.2         \\ \hline
\multicolumn{4}{c}{\texttt{BDD}} \\ \hline
Target                  & 24.3                 & 46.9                 & 24.0                 \\
Base                    & 27.1                 & 48.7                 & 25.4                 \\
HPT: Base-Target        & \textbf{28.1}       & \textbf{50.0}        & \textbf{26.3}        \\
HPT: Base-Target  (BN)  & \textbf{28.0}        & 49.6                 & \textbf{26.3}                 \\\hline
\end{tabular}
\end{table}

The Base-Source-Target Pascal results show an improvement when pretraining all model parameters, but remain consistent when only pretraining the batch norm parameters. This indicates that while the batch norm parameters can find a better pretraining model, sequentially pretraining from the source to the target on these values does not always yield an improved result. Across datasets, the overall gains are relatively modest, but we view these results as an indication that HPT is not directly learning redundant information with either the MoCo pretraining on ImageNet or the finetuning task on the target dataset. Furthermore, it is surprising that only tuning the batch norm parameters on the target dataset leads to an improvement in object detection. From this result, we note that pretraining specific subsets of object detector backbone parameters may provide a promising direction for future work.

\begin{table}[t]
\small
\caption{Transfer Result: This table reports the median $\mbox{AP}$, $\mbox{AP}_{50}$, $\mbox{AP}_{75}$ over three runs of finetuning a Mask-RCNN-C4 detector on COCO-2017. A bold result indicates at least a $0.2$ improvement over all other pretraining strategies.}
\resizebox{\columnwidth}{!} {
\label{tab:coco}
\begin{tabular}{l|ccc|ccc}
\hline
\textbf{Pretrain} & $\mbox{AP}^{\mbox{bb}}$           & $\mbox{AP}^{\mbox{bb}}_{50}$   & $\mbox{AP}^{\mbox{bb}}_{75}$   & $\mbox{AP}^{\mbox{mk}}$           & $\mbox{AP}^{\mbox{mk}}_{50}$   & $\mbox{AP}^{\mbox{mk}}_{75}$   \\ \hline\hline
 Target        & 36.0                                 & 54.7                                 & 38.6                                 & 19.3     & 40.6                                 & 49.1                                 \\
Base        & 38.0    & 57.4                  & 41.3                                 & 20.7                                 & 43.3                                 & 51.4 \\
HPT: B-T     & \textbf{38.4}    & \textbf{58.0}          & 41.3                                 & \textbf{21.6}                                 & \textbf{43.5}                                & \textbf{52.2}                                 \\
HPT: B-T (BN)          & 38.2    & 57.4           & 40.9                                 & 20.6                                 & 43.4                                 & \textbf{52.2}                                 \\
\hline
\end{tabular}
}
\end{table}

\begin{figure*}[t]
\captionsetup[subfigure]{labelformat=empty}
\centering
    \includegraphics[width=5.5cm]{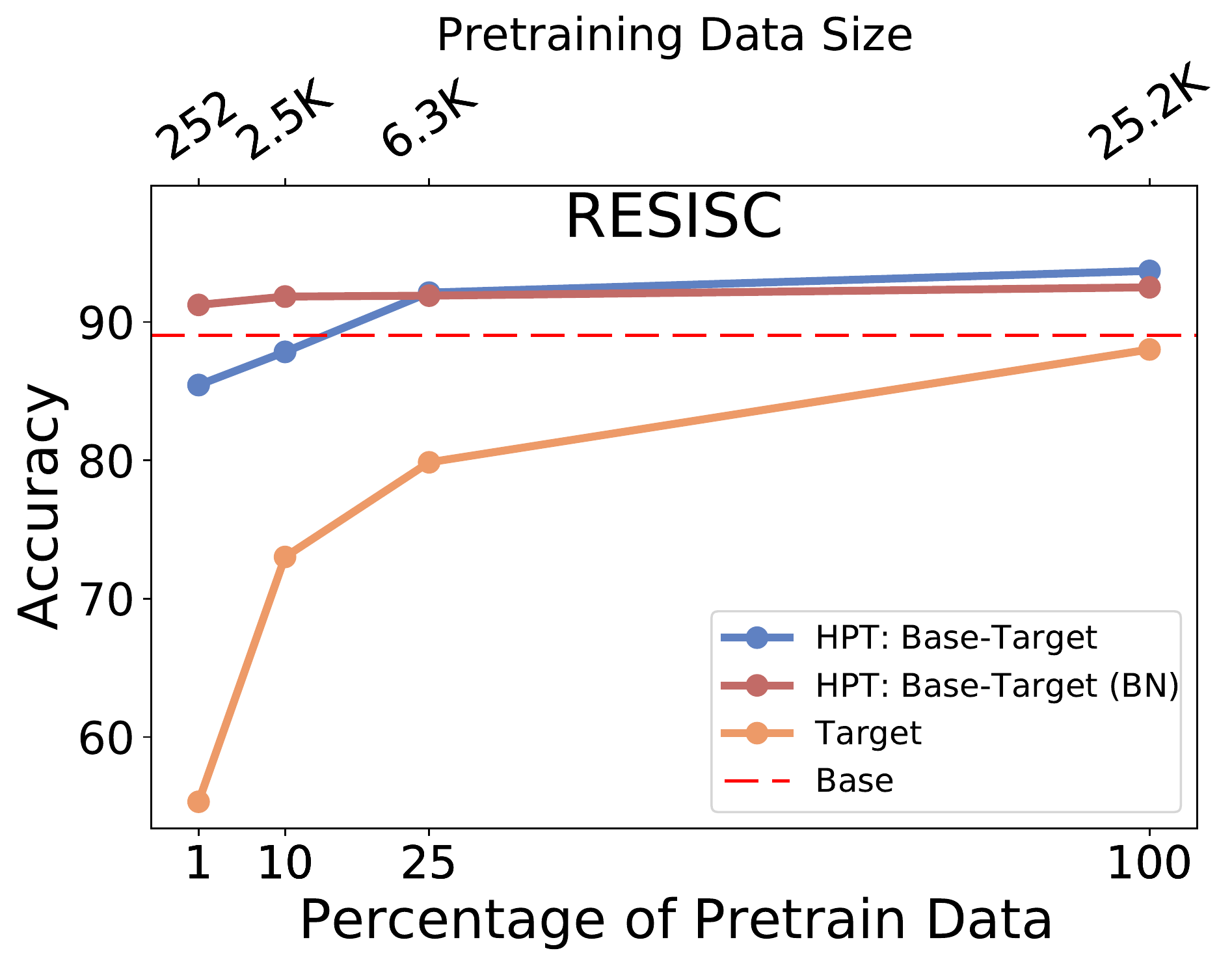}
    \includegraphics[width=5.5cm]{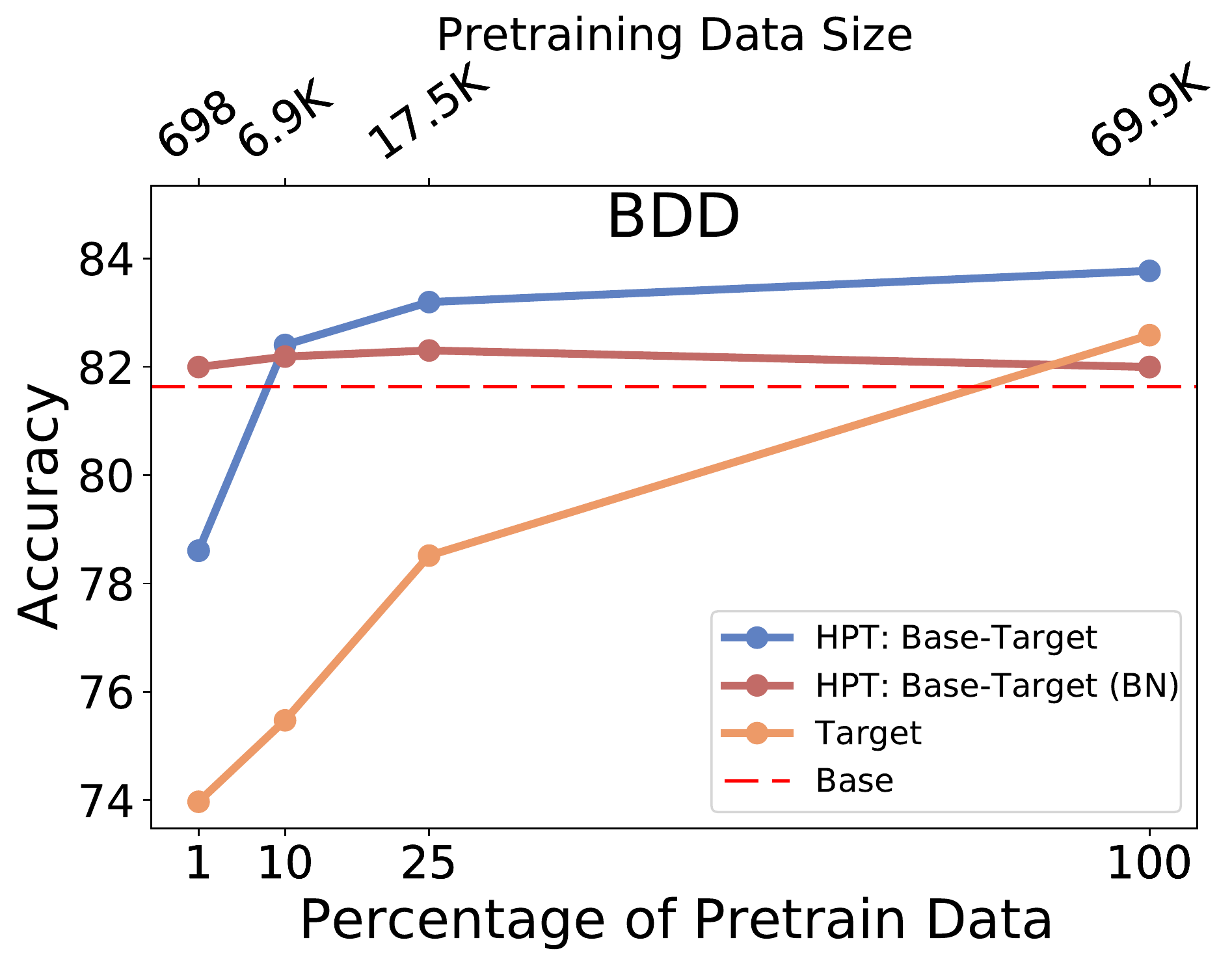}
\hspace{8pt}%
    \includegraphics[width=5.5cm]{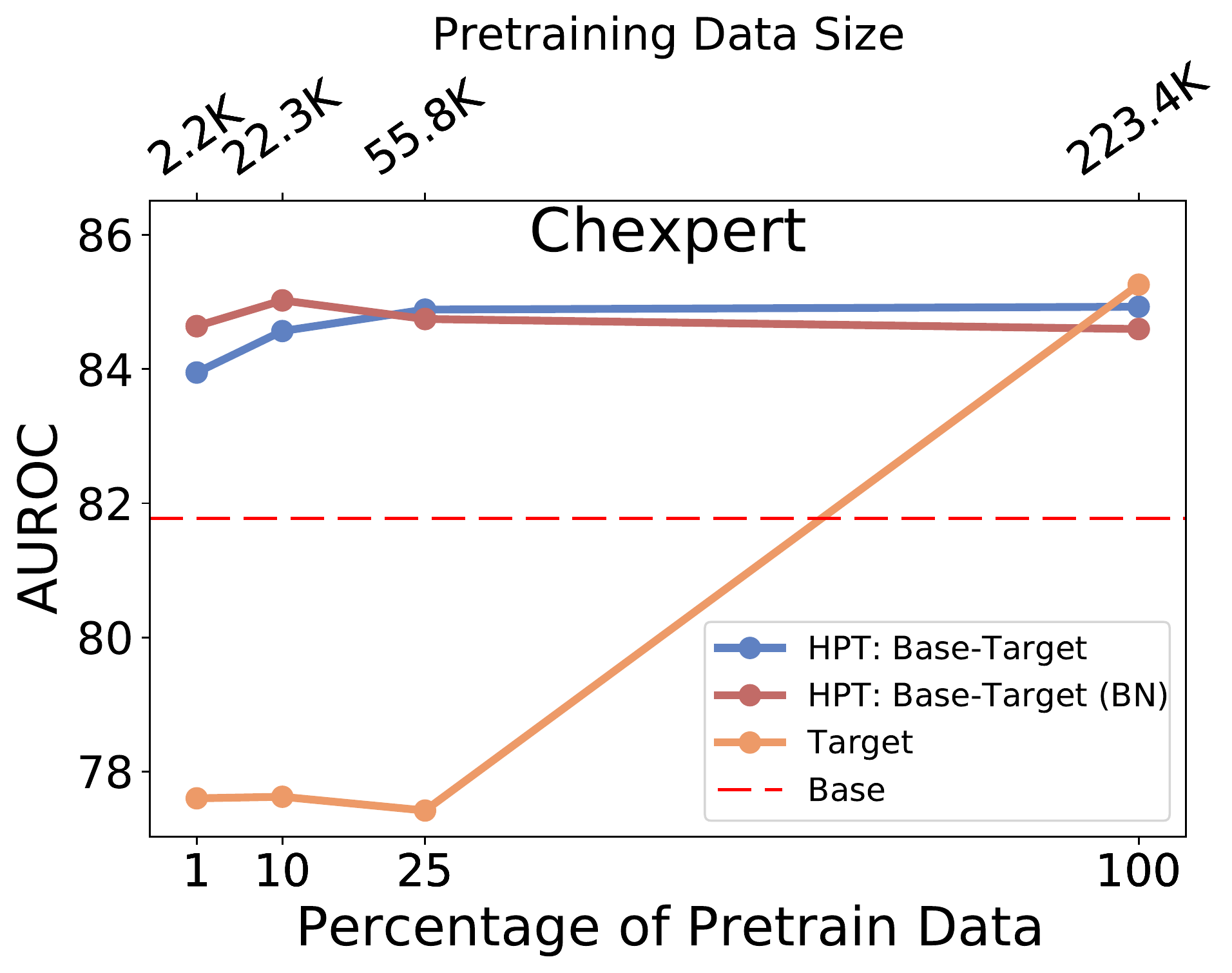}
\caption{
\emph{HPT performance as the amount of pretraining data decreases}. We show the linear evaluation performance as the amount of pretraining data varies. Top axis is the number of images, and the bottom is the percentage of pretraining data. HPT outperforms Base model transfer or Target pretraining with limited data. Notably, HPT-BN consistently outperforms Target pretraining with only 1\% of images. 
}
\label{fig:data_robustness}
\end{figure*}
\subsection{HPT Robustness}
Here, we investigate the robustness of HPT to common factors that impact the effectiveness of self-supervised pretraining such as the augmentation policy~\cite{chen_simple_2020, reed2020selfaugment} and pretraining dataset size~\cite{newell2020useful}.
For these robustness experiments, we used the BDD, RESISC, and Chexpert datasets as they provided a diversity in data domain and size. We measured separability, with the same hyperparameters as in \S\ref{ss:eval}.

\textbf{Augmentation robustness:}
MoCo-V2 sequentially applies the following image augmentations: \texttt{RandomResizedCrop}, \texttt{ColorJitter}, \texttt{Grayscale}, \texttt{GaussianBlur}, \texttt{RandomHorizontalFlip}. 
We studied the robustness of HPT by systematically removing these augmentations and evaluating the change in the linear evaluation for HPT and Target pretraining.

\textbf{Key observations:} Figure~\ref{fig:augrobust} shows separability results across datasets after sequentially removing augmentations. In all three data domains, HPT maintained strong performance compared to Target pretraining.
Unlike BDD and RESISC, the Chexpert performance decreased as the augmentation policy changed. This illustrates that changes to the augmentation policy can still impact performance when using HPT, but that the overall performance is more robust. In turn, as a practitioner explores a new data domain or application, they can either use default augmentations directly or choose a conservative set, e.g.~ only cropping.

\textbf{Pretraining data robustness:}
We pretrained with \{1\%, 10\%, 25\%, 100\%\} of the target dataset.
For HPT we used 5k pretraining steps.
For other methods with  25\% or 100\% of the data, we used the same number of steps as the top performing result in~Figure~\ref{fig:hpt-exp1-linear-analysis}. 
With 1\% or 10\% of the data, we use $1/10$ of the steps.

\textbf{Key observations:}
Figure~\ref{fig:data_robustness} shows separability results.
CheXpert has 3x more training data than BDD, which in turn has 3x more training data than Resisc.
While more data always performed better, the accuracy improvements of HPT increased as the amount of pretraining data decreased.
HPT-BN, while not achieving as high performance as HPT in all cases, had minimal accuracy degradation in low data regimes.
It consistently outperformed other methods with $<$5k samples.

\subsection{Domain Adaptation Case Study}
In this section, we explore the utility of HPT through a realistic case study experiment in which we apply HPT in a domain adaptation context. Specifically, in this experiment, the goal was to perform image classification on an unseen target domain given a labeled set of data in the source domain. We assume the target labels are scarcely provided with as few as 1 per class to 68 (see Table~\ref{tab:da-results}). We use a modern semi-supervised domain adaptation method called Minimax Entropy (MME)~\cite{mme} which consists of a feature encoder backbone, followed by a cosine similarity based classification layer that computes the features' similarity with respect to a set of prototypes estimated for each class. Adaptation is achieved by adversarially maximizing the conditional entropy of the unlabeled target data with respect to the classifier and minimizing it with respect to the feature encoder. 

The training procedure is as follows: we performed HPT to train a model using both source and target datasets on top of the standard MSRA ImageNet model~\cite{he2016deep}. 
We used this model to initialize the feature encoder in MME. At the end of each budget level we evaluated accuracy on the entire test set from the target domain. We perform two experiments on DomainNet datasets~\cite{peng2019moment} with 345 classes in 7 budget levels with increasing amount of target labels: (i) from \texttt{real} to \texttt{clip} and (ii) from \texttt{real} to \texttt{sketch}. We use \texttt{EfficientNet\_B2}~\cite{efficientnet} as the backbone architecture. 

Table~\ref{tab:da-results} shows our results for both domain adaptation experiments using MME with and without HPT. From the results, we observe that HPT consistently outperforms the baseline on both domains by achieving a higher accuracy across all the budget levels.  On the extreme case of low data regime (one shot/class), HPT achieves nearly 8\% better accuracy in both \texttt{clipart} and \texttt{sketch} domains in the extreme case of providing one shot per class in the target domain. This gap shrinks to 2\% as we increase the number of labeled target samples to 68 shots per class which is equivalent to 23,603 samples. These results demonstrate the effectiveness of HPT when applied as a single component in a realistic, end-to-end inference system.

\begin{table*}[t]
\begin{center}
\begin{small}
\caption{Budget levels and test accuracy in target domain for semi-supervised domain adaptation at 7 budget levels using MME with and without HPT between \texttt{real}$\rightarrow$\texttt{clip} and \texttt{real}$\rightarrow$\texttt{sketch}. At the single shot/class budget level, HPT achieves nearly 8\% better accuracy in both \texttt{clipart} and \texttt{sketch} domains. This gap shrinks to 2\% as we increase the number of labeled target samples to 68 shots per class which is equivalent to 23,603 samples.}
\label{tab:da-results}
\vskip 0.1in
\begin{tabular}{lcccccccc}
\toprule
\multicolumn{8}{c}{Budget levels in target domains} \\ \hline
\# of shots per class & \multicolumn{1}{c}{1} & \multicolumn{1}{c}{11} & \multicolumn{1}{c}{16} & \multicolumn{1}{c}{22} & \multicolumn{1}{c}{32} & \multicolumn{1}{c}{46} & \multicolumn{1}{c}{68} \\
\# of samples & \multicolumn{1}{c}{345} & \multicolumn{1}{c}{3795} & \multicolumn{1}{c}{5470} & \multicolumn{1}{c}{7883} & \multicolumn{1}{c}{11362} & \multicolumn{1}{c}{16376} & \multicolumn{1}{c}{23603} \\
\midrule
\multicolumn{8}{c}{Test accuracy (\%) for \texttt{real}$\rightarrow$\texttt{clip}} \\
\hline
MME   & 49.74 & 61.11 & 63.87 & 66.68 & 68.01 & 69.99 & 71.09 \\
MME+HPT  & \textbf{57.15} & \textbf{64.36} & \textbf{66.67} & \textbf{68.20} & \textbf{69.66} & \textbf{71.47} & \textbf{72.35 }\\
\midrule
\multicolumn{8}{c}{Test accuracy (\%) for \texttt{real}$\rightarrow$\texttt{sketch}} \\ \hline
MME   & 41.35 & 51.78 & 54.90 & 57.51 & 59.70 & 61.36 & 62.45 \\
MME+HPT & \textbf{50.17} & \textbf{56.43} & \textbf{58.77} & \textbf{60.72} & \textbf{62.80} & \textbf{63.91} & \textbf{64.90} \\
\bottomrule
\end{tabular}
\end{small}
\end{center}
\vskip -0.1in
\end{table*}

\section{Discussion}
We have shown that HPT achieves faster convergence, improved performance, and increased robustness, and that these results hold across data domains. Here, we further reflect on the utility of the HPT.

\textbf{What is novel about HPT?} 
The transfer learning methodology underlying HPT is well established in transfer learning. 
That is, transfer learning tends to work in a lot of situations, and our work could be perceived as a natural extension of this general observation.
However, our work provides the first thorough empirical analysis of transfer learning applied to self-supervised pretraining in computer vision. We hope this analysis encourages practitioners to include an HPT baseline in their investigations -- a baseline that is surprisingly absent from current works.

\textbf{How should I use HPT in practice?} 
We provide our code, documentation, and models to use HPT and reproduce our results\footnote{Code and pretrained models are available at \url{https://github.com/cjrd/self-supervised-pretraining}.}. For existing codebases, using HPT is usually as simple as downloading an existing model and updating a configuration. If working with a smaller dataset (e.g. $< 10$k images), our analysis indicates that using HPT-BN is ideal.

\textbf{Does this work for supervised learning?}
Yes. In the appendix, we reproduce many of these analyses using supervised \inet base models and show that HPT further improves performance across datasets and tasks.

\section{Conclusion and Implications}
Our work provides the first empirical analysis of transfer learning applied to self-supervised pretraining for computer vision tasks. In our experiments, we have observed that HPT resulted in 80x faster convergence, improved accuracy, and increased robustness for the pretraining process. These results hold across data domains, including aerial, medical, autonomous driving, and simulation.
Critically HPT requires fewer data and computational resources than prior methods, enabling wider adoption of self-supervised pretraining for real-world applications. 
Pragmatically, our results are easy to implement and use: we achieved strong results without optimizing hyperparameters or augmentation policies for each dataset. Taken together, HPT is a simple framework that improves self-supervised pretraining while decreasing resource requirements.

\paragraph{Funding Acknowledgements}
Prof. Darrell’s group was supported in part by DoD, NSF as well as BAIR and BDD at Berkeley, and Prof. Keutzer's group was supported in part by Alibaba, Amazon, Google, Facebook, Intel, and Samsung as well as BAIR and BDD at Berkeley.

{\small
\bibliographystyle{ieee_fullname}
\bibliography{refs}
}

\onecolumn
\clearpage
\appendix
\noindent\textbf{\Large Appendix}
\section{Implementation details}
\label{a:expdetails}

Table \ref{tab:params} lists the parameters used in the various training stages of the HPT pipeline. When possible, we followed existing settings from \cite{chen_improved_2020}. For the finetuning parameter sweeps, we followed a similar setting as the ``lightweight sweep'' setting from \cite{zhai2019large}.
We performed pretraining with the train and val splits. For evaluation, we used only the train split for training the evaluation task and then use the val split evaluation performance to select the top hyperparameter, training schedule, and evaluation point during the training. We then reported the performance on the test split evaluated with the best settings found with the val split. 

For the linear analysis and finetuning experiments, we used \texttt{RandomResizedCrop} to 224 pixels and \texttt{RandomHorizontalFlip} augmentations (for more on these augmentations, see \cite{chen_improved_2020}) during training. During evaluation, we resized the long edge of the image to 256 pixels and used a center crop on the image. All images were normalized by their individual dataset's channel-wise mean and variance. For classification tasks, we used top-1 accuracy and for multi-label classification tasks we used the Area Under the ROC (AUROC)~\cite{bradley1997use}.

For the 1000-label semi-supervised finetuning experiments, we randomly selected 1000 examples from the training set to use for end-to-end finetuneing of all layers, where each class occurred at least once, but the classes were not balanced. Similar to \cite{zhai2019large}, we used the original validation and test splits to improve evaluation consistency.

For all object detection experiments, we used the R50-C4 available in Detectron2 \cite{wu2019detectron2}, where following \cite{he2019momentum}, the backbone ends at conv4 and the box prediction head consists of conv5 using global pooling followed by an additional batchnorm layer. 
For PASCAL object detection experiments, we used the \texttt{train 2007+2012} split for training and the \texttt{val2012} split for evaluation. We used 24K training steps with a batch size of 16 and all hyperparameters the same as \cite{he2019momentum}.
For BDD, we used the 70K BDD \texttt{train} split for training and 10K \texttt{val} split for evaluation. We used 90K training steps with a batch size of 8 on 4 GPUs.
For CoCo object detection and segmentation, we used the 2017 splits, with the 1$\times (\sim$12$)$ epochs training schedule with a training batch size of 8 images over 180K iterations on 4 GPUs half of the default learning rate (note: many results in the literature (e.g. \cite{he2019momentum}) use a batch size of 16 images over 90K iterations on 8 GPUs with the full default learning rate, which leads to slightly improved results ($+$0.1-0.5 AP). For semantic segmentation, we used Mask-RCNN~\cite{he2017mask} with a C4 backbone setting as in \cite{he2019momentum}. 

\begin{table}[htb]
\centering
\caption{This table provides the parameters that were used for pretraining, linear, and finetuning analyses carried out in this paper (unless otherwise noted). Multiple values in curly braces indicate that all combinations of values were tested, i.e.~in order to find an appropriate evaluation setting. 10x decay at $\frac{1}{3}$ and $\frac{2}{3}$ corresponds to decaying the learning rate by a factor of 10 after $\frac{1}{3}$ and $\frac{2}{3}$ of training steps have occurred, respectively.}  
\label{tab:params}
\begin{tabular}{llll}
\toprule
\textbf{Parameter} & \textbf{MoCo-V2 Value} & \textbf{Linear Value} & \textbf{Finetune Value} \\
\midrule
 batch size & 256 & 512 & 256 \\
 num gpus & 4 & 4 & 4  \\
 lr & 0.03 & \{0.3, 3, 30\} & \{0.001, 0.01\}  \\
 schedule & cosine & 10x decay at $\frac{1}{3}, \frac{2}{3}$ & 10x decay at $\frac{1}{3}, \frac{2}{3}$ \\
 optimizer & SGD & SGD & SGD \\
 optimizer momentum & 0.9 & 0.9 & 0.9 \\
 weight decay & 1e-4 & 0.0 & 0.0 \\
 duration & 800 epochs & 5000 steps &  \{2500 steps, 90 epochs\} \\
 moco-dim & 128  & - & -\\
 moco-k & 65536 & - & -\\
 moco-m & 0.999 & - & -\\
 moco-t & 0.2 & - & - \\

\bottomrule

\end{tabular}
\end{table}

\section{Datasets}
\label{a:data}
Table~\ref{tab:Dataset} lists the datasets used throughout our experiments.
For all evaluations, unless otherwise noted, we used top-1 accuracy for the single classification datasets and used the Area Under the ROC (AUROC)~\cite{bradley1997use} for multi-label classification tasks.

\begin{table*}[!t]
\centering%
\caption{Dataset Descriptions. We use $x/y/z$ to denote train/val/test split in each dataset.}
\begin{tabular}
{ccccc}
\toprule
\multicolumn{2}{c}{\textbf{Dataset}}& \textbf{Train/Validation/Test Size} & \textbf{Labels} & \textbf{Classification Type} \\
\hline
\multicolumn{2}{c}{BDD~\cite{yu2020bdd100k}} & 60K/10K/10K & 6 classes & singular \\
\multicolumn{2}{c}{Chest-X-ray-kids\cite{kermany2018large}} & 4186/1046/624 & 4 classes & singular \\
\multicolumn{2}{c}{Chexpert~\cite{irvin2019chexpert}} & 178.7K/44.6K/234 & 5 classes & multi-class \\
\multicolumn{2}{c}{Coco-2014~\cite{lin2014microsoft}} & 82.7K/20.2K/20.2K & 80 classes & multi-class \\
& Clipart & 27.2K/6.8K/14.8K & 345 classes & singular \\
& Infograph & 29.6K/7.4K/16.1K & 345 classes & singular \\
Domain Net & Painting & 42.2K/10.5K/22.8K & 345 classes & singular \\
\cite{peng2019moment} & Quickdraw & 96.6K/24.1K/51.7K & 345 classes & singular \\
& Real & 98K/24.5K/52.7K & 345 classes & singular \\
& Sketch & 39.2K/9.8K/21.2K & 345 classes & singular \\
\multicolumn{2}{c}{RESISC~\cite{cheng2017remote}} & 18.9K/6.3K/6.3K & 45 classes & singular \\
\multicolumn{2}{c}{VIPER~\cite{richter2017playing}} & 13.3K/2.8K/4.9K & 5 classes & singular \\
\multicolumn{2}{c}{UC Merced~\cite{yang2010bag}} & 1.2K/420/420 & 21 classes & singular \\
\multicolumn{2}{c}{Pascal VOC~\cite{everingham2015pascal}} & 13.2K/3.3K/4.9K & 20 classes & multi-class \\
\multicolumn{2}{c}{Flowers~\cite{nilsback2008automated}} & 1K/1K/6.1K & 103 classes & singular \\
\multicolumn{2}{c}{xView~\cite{lam2018xview}} & 39K/2.8K/2.8K & 36 classes & multi-class\\
\bottomrule
\end{tabular}
\label{tab:Dataset}
\end{table*}

\section{Additional Experiments and Ablations}
\label{a:generalization}

\subsection{HPT Learning Rate}
We investigated the choice of the learning rate used during HPT and its effect on linear evaluation performance (see Table~\ref{tab:lr}). Specifically, we tested initial learning rates \{0.1, 0.03, and 0.001\}, on  datasets: RESISC, BDD, and Chexpert. The following table shows that the default learning rate of 0.03 for batch size 256 from \cite{chen_improved_2020} outperformed the other configurations. Based on this experiment, we used the default 0.03 learning rate for all HPT pretraining runs.

\begin{table}[H]
\begin{center}
\caption{The following table shows the linear evaluation performance with HPT pretraining learning rates of \{0.1, 0.03, and 0.001\}. Based on this experiment, we continues to use the default 0.03 learning rate from \cite{chen_improved_2020}.}
\label{tab:lr}
\vskip 0.1in
\begin{tabular}{llll}
\multicolumn{1}{c}{\textbf{}}          & \multicolumn{3}{c}{\textbf{Learning Rate}} \\ \toprule
\multicolumn{1}{l|}{\textbf{Datasets}} & 0.1       & 0.03               & 0.001     \\ \midrule
\multicolumn{1}{l|}{RESISC}            & 92.5      & \textbf{93.7}      & 91.6      \\
\multicolumn{1}{l|}{BDD}               & 81.9      & \textbf{83.2}      & 82.4      \\
\multicolumn{1}{l|}{Chexpert}          & 79.5      & \textbf{85.8}      & 83.9      \\
\bottomrule
\end{tabular}
\end{center}
\end{table}

\subsection{HPT with Supervised Base Model}
\begin{figure*}[tbh] 
    \centering
    \includegraphics[width=\linewidth]{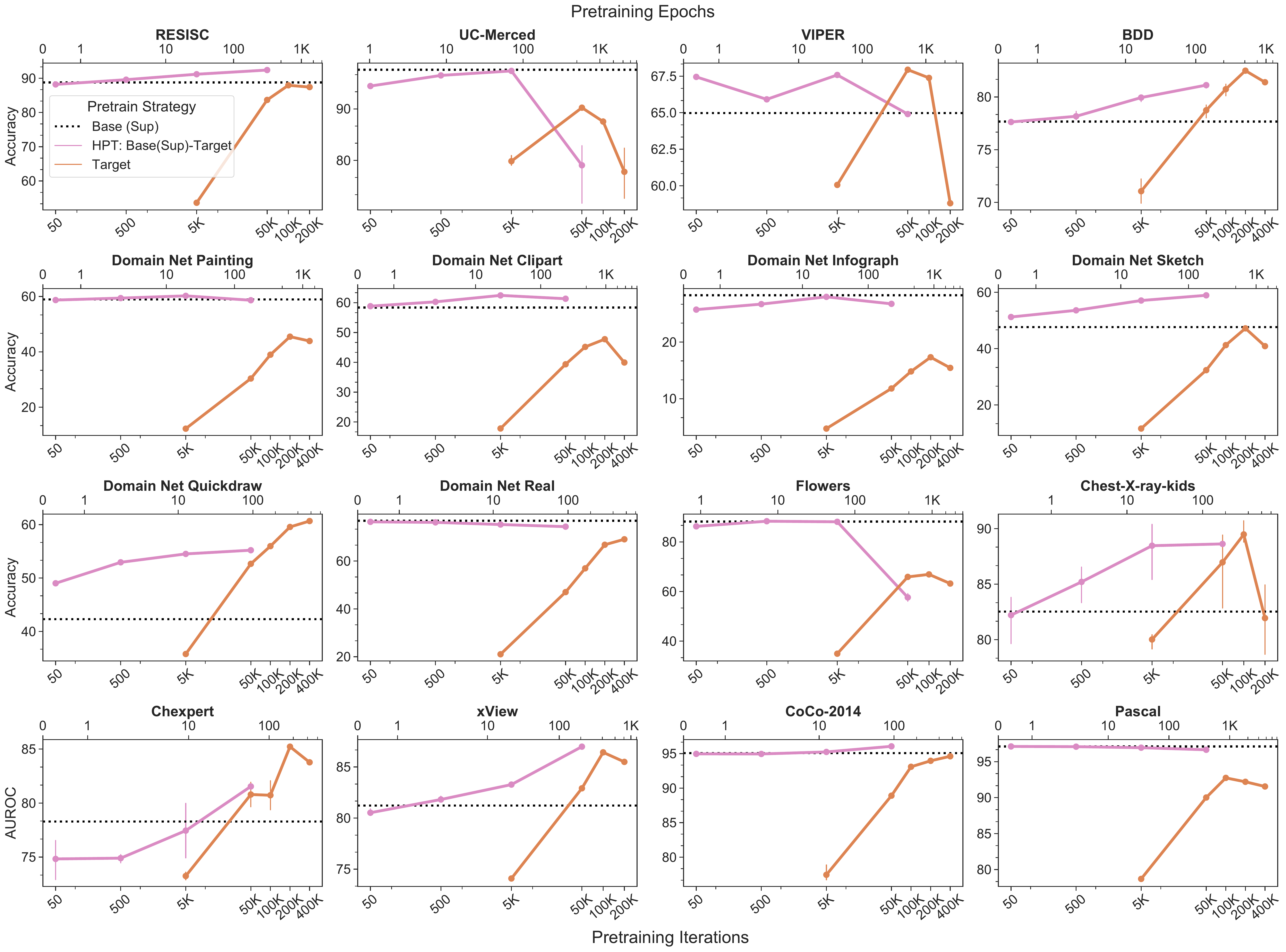}
     \caption{Linear eval: For each of the 16 datasets, we use a supervised ImageNet Base model~\cite{he2016deep}. We train the HPT framework for 50-50k iterations (HPT Base(sup)-Target). We compare it to a model trained from a random initialization (Target) trained from 5K-400K iterations.
     For each, we train a linear layer on top of the final representation. With a supervised base model, HPT obtains as good or better results on 13/16 datasets without hyperparameter tuning.}
     \label{fig:hpt-exp1-linear-analysis-supervised}
\end{figure*}

We explored using a supervised ImageNet base model from \cite{he2016deep} instead of the self-supervised MoCo model from \cite{chen_improved_2020}. Similar to the expepriments shown in Figure~\ref{fig:hpt-exp1-linear-analysis} in the main paper, Figure~\ref{fig:hpt-exp1-linear-analysis-supervised} shows the same results using a supervised base ImageNet model. 
We observe similar behavior as with the self-supervised base model: HPT with the supervised base model tends to lead to improved results compared to directly transferring the base model or pretraining entirely on the target data.
Unlike the self-supervised base model, HPT with a supervised base model often shows improved performance after 5K iterations, e.g.~at 50K iterations (RESISC, BDD, DomainNet \texttt{Sketch}, DomainNet \texttt{Quickdraw}, xView, Chexpert, CoCo-2014), indicating that the supervised base model needs longer training to obtain comparable linear evaluation results.
Also unlike the self-supervised base model, these results show clearly better Target model results for BDD and Chexpert, and a larger gap with DomainNet \texttt{Quickdraw}. This indicates that the supervised pretraining is less beneficial as a base model when the domain gap is large -- an observation further supported by the experiments in \cite{yang2020transfer}.

\begin{figure*}[tbh] 
    \centering
    \includegraphics[width=\linewidth]{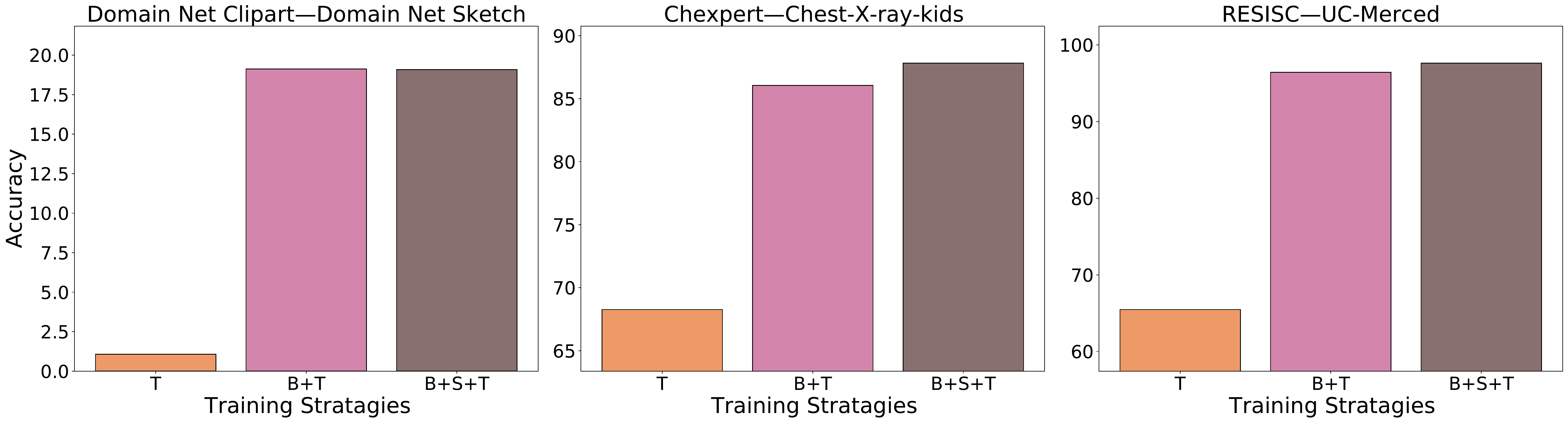}
     \caption{Finetuning performance on target datasets with supervised pretrainings. Here, we show results similar to the finetuning experiment results displayed in Figure~\ref{fig:ex3}, we investigated the finetuning performance on the same set of target datasets except using a supervised ImageNet base model~\cite{he2016deep} and supervised (S)ource pretraining for B+S+T. Overall, HPT again leads to improved performance over finetuning on the Target pretrained models for all three datasets. Different from the self-supervised base model used in Figure~\ref{fig:ex3}, all results for the Domain Net framework are considerably worse, and incorporating training on the source dataset (Domain Net Clipart) does not demonstrate an improvement in this case. Overall, however, HPT with the supervised base model leads to improved finetuning performance with and without the source training step.}
     \label{fig:exp-3-supervised}
\end{figure*}

In Figure~\ref{fig:exp-3-supervised}, we show results similar to the finetuning experiment results displayed in Figure~\ref{fig:ex3} in the main paper, we investigate the finetuning performance on the same set of target datasets except using a supervised ImageNet base model~\cite{he2016deep}. Overall, HPT again leads to improved performance over finetuning on the Target pretrained models for all three datasets. Different from the self-supervised base model used in Figure~\ref{fig:ex3}, all results for the DomainNet framework are considerably worse, and incorporating training on the source dataset (DomainNet \texttt{Clipart}) does not demonstrate an improvement in this case. Overall, however, HPT with the supervised base model leads to improved finetuning performance with and without the source training step.

\begin{figure*}[t] 
    \centering
    \includegraphics[width=0.33\textwidth]{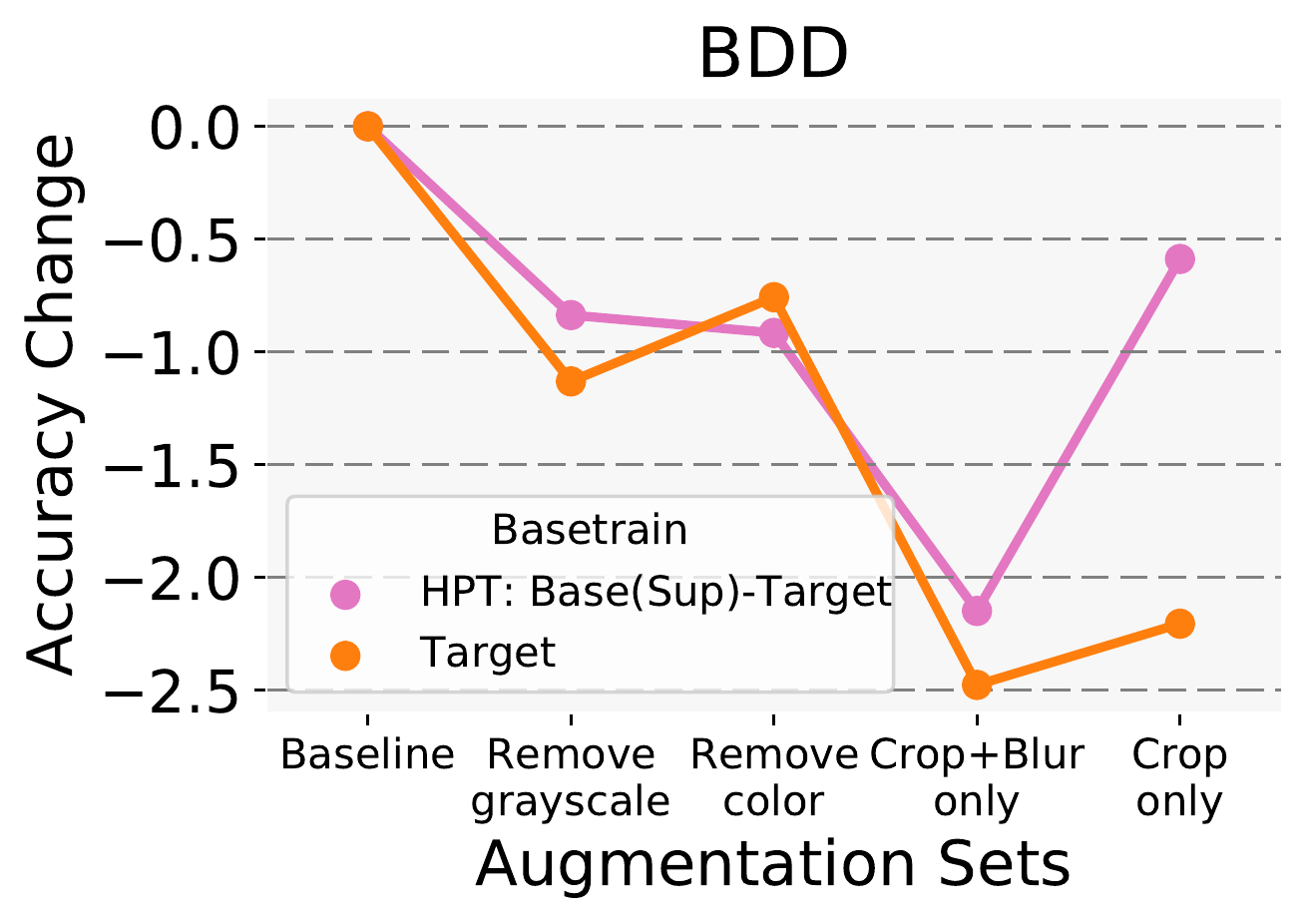}
    \includegraphics[width=0.33\textwidth]{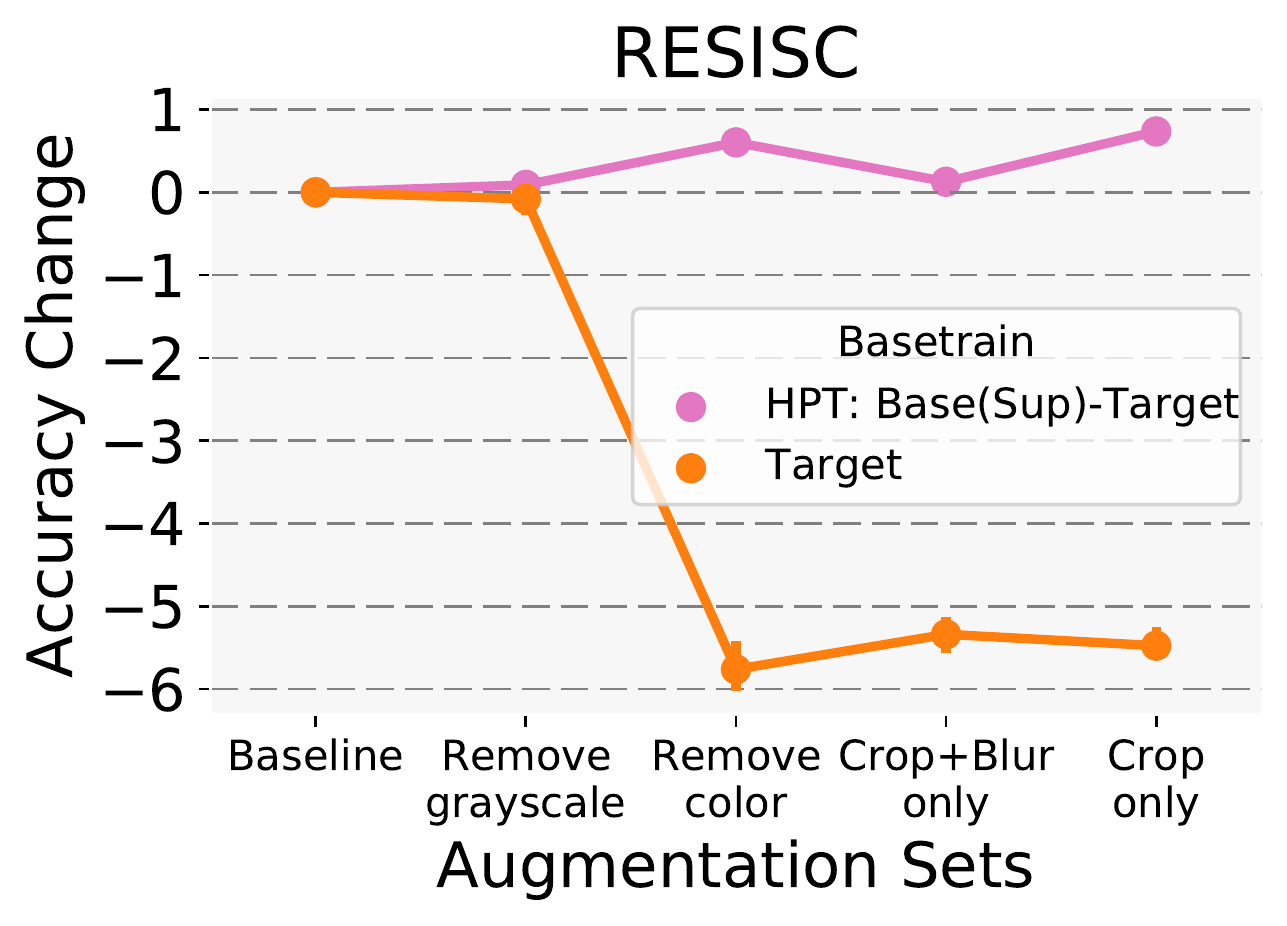}
    \includegraphics[width=0.33\textwidth]{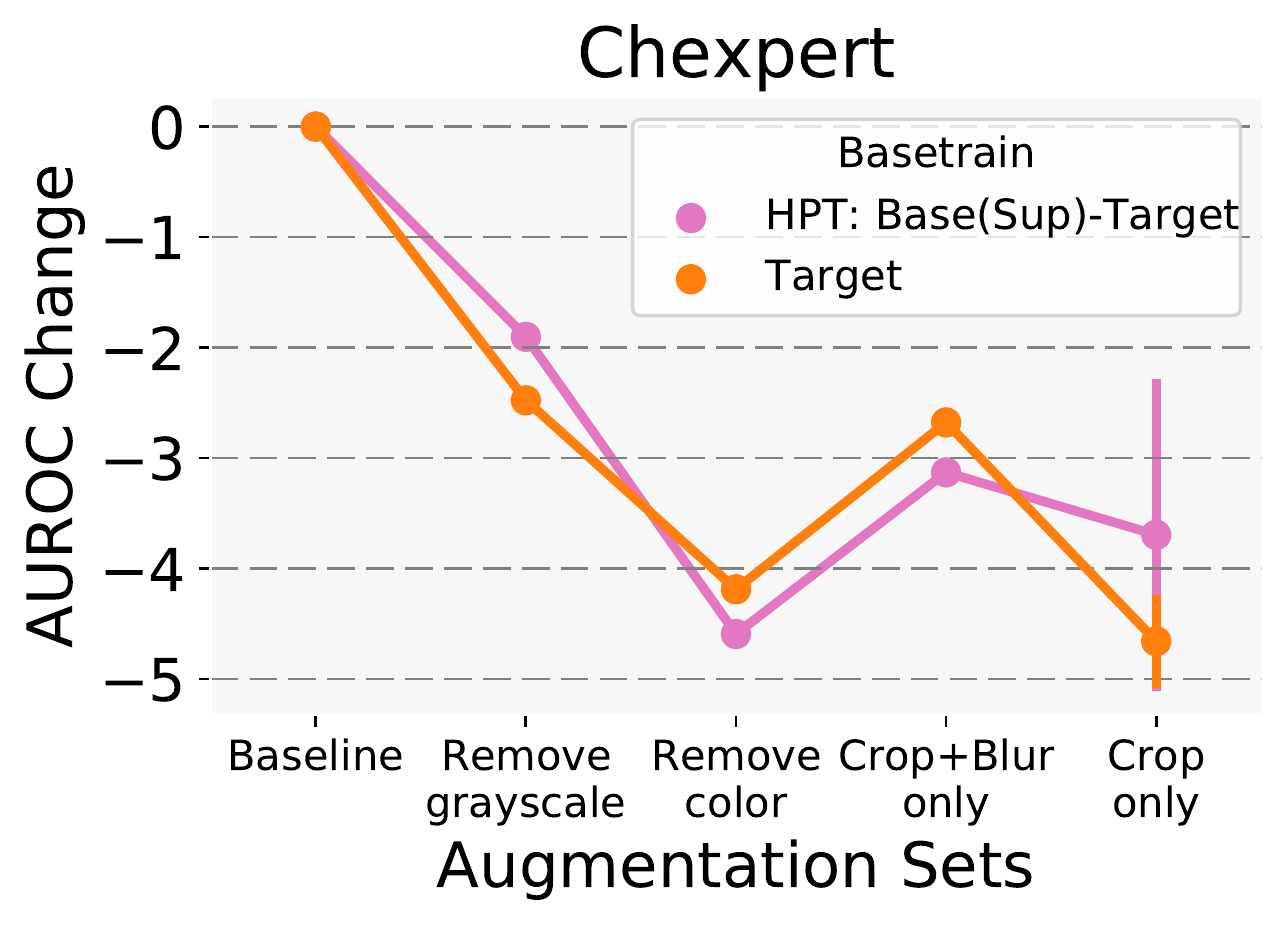}
     \caption{Supervised base model augmentation robustness. Here, we further studied the augmentation robustness of HPT when using a supervised base model (HPT-sup). We followed the same experimental procedure described in Section 4.4. As discussed in the text, these results show that HPT-sup exhibits more variation to the augmentations used during pretraining on BDD and Chexpert. As the supervised model was only trained with cropping and flipping augmentations, this indicates that the robustness from the base augmentations used in the self-supervised pretraining remain when performing further pretraining on the target dataset. }
     \label{fig:aug-robust-supervised}
\end{figure*}

\textbf{Augmentation robustness:} We studied the augmentation robustness of HPT when using a supervised base model (HPT-sup). We followed the same experimental procedure described in Section 4.4. Figure \ref{fig:aug-robust-supervised} shows the results using HPT-sup, while for comparison, Figure \ref{fig:augrobust} shows the results with HPT. Both HPT-sup and HPT demonstrate their robustness to the set of augmentations used while pretraining on RESISC. However, HPT-sup exhibits more variation to the augmentations used during pretraining on BDD and Chexpert. The supervised model was trained with only cropping and flipping augmentations, while the self-supervised pretraining took place with all augmentations in the shown policy. The robustness of the self-supervised base model indicates that the selection of the augmentation policy for further pretraining with the target dataset is resilient to changes in the set of augmentations used for pretraining the base model, and if these augmentations are not present, then the HPT framework loses its augmentation policy robustness.
\subsection{Basetrain Robustness}
\begin{figure}[t]
\captionsetup[subfigure]{labelformat=empty}
\centering
    \includegraphics[width=\linewidth/3]{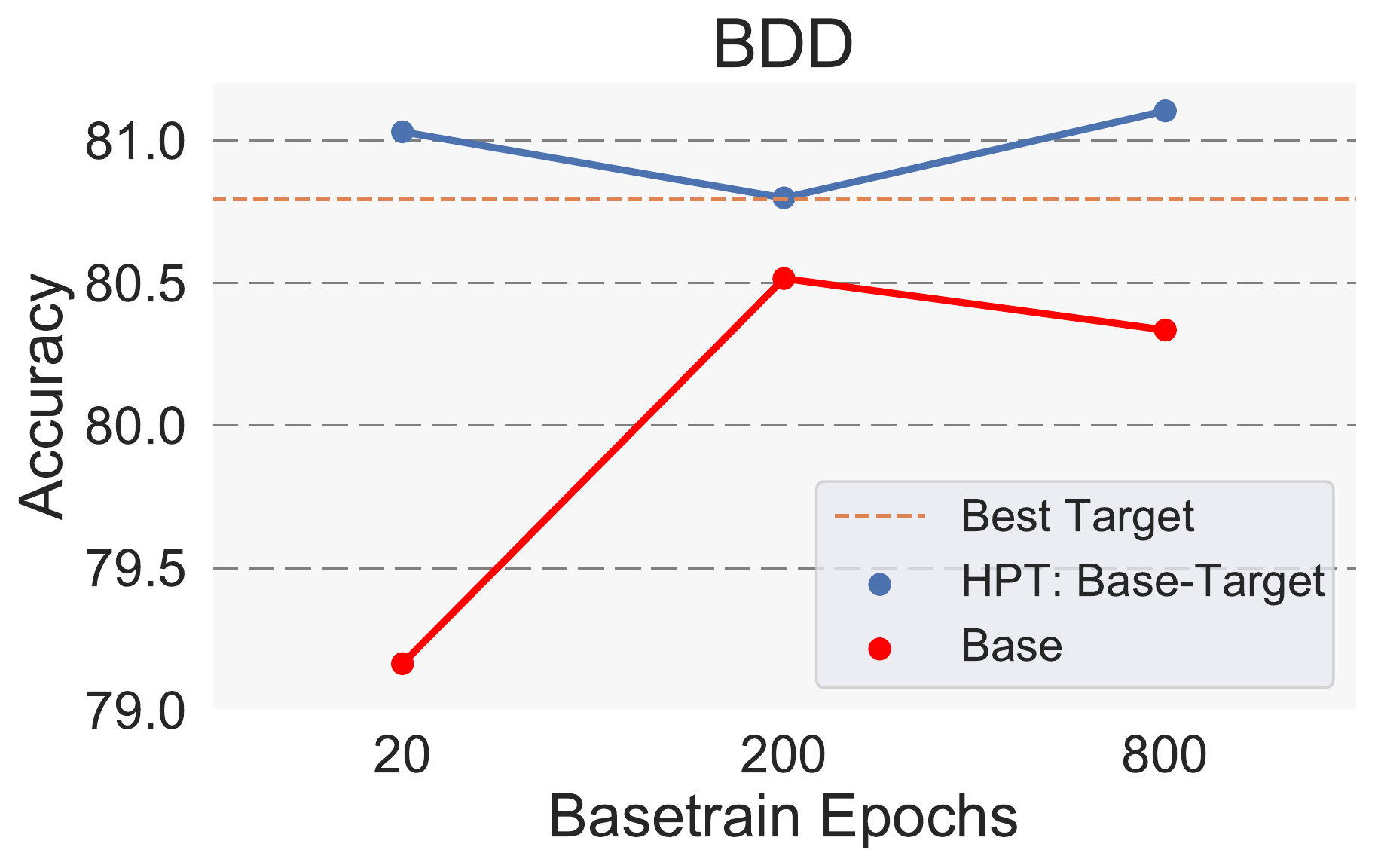}
    \includegraphics[width=\linewidth/3]{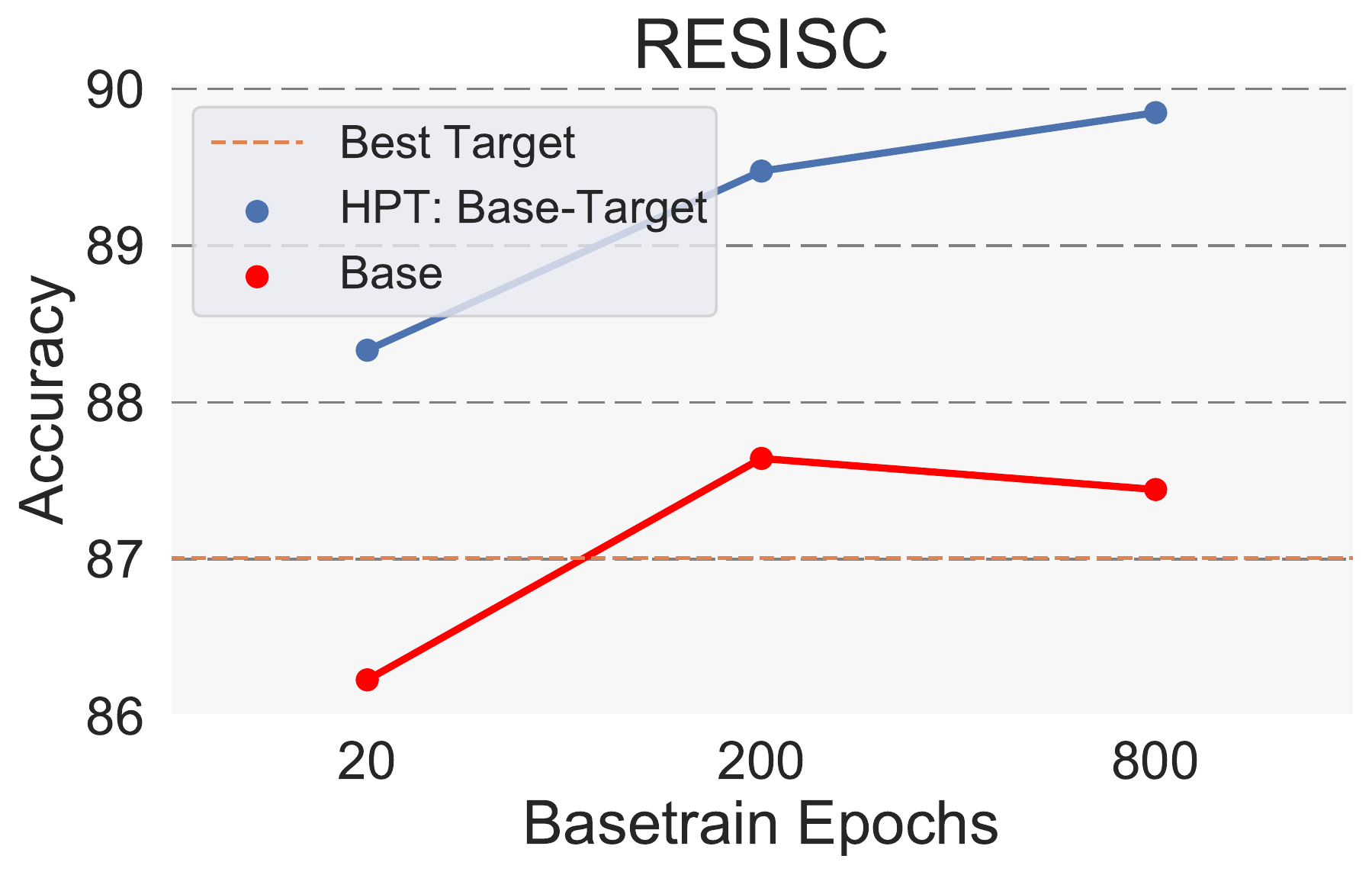}
    \includegraphics[width=\linewidth/3]{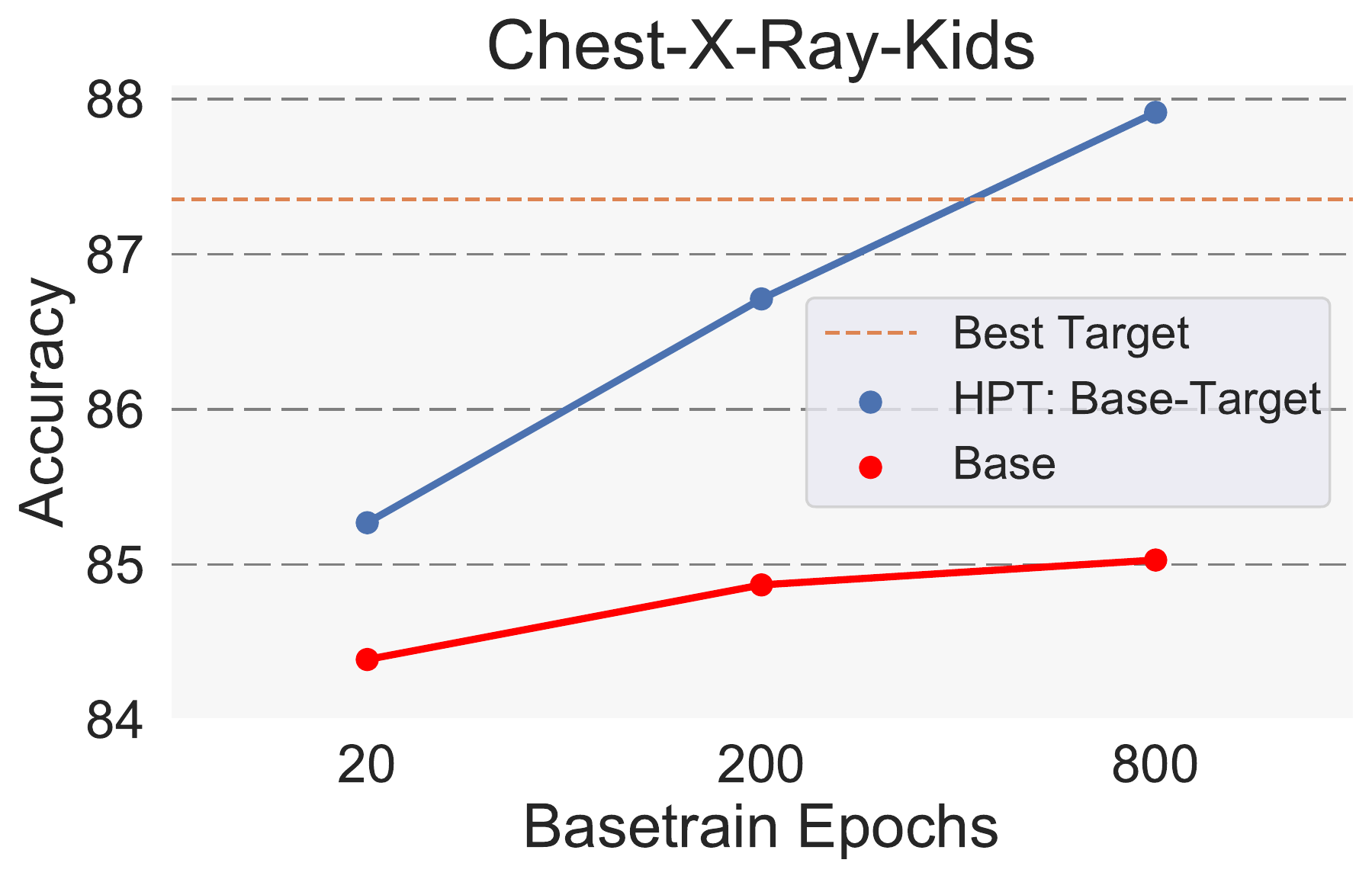}
    \includegraphics[width=\linewidth/3]{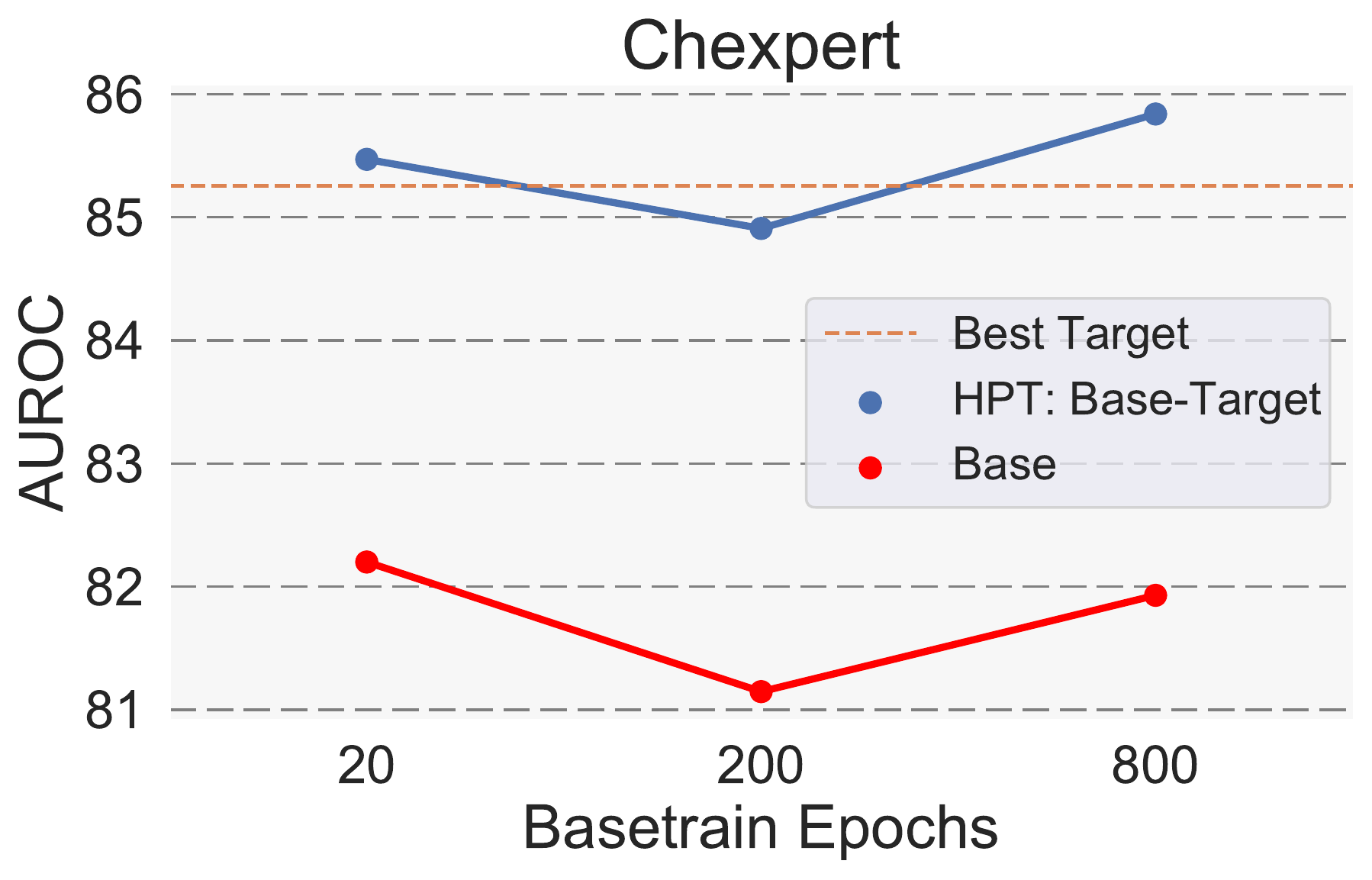}
    \includegraphics[width=\linewidth/3]{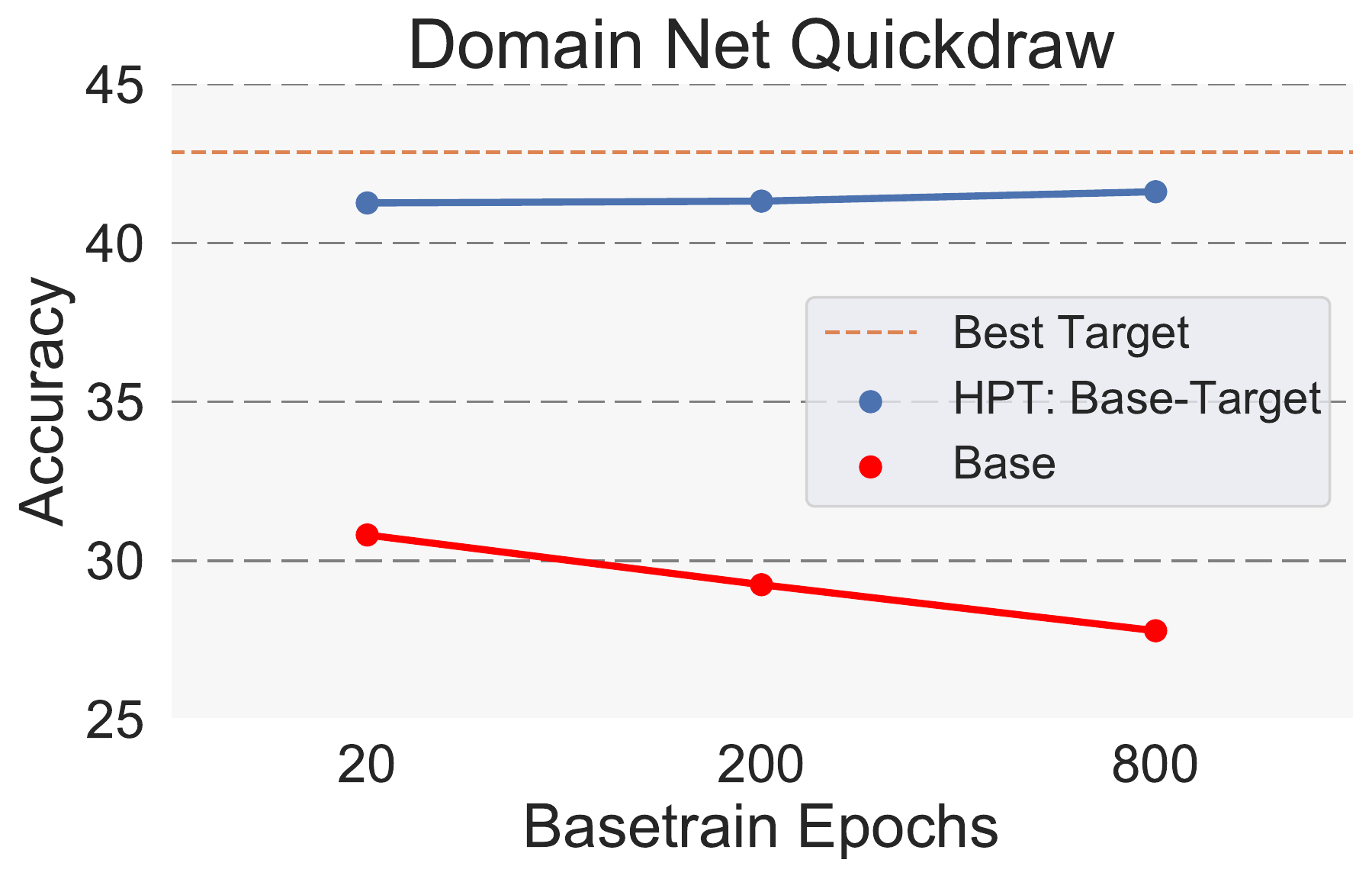}
\caption{These figures shows performance of linear evaluation on models pretrained on ImageNet for various epochs (in \textcolor{myblue}{blue}) and with addition HPT training (in \textcolor{myred}{red}). The best baseline model pretrained only on target data for at least the equivalent of 20 ImageNet epochs and 5K HPT steps is show as the orange dotted-line.}
\label{fig:base_robustness}%
\end{figure}

We explored how the linear analysis evaluation changed with varying the amount of self-supervised pretraining on the base model, e.g.~the initial ImageNet model. We tested base models trained for 20, 200, and 800 epochs, where the 200 and 800 epoch models were downloaded from the research repository from \cite{chen_improved_2020}\footnote{\url{https://github.com/facebookresearch/moco}}, and the 20 epoch model was created using their provided code and exact training settings. For each base model, we performed further pretraining on the target dataset for 5000 steps.

Figure~\ref{fig:base_robustness} shows the results for the RESISC, BDD, Chexpert, Chest-X-ray-kids, and DomainNet \texttt{Quickdraw} datasets. We note several characteristics of these plots: the 200 and 800 epoch datasets performed comparably across all datasets except Chest-X-ray-kids which displayed a drop in performance at 200 epochs, indicating that the extra self-supervised pretraining needed to obtain state-of-the-art linear ImageNet classification performance is typically not be necessary for strong HPT performance. Surprisingly, BDD, \texttt{Quickdraw}, and Chexpert show similar or improved performance at 20 epochs of basetraining.
This indicates that even a relatively small amount of self-supervised pretraining at the base level improves transfer performance. 
Furthermore, as mentioned in \S\ref{sec:exp}, the \texttt{Quickdraw} dataset has a large domain gap with ImageNet, and indeed, we observe that directly transferring ImageNet models with less base training leads to improved results on \texttt{Quickdraw}, but HPT maintains consistent performance regardless of the amount of basetraining.

The computation needed for 20 epochs of basetraining + 5000 iterations of pretraining on the target dataset is approximately equal to 100,000 iterations of pretraining on only the target dataset. For all datasets except Chest-X-ray-kids, HPT at 20 epochs of basetraining exceeded the best Target-only pretraining performance, which was $\geq 100$k iterations for all datasets.
Indeed, for RESISC, the HPT results at 20 epochs are worse than 200 and 800 epochs, but they still exceed the best Target-only pretraining results (the dashed, orange line). 

\subsection{Source Pretraining}
\begin{figure*}[t]
\tiny
\centering

\stackunder[5pt]{\includegraphics[width=5.3cm]{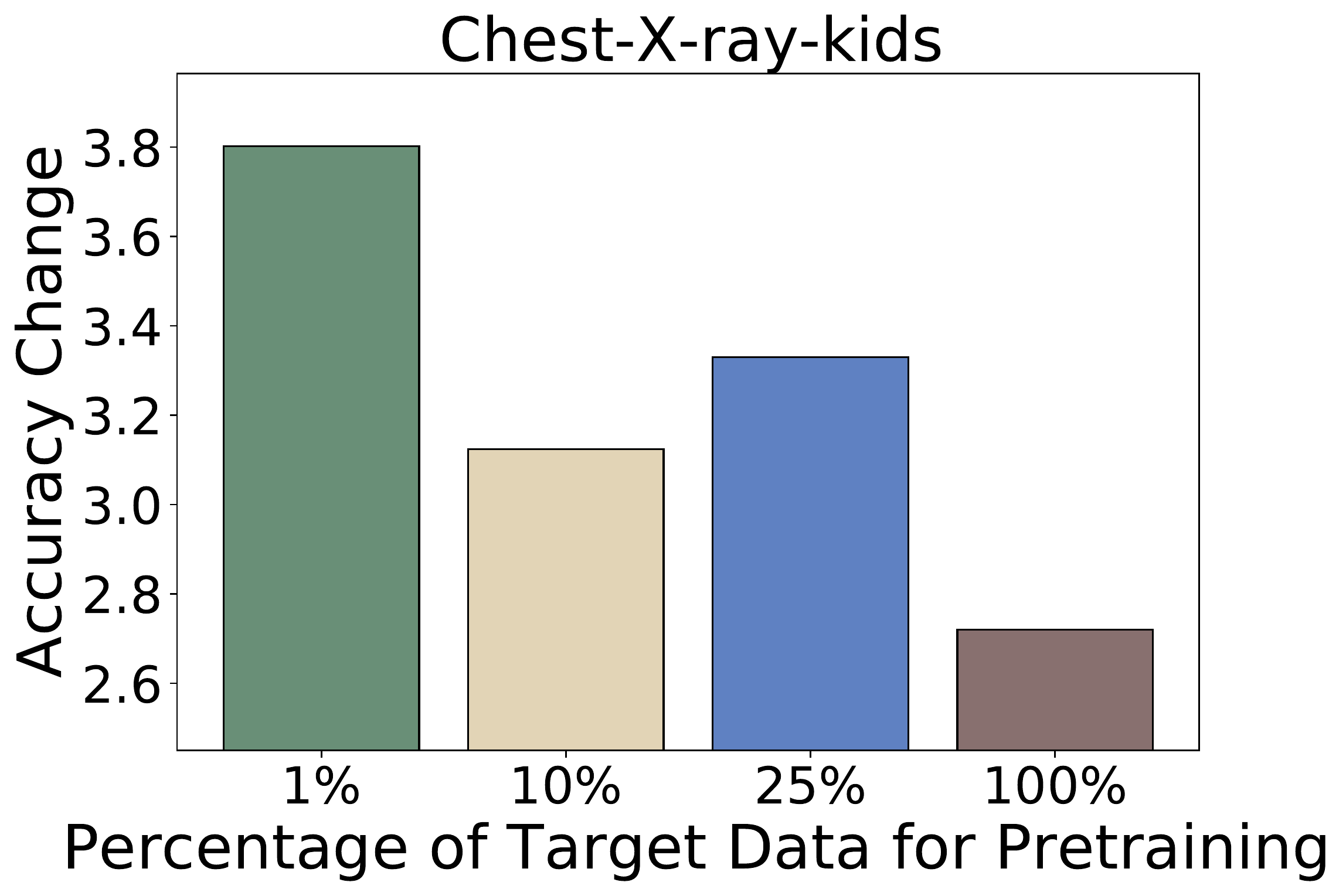}}{} 
\hspace{6pt}%
\stackunder[5pt]{\includegraphics[width=5.3cm]{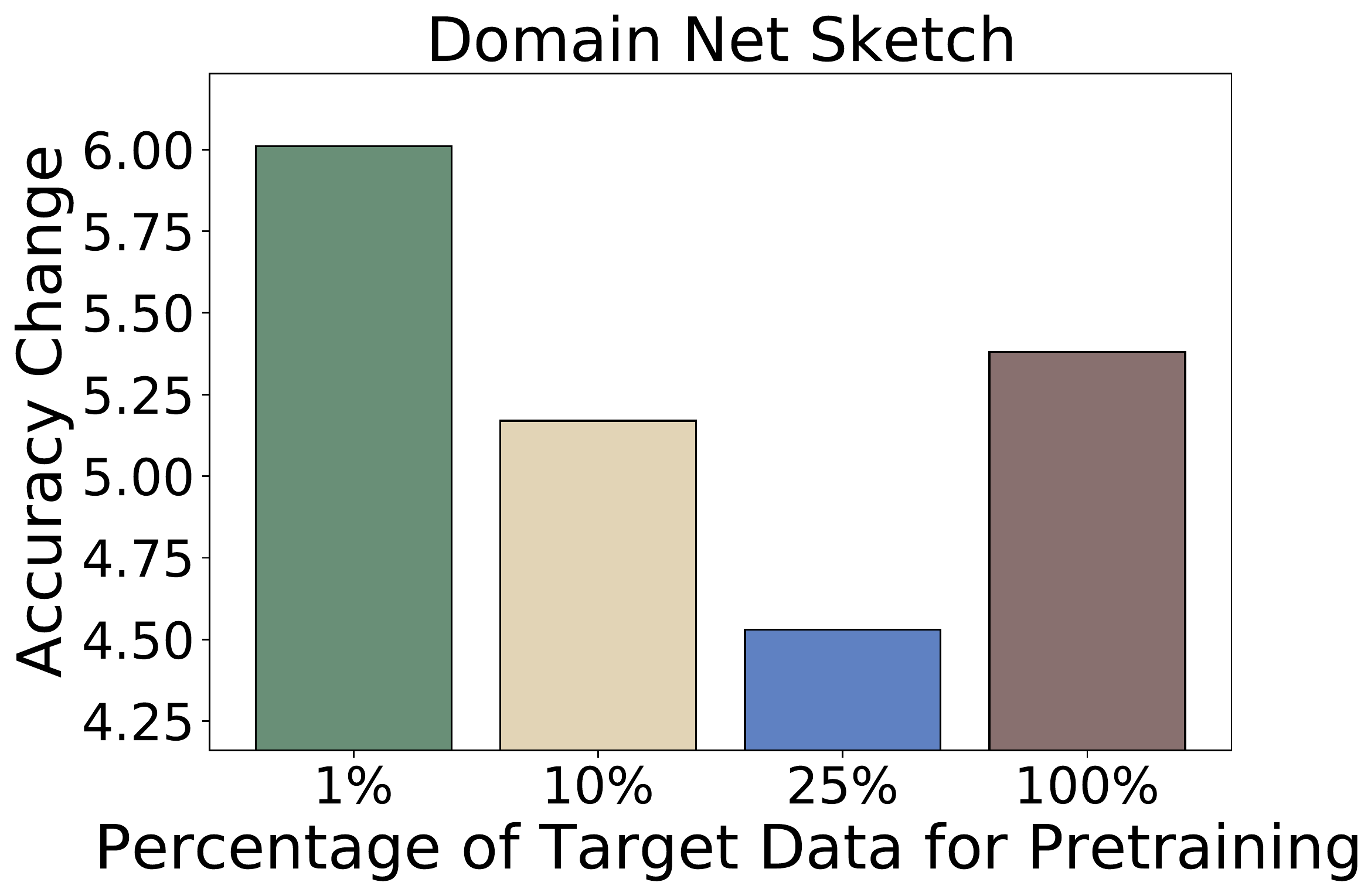}}{}
\hspace{6pt}%
\stackunder[5pt]{\includegraphics[width=5.3cm]{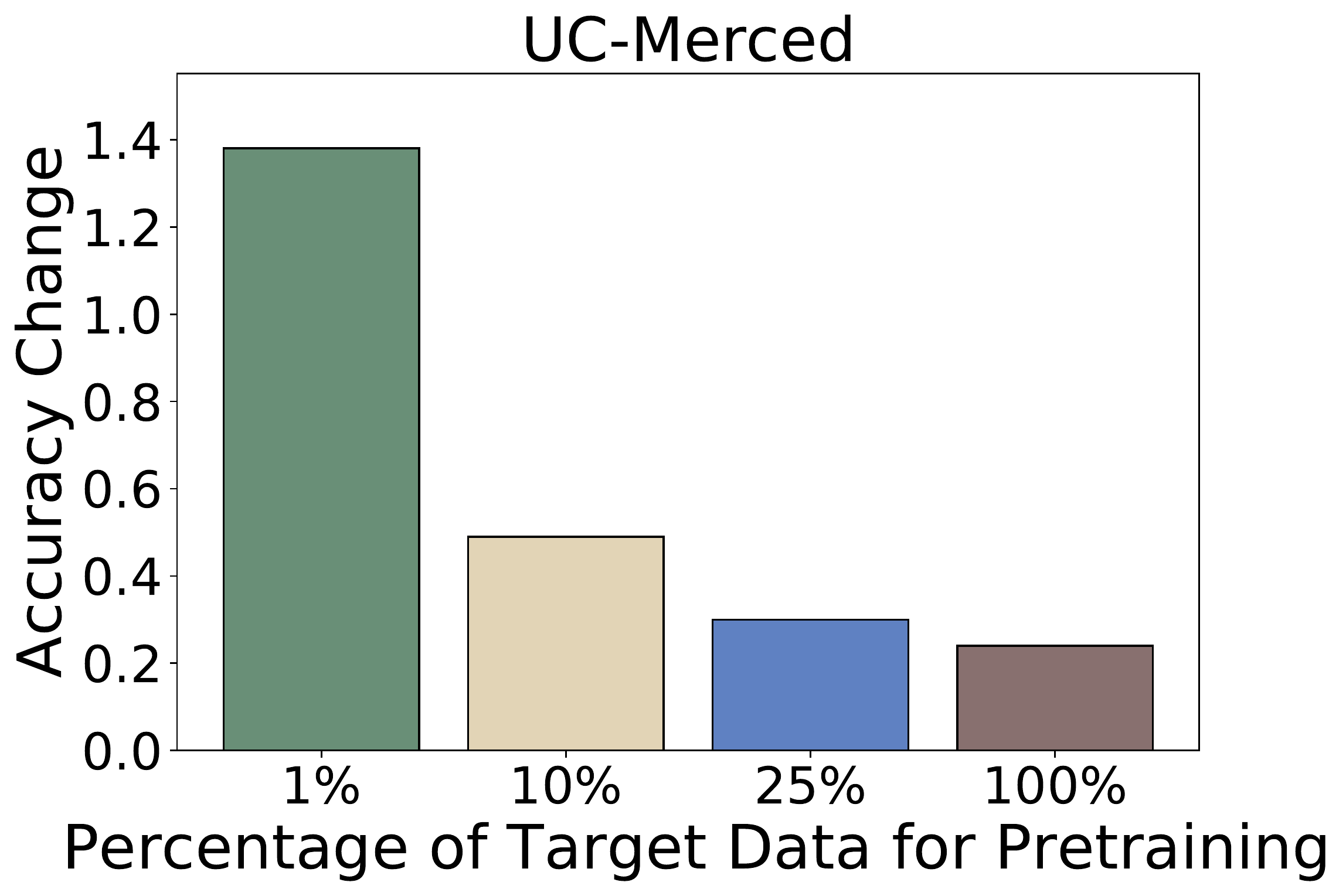}}{} 
\caption{These figures show the change in the linear evaluation results by adding the source pretraining for each of the target data amounts for the given datasets. We observed that for all three datasets, adding the source pretraining had a larger benefit as the amount of target data was reduced. In other words, these results show that the impact of pretraining on an intermediate source dataset decreases as the size of the target dataset increases.}
\label{fig:target-source-robustness}%
\end{figure*}
In this ablation, we investigated the impact of the source pretraining stage in the HPT framework as the amount of target data changes. Intuitively, we expected that the source pretraining stage would have less impact as the amount of target data increased. For this ablation, we used the three HPT frameworks studied in \S\ref{sec:exp}: ImageNet (base) then Chexpert (source) then Chest-X-ray-kids (target), ImageNet (base) then DomainNet \texttt{Clipart} (source) then DomainNet \texttt{Sketch} (target), and ImageNet (base) then RESISC (source) then UC-Merced (target). For each framework, we pretrained with \{1\%,10\%,25\%,100\%\} of the target data on top of the base+source model and on top of only the source model, before performing a linear evaluation with the target data. 

Figure~\ref{fig:target-source-robustness} shows the change in the linear evaluation results by adding the source pretraining for each of the target data amounts. We observed that for all three datasets, adding the source pretraining had a larger benefit as the amount of target data was reduced. In other words, these results show that the impact of pretraining on an intermediate source dataset decreases as the size of the target dataset increases.

\subsection{ResNet-18 Experiments}
\begin{figure*}[t]
\captionsetup[subfigure]{labelformat=empty}
\centering
\includegraphics[width=\linewidth]{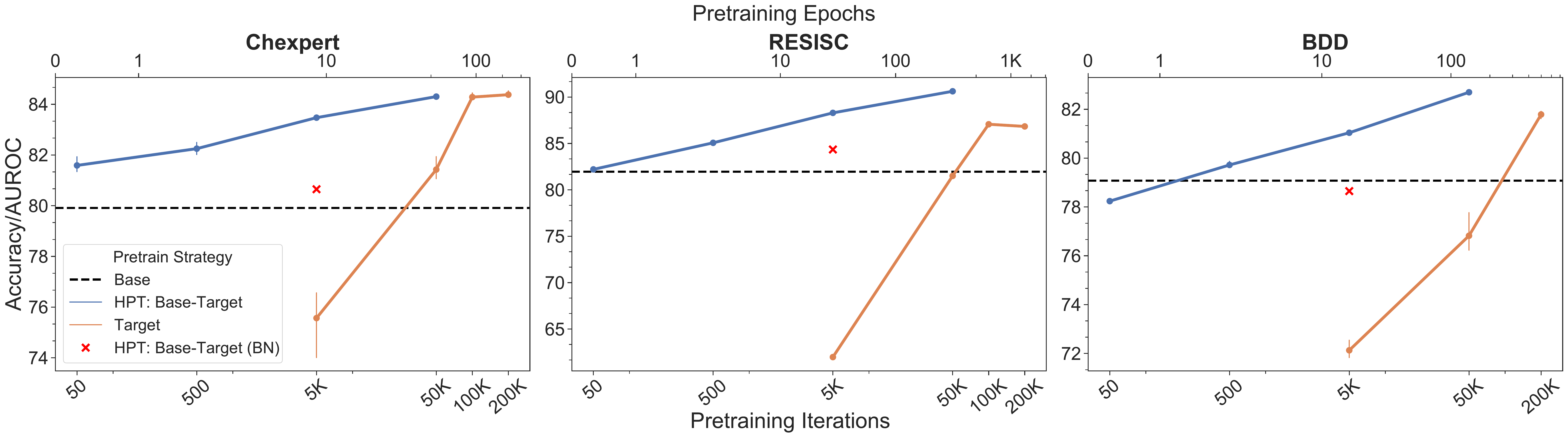}
\caption{This figure shows the HPT linear evaluation performance on BDD, Chexpert, and RESISC using a ResNet-18. For these datasets, we observe similar behavior as with ResNet-50 though the evaluation performance is lower for all datasets and HPT shows improved performance at 50K iterations for all datasets. Generalizing from this experiment, we expect HPT to be broadly applicable across architectures, and we will report additional, ongoing results in our online code repository.}
\label{fig:resnet18}%
\end{figure*}
Similar to Figure~\ref{fig:hpt-exp1-linear-analysis}, Figure~\ref{fig:resnet18} shows the same results using a ResNet-18 on BDD, Chexpert, and RESISC. For these datasets, we observe similar behavior as with ResNet-50, though the evaluation performance is lower. Generalizing from this experiment, we expect HPT to be broadly applicable across architectures, and we will report additional, community results in our online code repository.

\subsection{BYOL Experiments}

\begin{figure*}[tbh] 
    \centering 
    \includegraphics[width=1.0\linewidth]{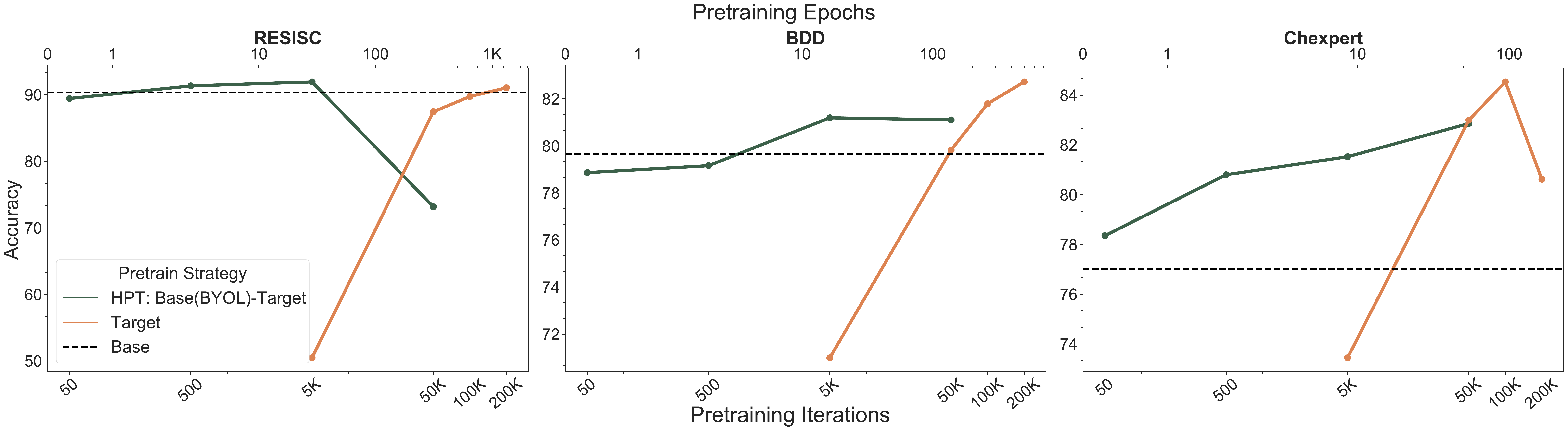}
     \caption{Linear eval with BYOL~\cite{byol} pretraining on RESISC, BDD and Chexpert. For each dataset, we we train a generalist model for 200 epochs on ImageNet (Base). We then train the whole model from 50-50K iterations (HPT: Base (BYOL)-Target). We compare the HPT model with a model trained from a random initialization on the target data (Target). We use a linear evaluation to evaluate the quality of the learned representations.}
     \label{fig:byol}
\end{figure*}

All of the pretraining results in the main paper are based on MoCo, here we use BYOL~\cite{byol} for pretraining and perform linear evaluation on RESISC, BDD and Chexpert. As shown in Figure~\ref{fig:byol}, we observe similar results as Figure~\ref{fig:hpt-exp1-linear-analysis} in the main paper. Generalizing from this experiment, we expect HPT to be broadly applicable across different Self-supervised pretraining methods, and we will report additional, community results in our online code repository. 

\subsection{Representational Similarity}
    We examine how the similarity of representations change during pretraining with the self-supervised base model HPT, supervised base model HPT, and self-supervised training on only the target data.
    
    \subsubsection{Defining metrics}
    We explore different metrics for measuring the similarity of two different pretrained models.
    The first metric we explore is the Intersection over Union (IoU) of 
    misclassified images.
    The IoU is the ratio between the number of images misclassified by both models and the number of images misclassified by at least one model.
    
    \begin{algorithm}
    \caption{IoU}\label{IoU}
    \begin{algorithmic}[1]
    \REQUIRE{data, labels, modelA, modelB}
    \STATE predictionA = modelA(data) 
    \STATE predictionB = modelB(data)
    \STATE commonErrors, totalErrors $\leftarrow 0$
    \FORALL {$l,pA,pB \in $ zip(labels, predictionA, predictionB)}
    \IF{$l == pA$ \AND $l == pB$} 
    \STATE commonErrors+=1 \ENDIF
    \IF{$l == pA$ \OR $l == pB$} 
    \STATE totalErrors+=1 \ENDIF
    \ENDFOR
    
    \textbf{return:} $\frac{\text{commonErrors}}{\text{totalErrors}}$
    \end{algorithmic}
    \end{algorithm}

    The activation similarity metric we used wass RV2 
    \cite{DBLP:journals/corr/abs-1912-02260} which, 
    instead of comparing predictions, aims to compare the similarity between two different layers' outputs computed on the same data.
    The pseudocode for our algorithm is shown in Algorithm~\ref{rv2}.
    
    \begin{algorithm}
    \caption{RV2}\label{rv2}
    \begin{algorithmic}
    \REQUIRE{activation A, activation B} 
    \COMMENT{Both activations are size $n\times p$, 
    where $n$ is the number of data points, and
    $p$ is the size of the layer output}
    
    \STATE $A'$ = $AA^T$
    
    \STATE $B'$ = $BB^T$
    \STATE $A'' = A' - diag(A')$
    \STATE $B'' = B' - diag(B')$
    \STATE
    \textbf{return} 
    \[
    \frac{\text{tr}(A''B''^T)}
    {
    \sqrt{\text{tr}(
     A''A''^T) *
      \text{tr}(
     B''B''^T))}}
    \]

    \end{algorithmic}
    \end{algorithm}
    
   Because many of our evaluations were performed by finetuning a linear layer over the outputs of the final convolutional layer of a pretrained model, we evaluated activations for the final convolutional layer and the linear layer finetuned on top of it.

    For this analysis, we studied Resisc, UC Merced, Domain Net, Chexpert, Chest-X-ray-kids, and BDD datasets.
    \subsubsection{Effect of model initialization?}
    In this section, we present the results of two sets of experiments intended to analyze the representation of HPT models initialized from different base models. Overall, we found that HPT models with different base models learn different representations.
    
    \textbf{Random effects}: 
        In order to examine the effect of the random seed on model errors,
        we trained each combination of (basetrain, pretrain steps)
        five times for the RESISC dataset. 
    
\begin{figure}[ht]
        \centering
        \includegraphics[width=.35\textwidth]{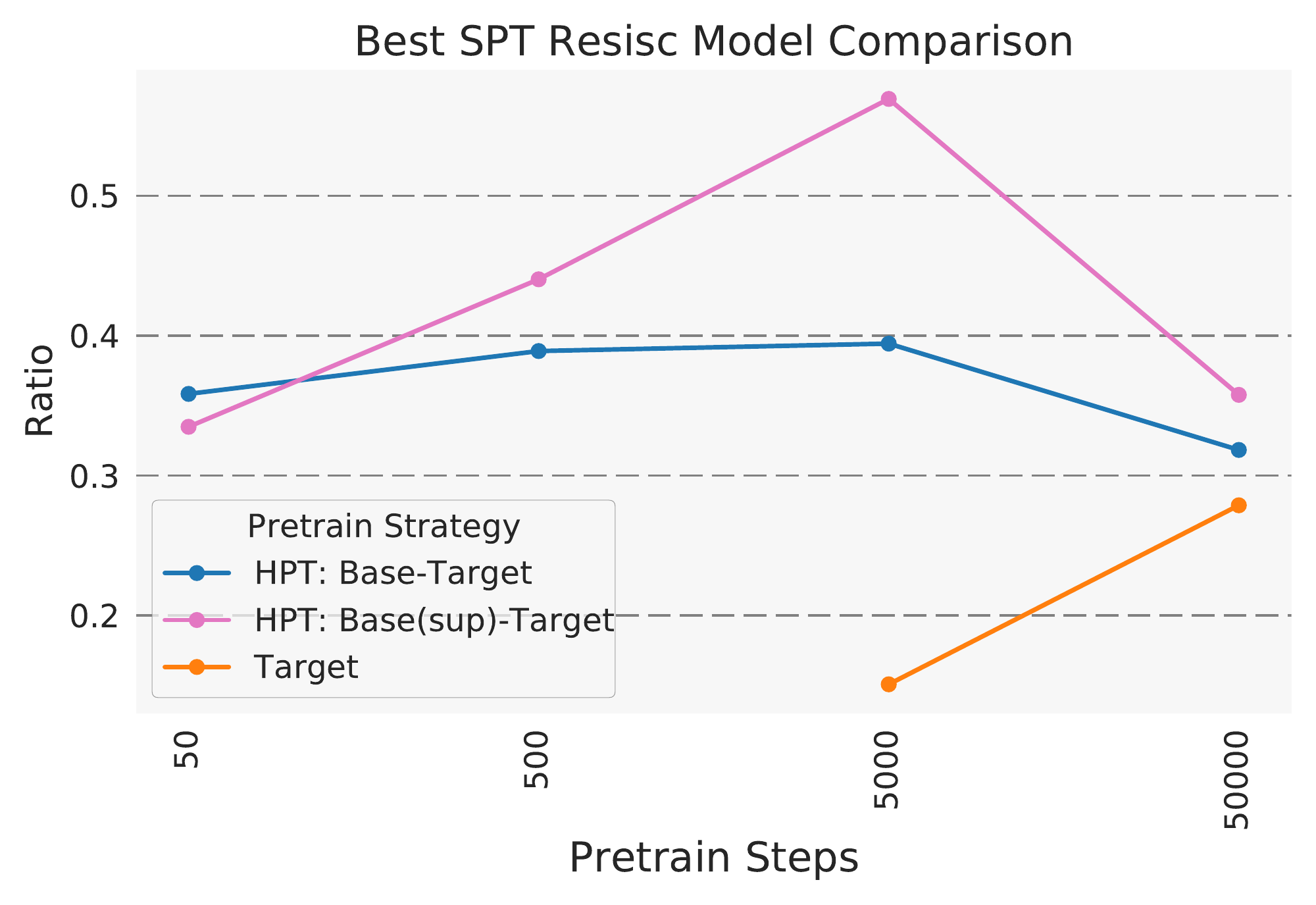}
        \includegraphics[width=.35\textwidth]{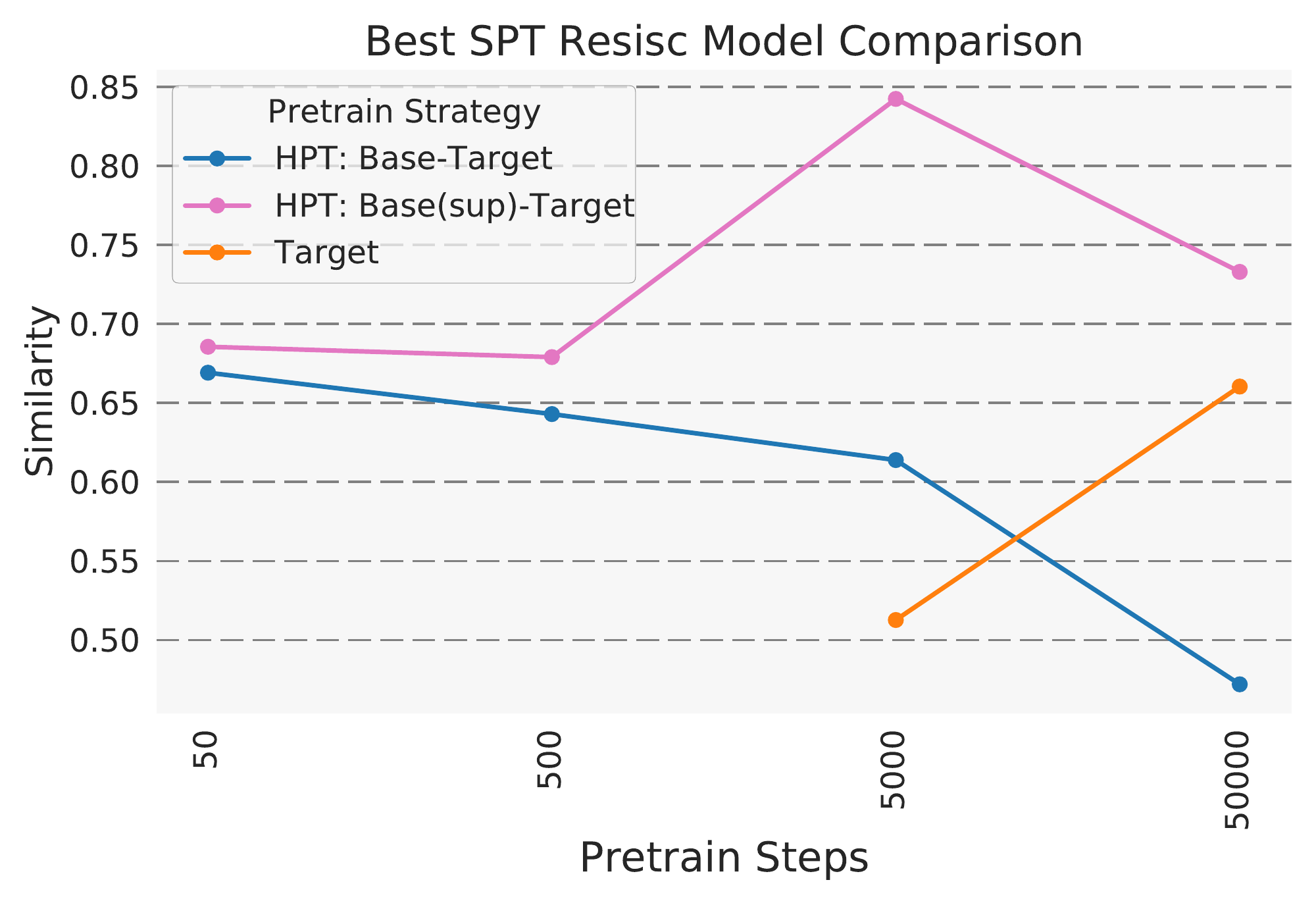}
        \includegraphics[width=.35\textwidth]{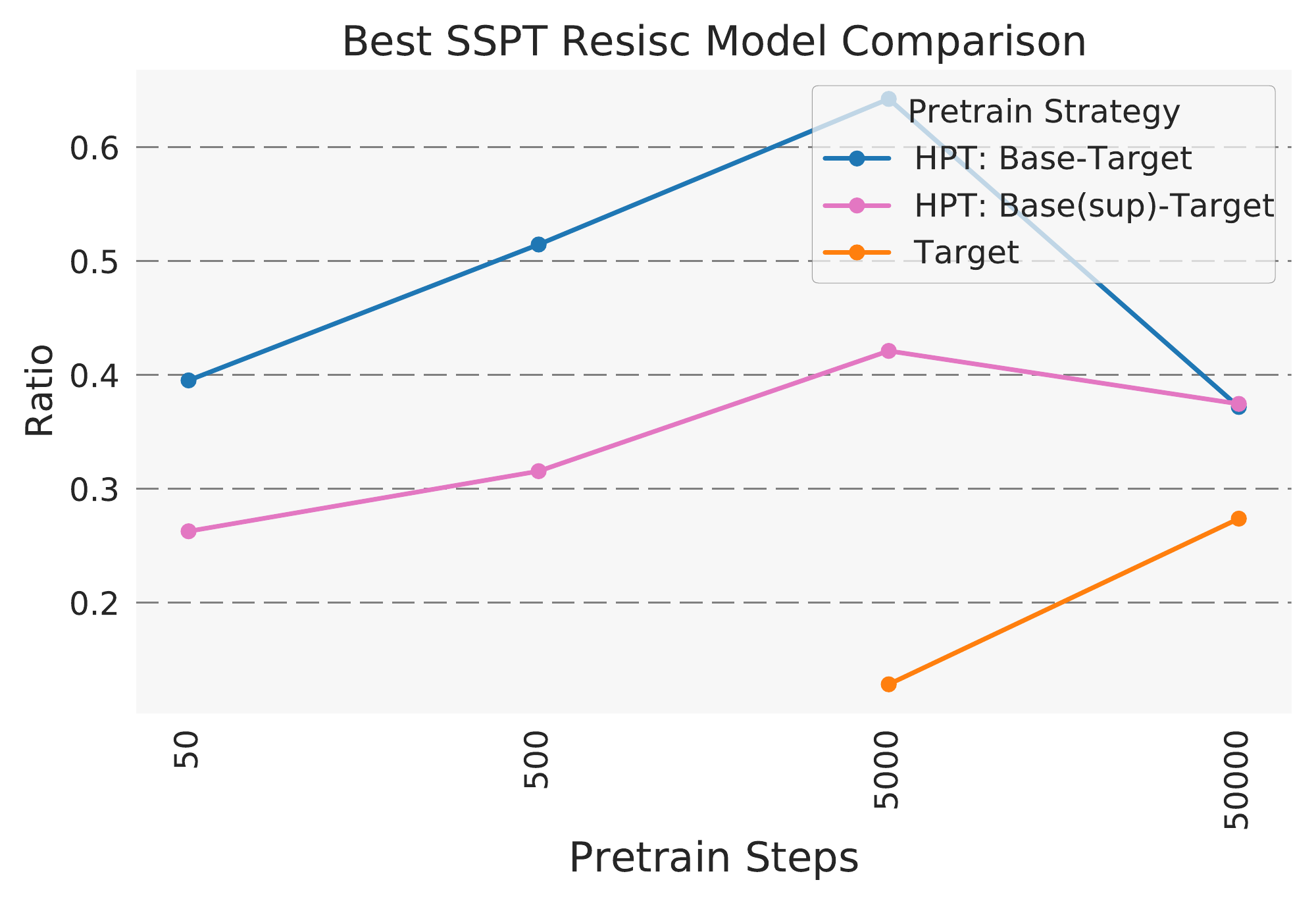}
        \includegraphics[width=.35\textwidth]{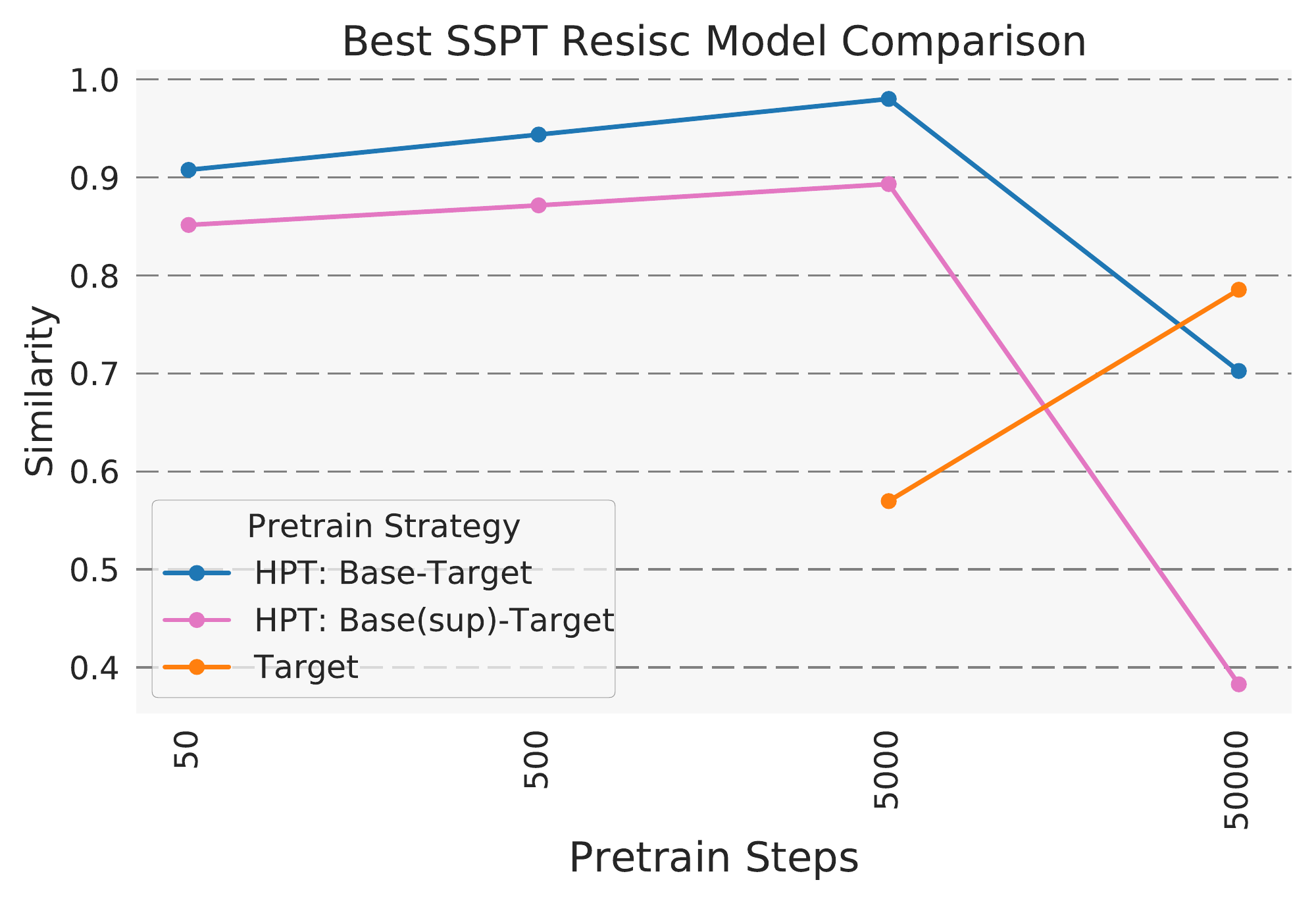}
        \caption{Supervised and semi-supervised models trained with random seeds. Left IoU, right final linear layer similarity.
        Top supervised, bottom semi-supervised}
        \label{fig:random-seeds}
        \end{figure}

        Representations which share the same basetrain and number of pretrain steps result in more similar errors and have 
        much more similar representations than other combinations, see Figure~\ref{fig:random-seeds}. The IoU is typically between 0.5 and 0.6, meaning that roughly half of the total error caused by mispredictions are unique between these runs. The similarity is generally much higher, but is varies depending on the dataset.

    \textbf{Same basetrain}: Out of three different basetrain configurations (random-initialized, supervised basetrain, and self-supervised basetrain), different runs starting from the same basetrain are 
        typically more similar to each other by both metrics than those with different basetrains. This holds true across models trained for different number of iterations with different random seeds, and further adds to the notion that models learn different representations based on their starting point.  However, after overfitting, models are less likely to follow this trend and become dissimilar to ones with the same basetrain. We believe this is most likely due to models learning representations that mainly distinguish input data and thus all become dissimilar to the best performing model.

        We determined how similar models were using our two metrics of layer similarity and IoU errors of models with different basetrains and number of pretrain steps compared to the best performing model. All  models used the same pretrain dataset and target dataset for evaluation.
        We focused on similarity to the highest performing model (in terms of both basetrain and training iterations) to see if different basetrains converged to the same representation.

\begin{figure}[!ht]
        \centering
        \includegraphics[width=.35\textwidth]{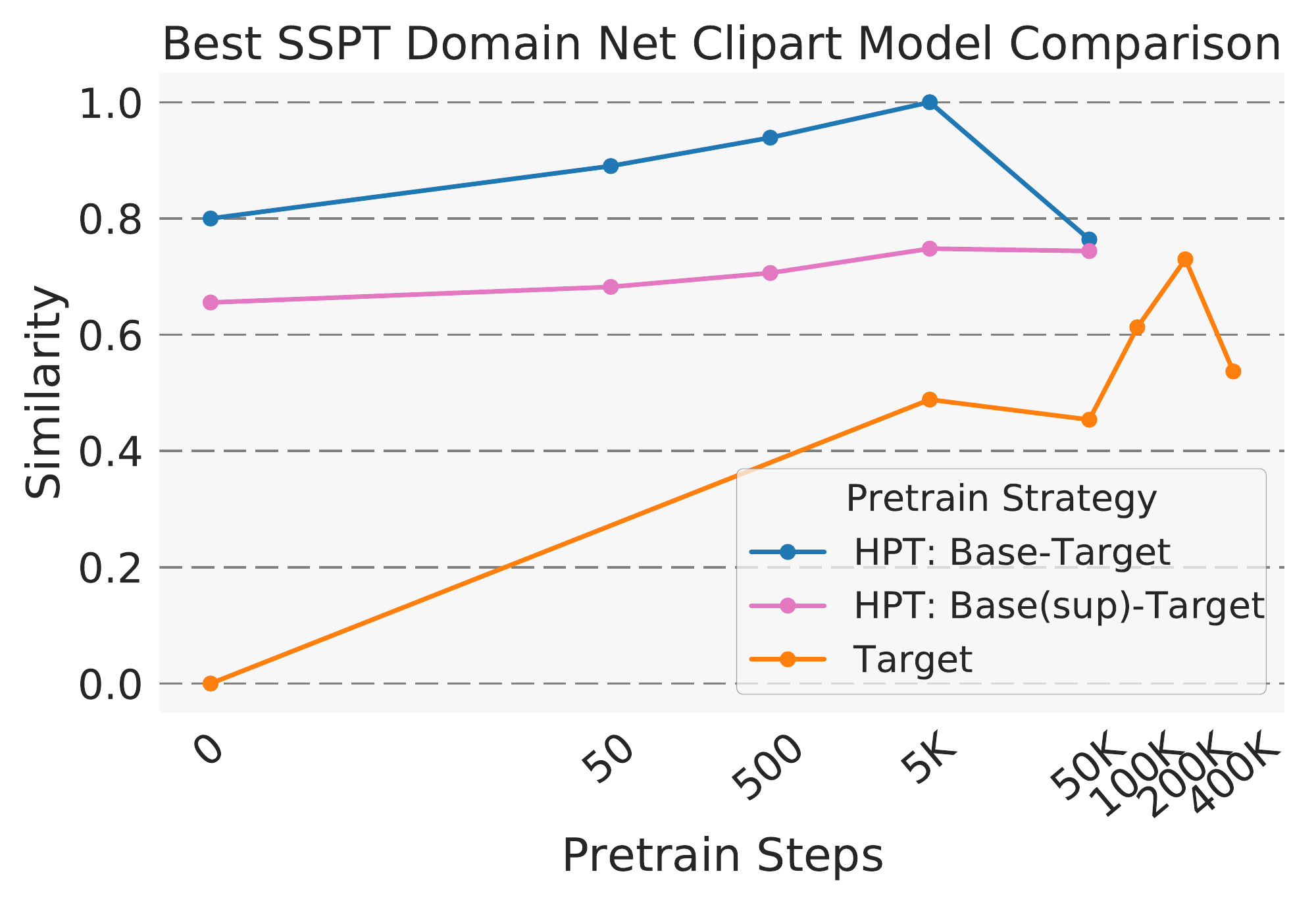}
        \includegraphics[width=.35\textwidth]{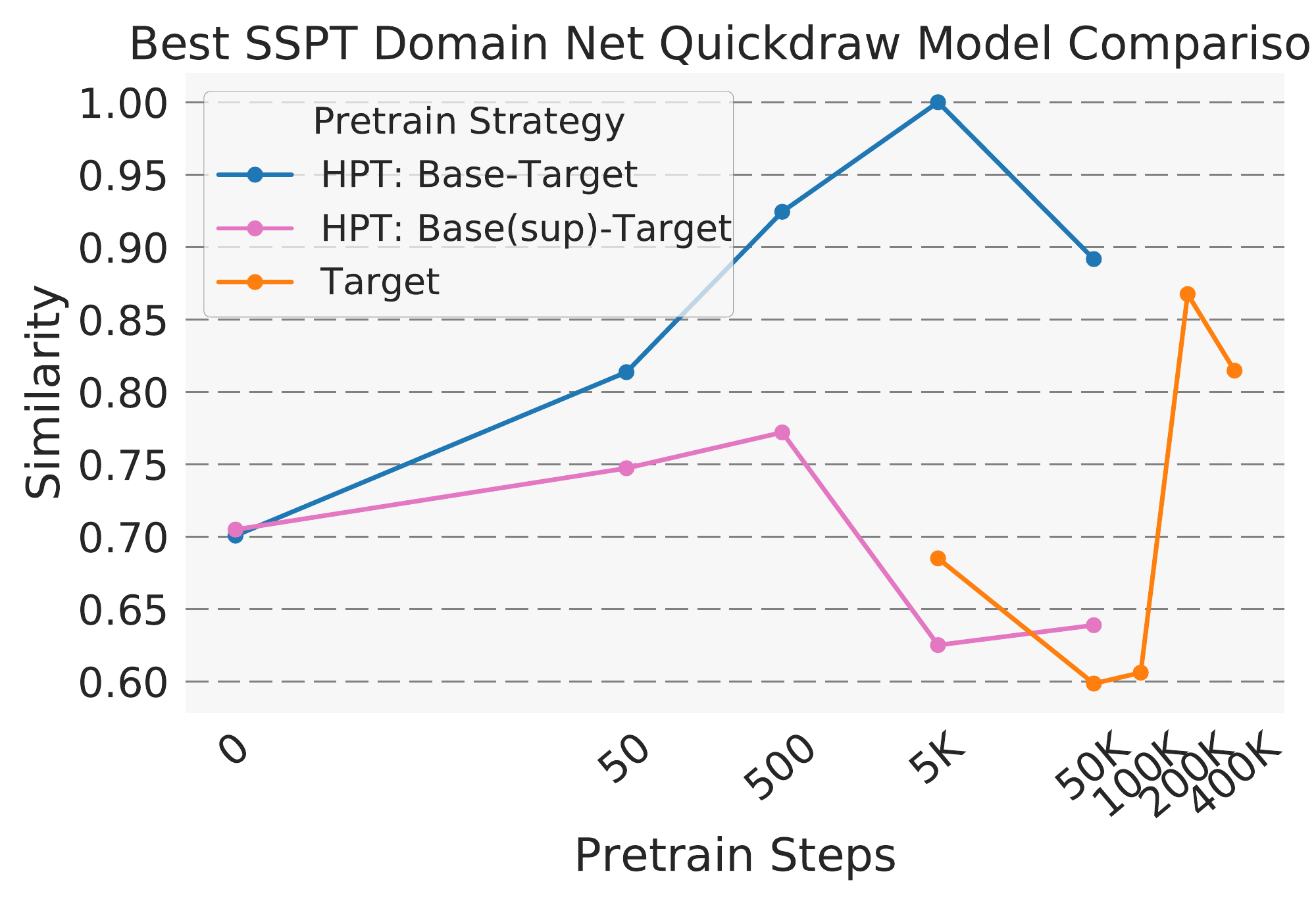}
        \includegraphics[width=.35\textwidth]{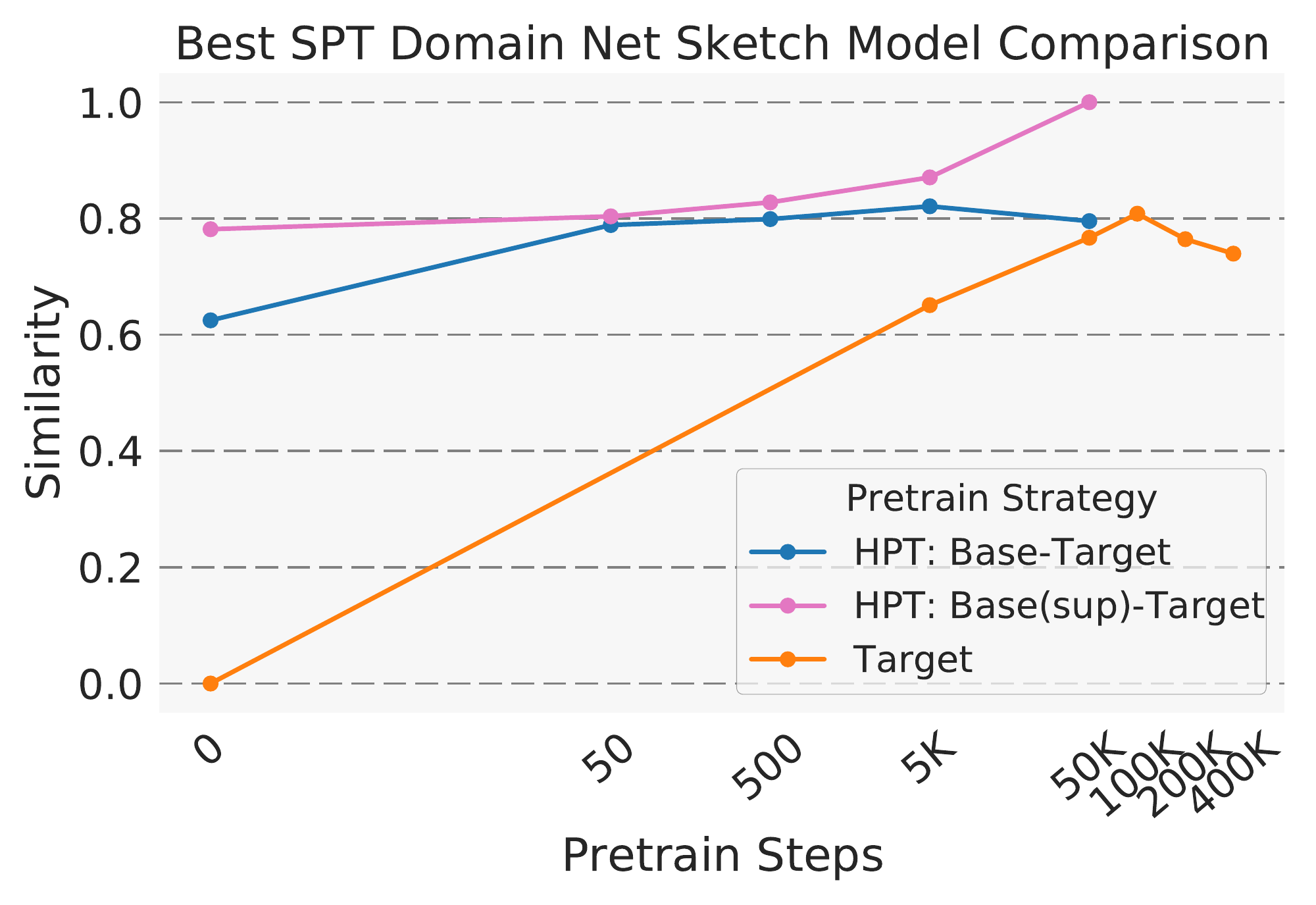}
        \includegraphics[width=.35\textwidth]{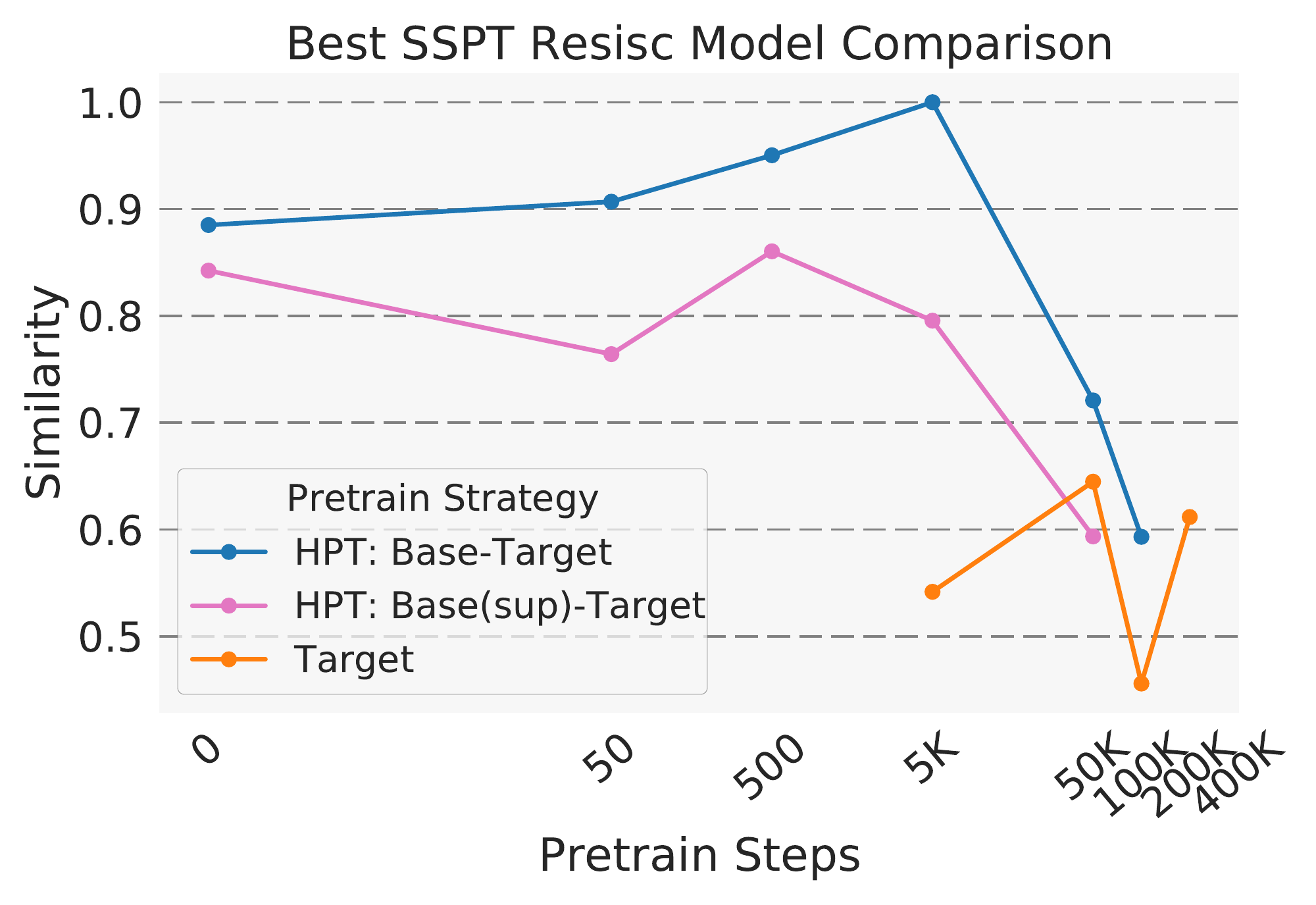}
        \caption{Linear Classification Layer comparison. For all the following graphs, SPT refers to Supervised Pretrain with ImageNet, and SSPT refers to Self-Supervised Pretrain with MoCo. The best model is the model that attains 1.0 similarity, and every other model is compared to that point. Up until the target model's pretrain steps (before overfitting), we can see that the similarity between the linear classification layers of every model with a different basetrain is much less than the models with the same basetrains.}
        \label{fig:linear}
        \end{figure}

       Linear classification layers from the same basetrain are consistently more similar than those with different basetrains.
       This trend becomes less consistent after around 50,000 iterations of training, which is also when the self-supervised models we examined start overfitting.
       In Figure~\ref{fig:linear} we plot the similarity of linear layers for each model relative to the best performing models on four sample datasets.
        
\begin{figure*}[!ht]
        \centering
        \includegraphics[width=.35\textwidth]{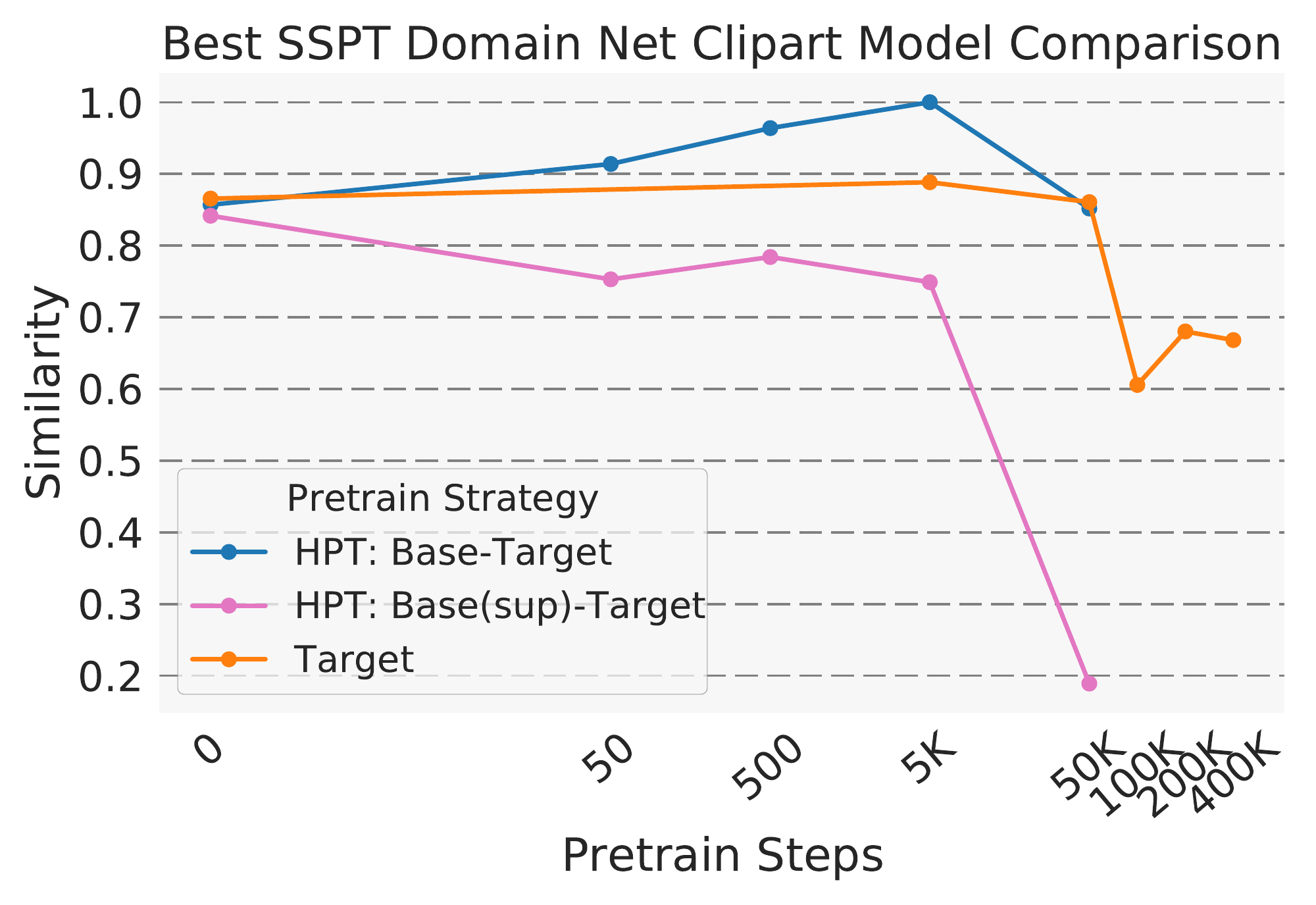}
        \includegraphics[width=.35\textwidth]{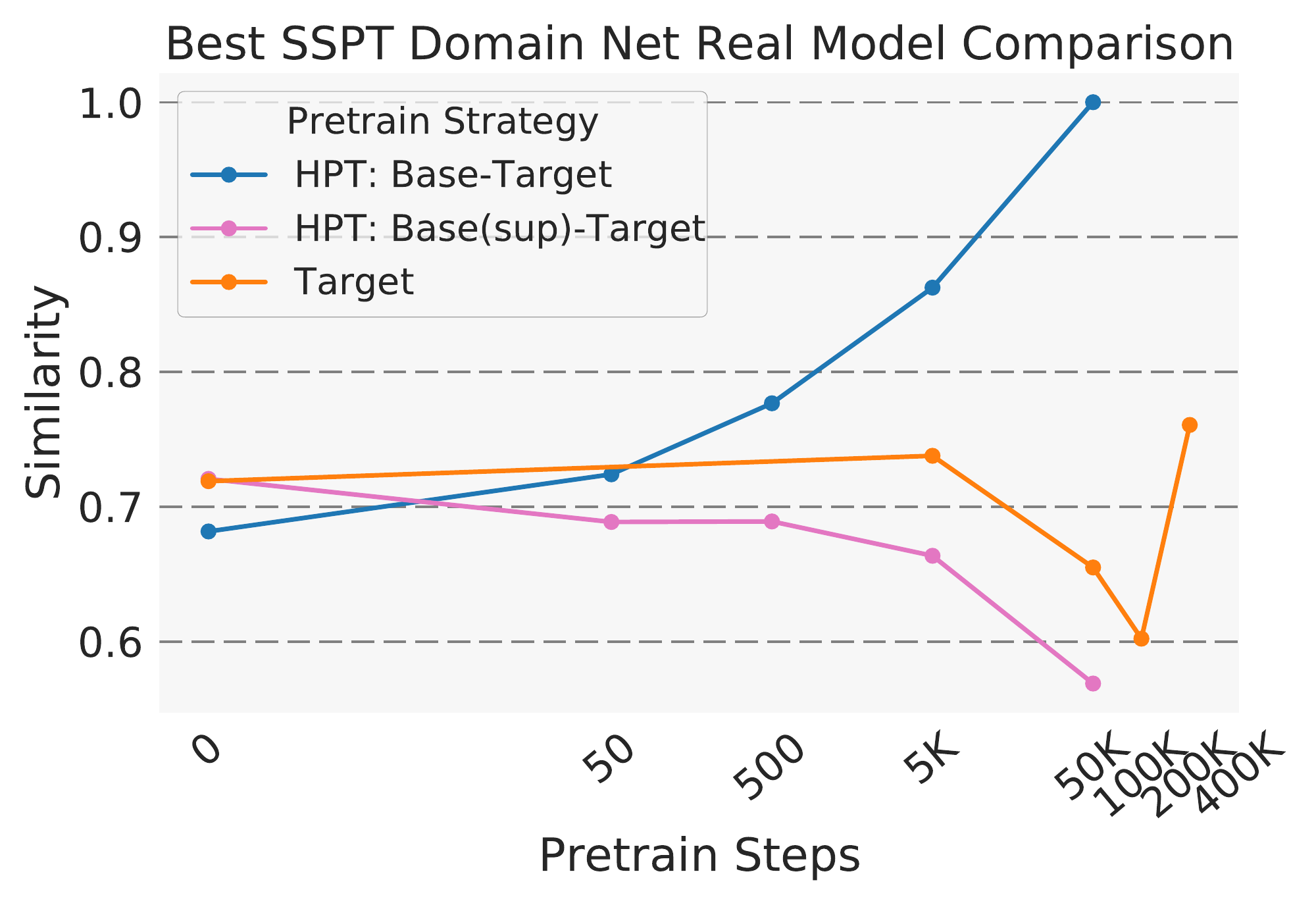}
        \includegraphics[width=.35\textwidth]{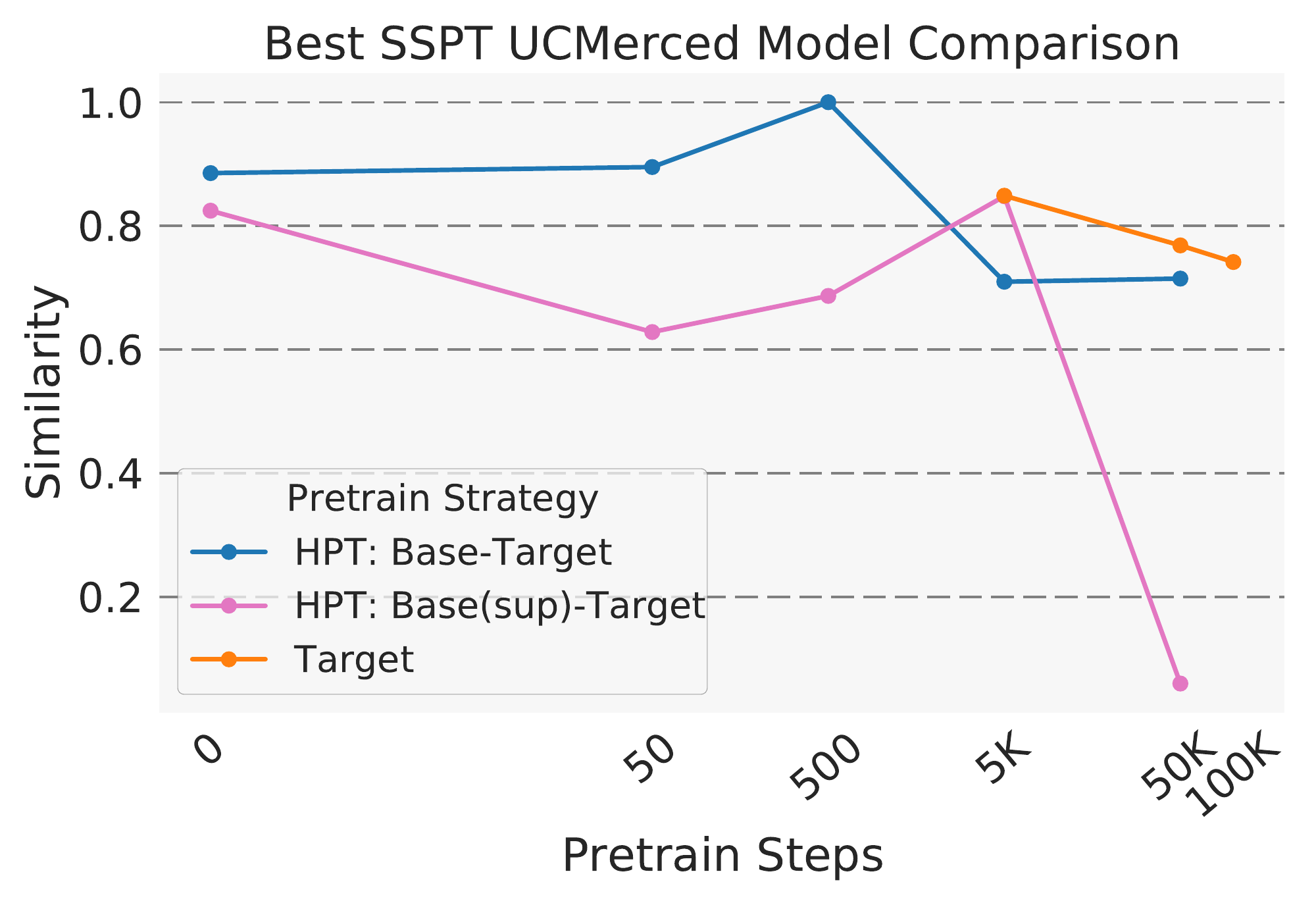}
        \includegraphics[width=.35\textwidth]{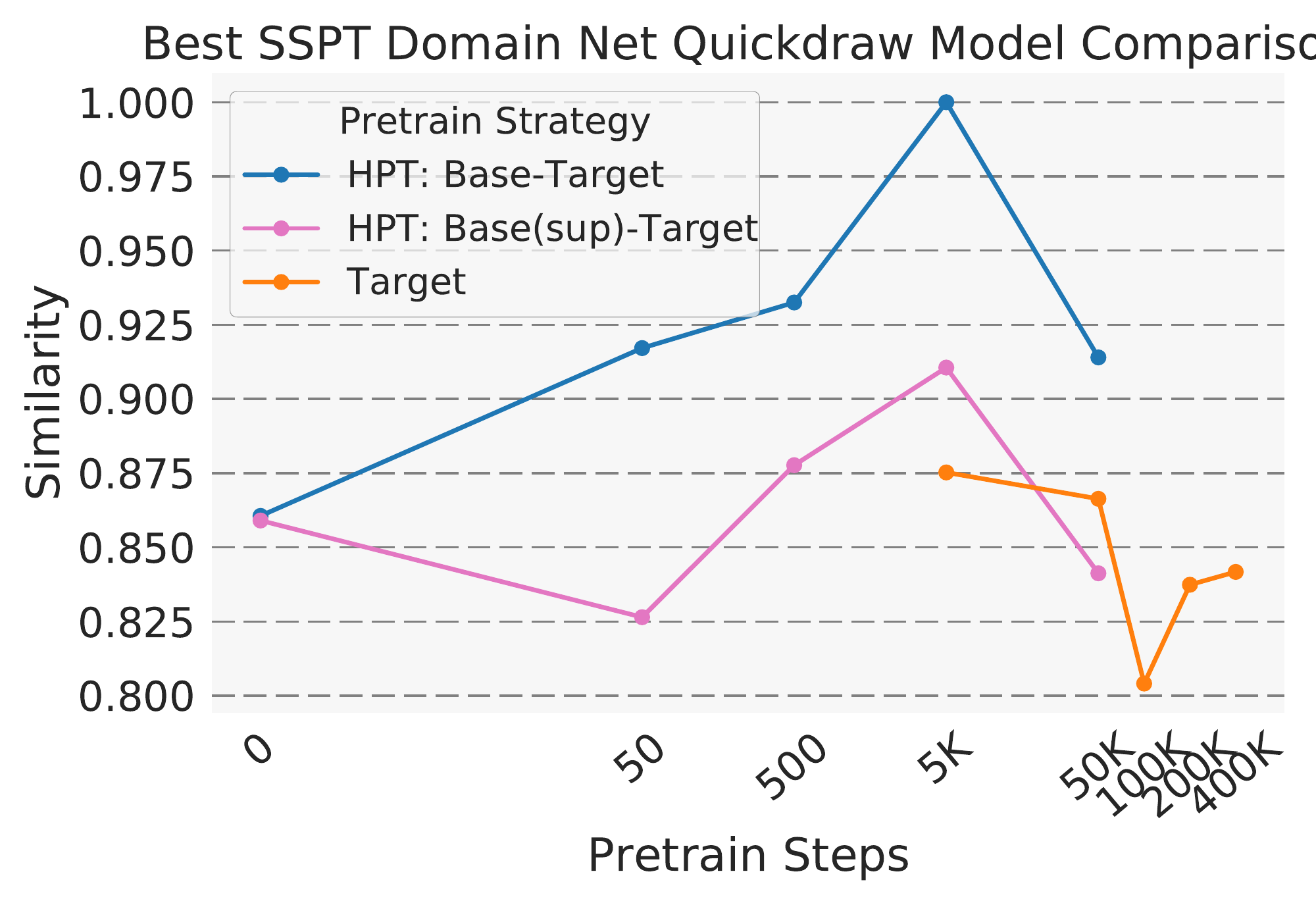}
        \caption{
        Final Convolutional Layer comparison. 
        We can see that the models with basetrains similar to the best model consistently have higher similarity scores. 
        }
        \label{fig:conv}
        \end{figure*}

        This same observation held when comparing the similarity of the final convolutional layers instead of the linear classification layers as shown in Figure~\ref{fig:conv}. Overall, the convolutional layers trained from the same basetrain were more similar to each other than other basetrains. There were just a few points of comparison that deviated from the trend in a little under half of the datasets we tested.        %

\begin{figure*}[!ht]
        \centering
        \includegraphics[width=6cm]{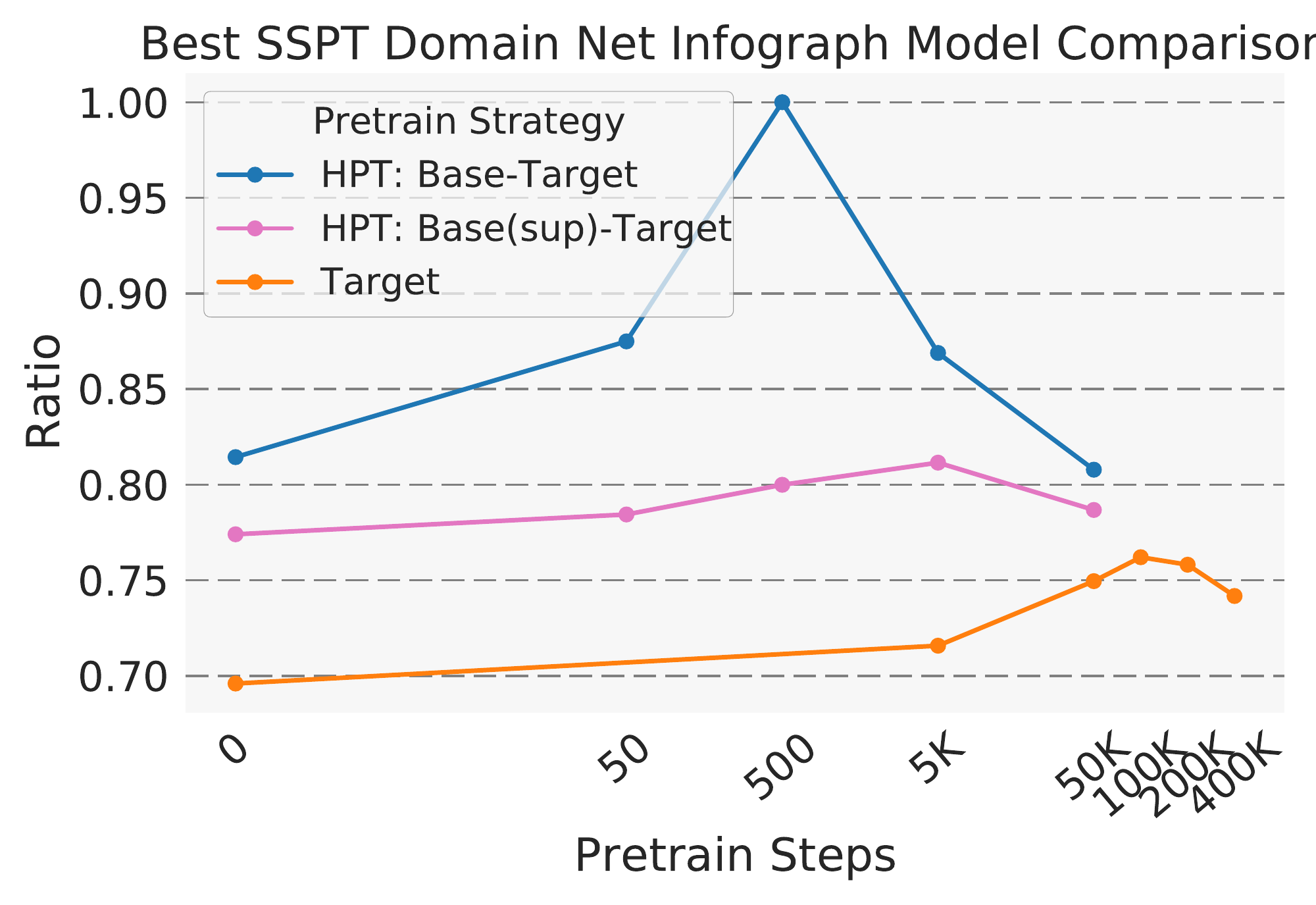}
        \includegraphics[width=6cm]{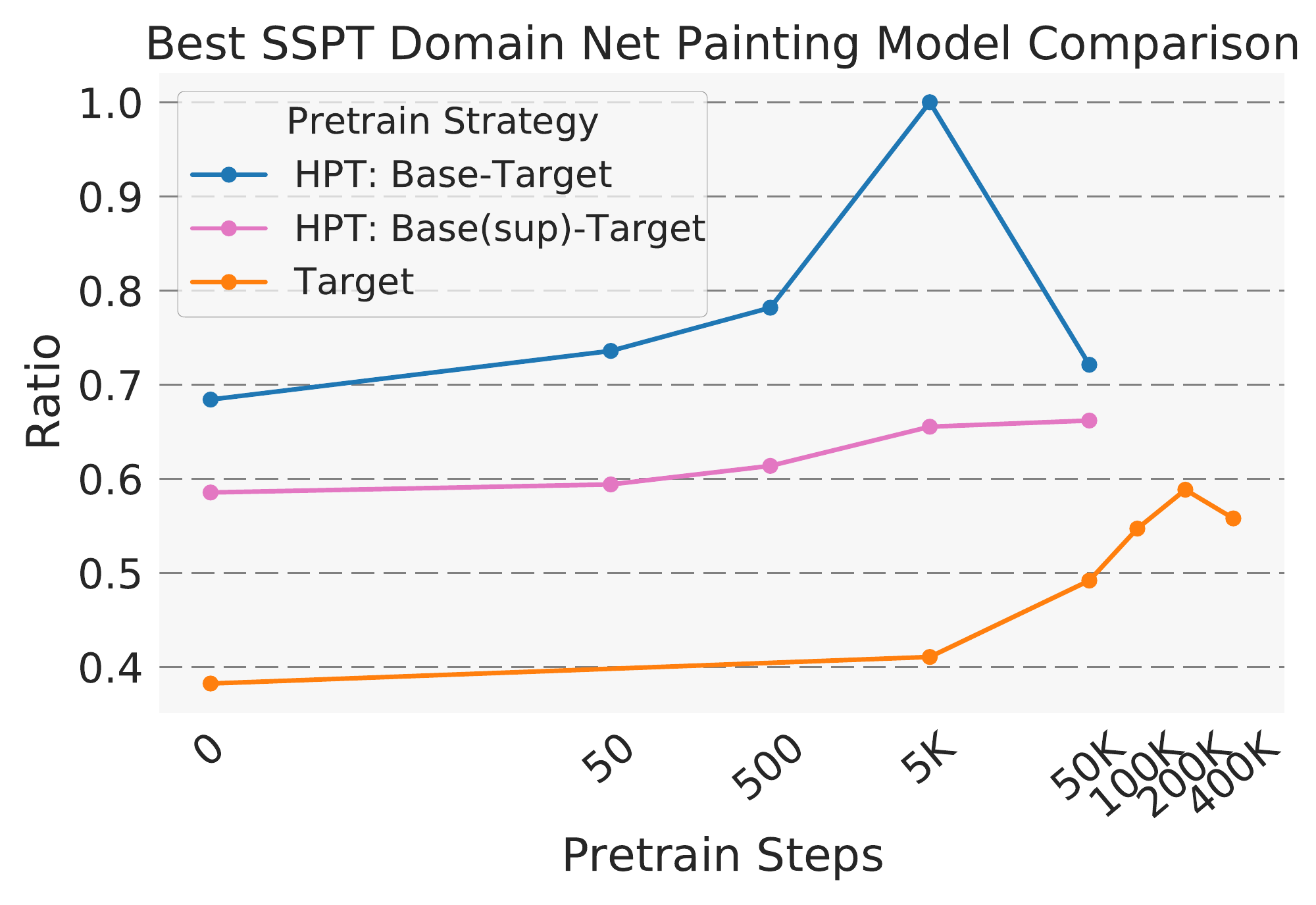}
        \includegraphics[width=6cm]{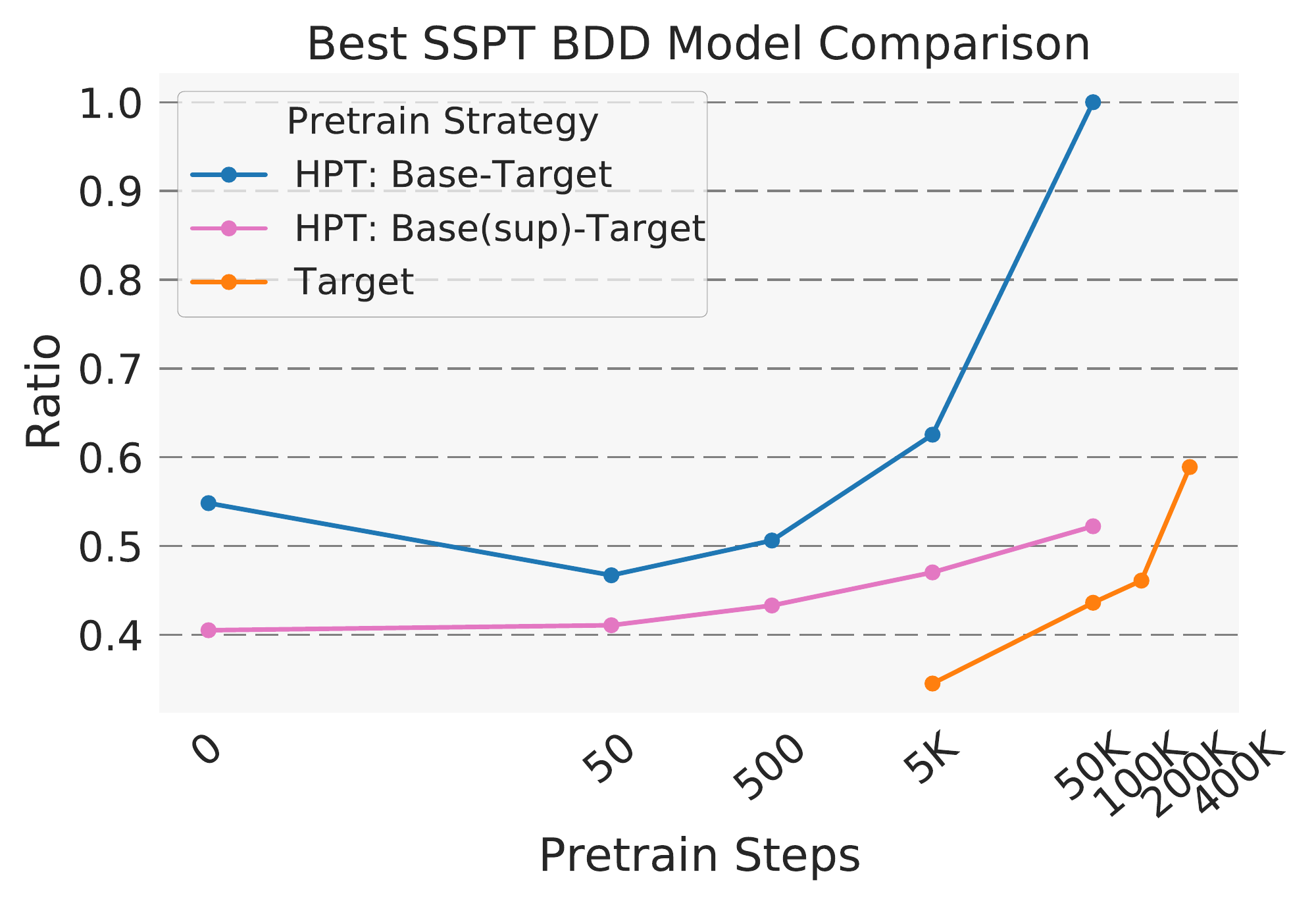}
        \includegraphics[width=6cm]{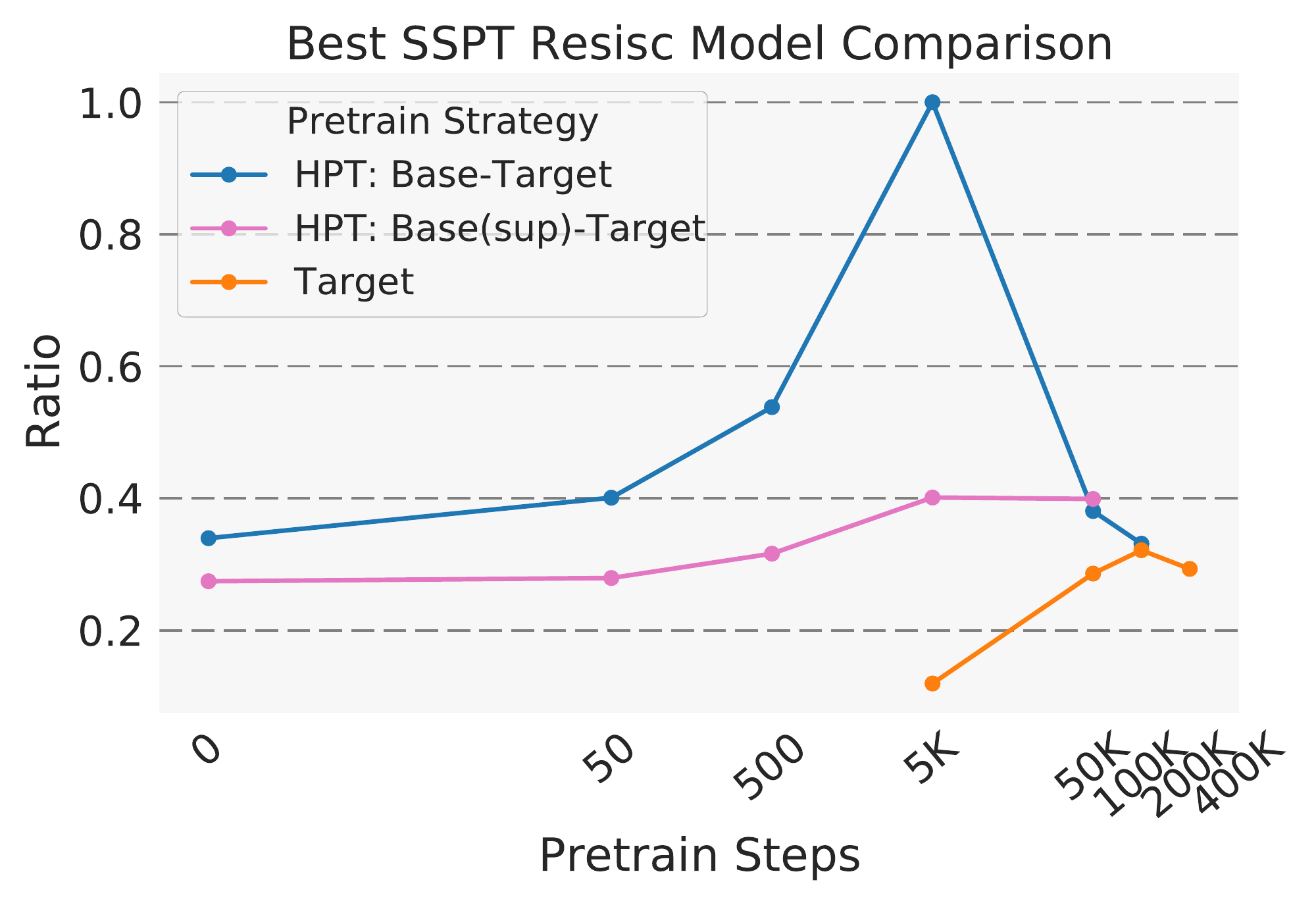}
        
        \caption{IoU comparison. IoU scores are consistently larger for models with the same basetrain as the best model compared to those of different basetrains, indicating that more similar errors are made by models with a similar initialization.} 
        \label{fig:iou}
        \end{figure*}

        The IoU error comparisons in Figure~\ref{fig:iou} showed a similar trend to the linear layers, 
        with models with the same basetrain being more similar on almost all random seeds and datasets until 50,000 iterations.

        Finally, we performed a significance test to demonstrate the significant difference between representations learned from different basetrain models. 
        We trained five pairs of models with identical hyperparameters, but with different basetrains (supervised vs self-supervised) and a different random seed.
        We also trained five pairs of models with identical hyperparameters, but with the same basetrain (self-supervised) and a different random seed.
        All models used ImageNet for the basetrain and RESISC for the pretrain.
        We calculated the linear layer similarity and IoU for each pair of models, and performed a Welsh's t-test on the results.
        We found that the similarities and IoUs were significantly different.
        The different basetrains had a mean similarity of 0.78 while the identical basetrains had a mean similarity of 0.98 ($p=2\times10^{-4}$).
        The different basetrains had a mean IoU of 0.40 while the identical basetrains had a mean IoU of 0.61 ($p=2\times10^{-6}$).

\end{document}